\documentclass[10pt,twocolumn,letterpaper]{article}

\usepackage{cvpr}      

\usepackage{graphicx}
\usepackage{amsmath}
\usepackage{amssymb}
\usepackage{booktabs}
\usepackage{gensymb}
\usepackage{bbm}
\usepackage{multirow}
\usepackage{array}
\usepackage{float}
\usepackage[accsupp]{axessibility}  

\usepackage[pagebackref,breaklinks,colorlinks]{hyperref}

\usepackage[capitalize]{cleveref}
\crefname{section}{Sec.}{Secs.}
\Crefname{section}{Section}{Sections}
\Crefname{table}{Table}{Tables}
\crefname{table}{Tab.}{Tabs.}

\def\cvprPaperID{XXXX} 
\def\confName{CVPR}
\def\confYear{2022}

\begin{document}


\title{Self-Supervised Keypoint Discovery in Behavioral Videos}

\author{Jennifer J. Sun$^\textnormal{1}$\footnotemark[1] $\quad$
Serim Ryou$^\textnormal{1}$\footnotemark[1]~~\!\footnotemark[2] $\quad$ 
Roni H. Goldshmid$^\textnormal{1}$ $\quad$
Brandon Weissbourd$^\textnormal{1}$ $\quad$
John O. Dabiri$^\textnormal{1}$ $\quad$ \\
David J. Anderson$^\textnormal{1}$ $\quad$
Ann Kennedy$^\textnormal{2}$ $\quad$
Yisong Yue$^\textnormal{1,3}$ $\quad$
Pietro Perona$^\textnormal{1}$  
\and
$^\textnormal{1}$Caltech$\quad\quad$
$^\textnormal{2}$Northwestern University$\quad\quad$
$^\textnormal{3}$Argo AI$\quad\quad$\\
{\small Code \& Project Website: \url{https://sites.google.com/view/b-kind}}
\vspace{-0.3em}
}

\twocolumn[{%
\renewcommand\twocolumn[1][]{#1}%
\maketitle
\begin{center}
    \centering
    \captionsetup{type=figure}
    \vspace{-0.1in}
    \includegraphics[width=\linewidth]{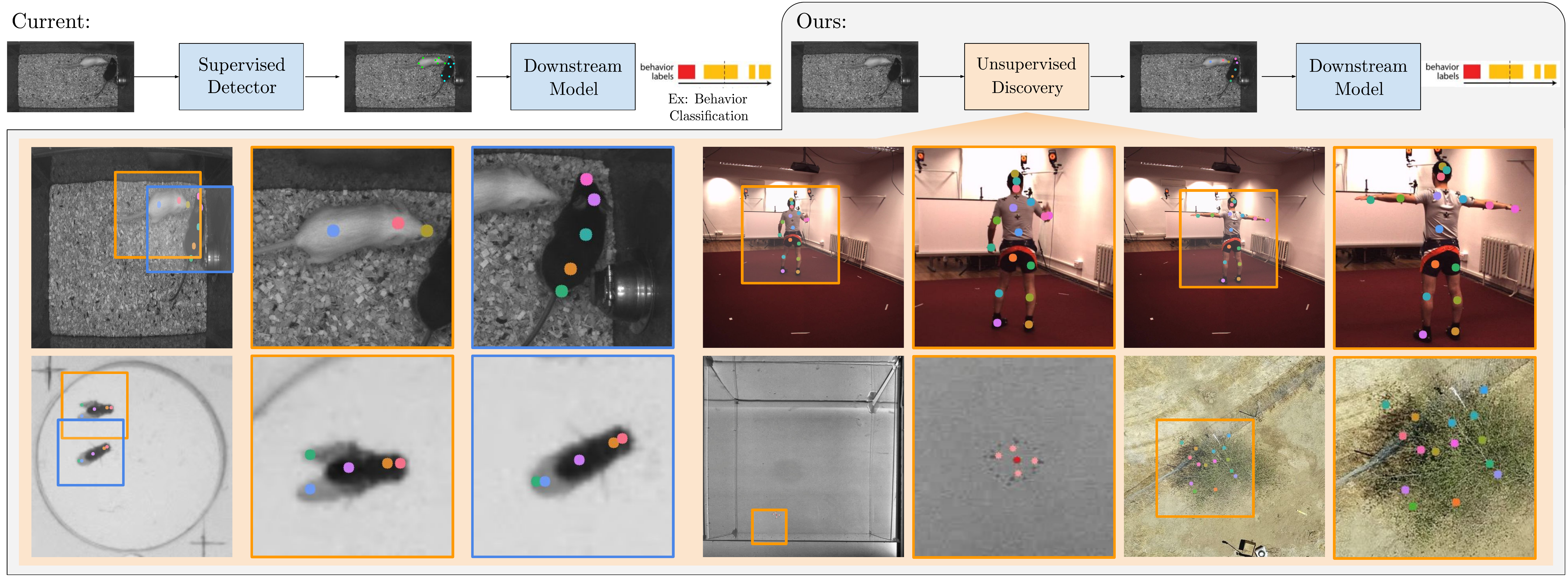}\vspace{-0.05in} 
    \captionof{figure}{\textbf{Self-supervised Behavioral Keypoint Discovery}. Intermediate representations in the form of keypoints are frequently used for behavior analysis. We propose a method to discover keypoints from behavioral videos without the need for manual keypoint or bounding box annotations. Our method works across a range of organisms (including mice, humans, flies, jellyfish and tree), works with multiple agents simultaneously (see flies and mice above), does not require bounding boxes (boxes visualized above purely for identifying the enlarged regions of interest) and achieves state-of-the-art performance on downstream tasks.}\label{fig:intro_fig}
\end{center}%
}]

{
  \renewcommand{\thefootnote}%
    {\fnsymbol{footnote}}
  \footnotetext[1]{Equal contribution. Correspondence to \url{jjsun@caltech.edu}.}
  \footnotetext[2]{Current affiliation: Samsung Advanced Institute of Technology}
}


\begin{abstract}
\vspace{-0.2cm}
We propose a method for learning the posture and structure of agents from unlabelled behavioral videos.
Starting from the observation that behaving agents are generally the main sources of movement in behavioral videos, our method, Behavioral Keypoint Discovery (B-KinD), uses an encoder-decoder architecture with a geometric bottleneck to reconstruct the spatiotemporal difference between video frames.
By focusing only on regions of movement, our approach works directly on input videos without requiring manual annotations.
Experiments on a variety of agent types (mouse, fly, human, jellyfish, and trees) demonstrate the generality of our approach and reveal that our discovered keypoints represent semantically meaningful body parts, which achieve state-of-the-art performance on keypoint regression among self-supervised methods.  
Additionally, B-KinD achieve comparable performance to supervised keypoints on downstream tasks, such as behavior classification, suggesting that our method can dramatically reduce model training costs vis-a-vis supervised methods.
\end{abstract}


\vspace{-0.2cm}
\section{Introduction}
\label{sec:intro}

Automatic recognition of object structure, for example in the form of keypoints and skeletons, enables models to capture the essence of the geometry and movements of objects.
Such structural representations are more invariant to background, lighting, and other nuisance variables and are much lower-dimensional than raw pixel values, making them good intermediates for downstream tasks, such as behavior classification~\cite{branson2009high,eyjolfsdottir2014detecting,segalin2020mouse,sun2021task,dankert2009automated}, video alignment~\cite{sun2020view,liu2021normalized}, and physics-based modeling~\cite{cardona2021wind,de2020discovery}. 

However, obtaining annotations to train supervised pose detectors can be expensive, especially for applications in behavior analysis.
For example, in behavioral neuroscience~\cite{pereira2020quantifying}, datasets are typically small and lab-specific, and the training of a custom supervised keypoint detector presents a significant bottleneck in terms of cost and effort. 
Additionally, once trained, supervised detectors often do not generalize well to new agents with different structures without new supervision. 
The goal of our work is to enable keypoint discovery on new videos without manual supervision, in order to facilitate behavior analysis on novel settings and different agents.

Recent unsupervised/self-supervised methods have made great progress in keypoint discovery~\cite{jakab20self-supervised,JakabNeurips18,ZhangKptDisc18} (see also Section~\ref{sec:previous-work}), but these methods are generally not designed for behavioral videos.
In particular, existing methods do not address the case of multiple and/or non-centered agents, and often require inputs as cropped bounding boxes around the object of interest, which would require an additional detector module to run on real-world videos.
Furthermore, these methods do not exploit relevant structural properties in behavioral videos (e.g., the camera and the background are typically stationary, as observed in many real-world behavioral datasets~\cite{segalin2020mouse,eyjolfsdottir2014detecting,burgos2012social,marstaller2019deepbees,pereira2020quantifying,jhuang2010automated}). 

To address these challenges, the key to our approach is to discover keypoints based on reconstructing the \textit{spatiotemporal difference} between video frames.
Inspired by previous works based on image reconstruction~\cite{JakabNeurips18,ryou2021weakly}, we use an encoder-decoder setup to encode input frames into a geometric bottleneck, and train the model for reconstruction. We then use spatiotemporal difference as a novel reconstruction target for keypoint discovery, instead of single image reconstruction. 
Our method enables the model to focus on discovering keypoints on the behaving agents, which are generally the only source of motion in behavioral videos.

Our self-supervised approach, \textbf{B}ehavioral \textbf{K}eypo\textbf{in}t \textbf{D}iscovery (B-KinD), works without manual supervision across diverse organisms (Figure~\ref{fig:intro_fig}). Results show that our discovered keypoints achieve state-of-the-art performance on downstream tasks among other self-supervised keypoint discovery methods. We demonstrate the performance of our keypoints on behavior classification~\cite{sun2021multi}, keypoint regression~\cite{JakabNeurips18}, and physics-based modeling~\cite{cardona2021wind}.
Thus, our method has the potential for transformative impact in behavior analysis: first, one may discover keypoints from behavioral videos for new settings and organisms; second, unlike methods that predict behavior directly from video, our low-dimensional keypoints are semantically meaningful so that users can directly compute behavioral features; finally, our method can be applied to videos without the need for manual annotations.

To summarize, our main contributions are:\\
{\bf 1. Self-supervised method for discovering keypoints} from real-world behavioral videos, based on spatiotemporal difference reconstruction. \\
{\bf 2.  Experiments across a range of organisms} (mice, flies, human, jellyfish, and tree) demonstrating the generality of the method and showing that the discovered keypoints are semantically meaningful. \\
{\bf 3. Quantitative benchmarking} on downstream behavior analysis tasks showing performance that is comparable to supervised keypoints.


\section{Related work}
\label{sec:related}
\label{sec:previous-work}

\textbf{Analyzing Behavioral Videos}. Video data collected for behavioral experiments often consists of moving agents recorded from stationary cameras~\cite{anderson2014toward,segalin2020mouse,eyjolfsdottir2014detecting,burgos2012social,nilsson2020simple,dankert2009automated,branson2009high,jhuang2010automated}. These behavioral videos contain different model organisms studied by researchers, such as fruit flies~\cite{eyjolfsdottir2014detecting,kabra2013jaaba,branson2009high,dankert2009automated} and mice~\cite{hong2015automated,segalin2020mouse,jhuang2010automated,burgos2012social}. From these recorded video data, there has been an increasing effort to automatically estimate poses of agents and classify behavior~\cite{kabra2013jaaba,hong2015automated,eyjolfsdottir2016learning,Mathisetal2018,egnor2016computational,segalin2020mouse}. 

Pose estimation models that were developed for behavioral videos~\cite{Mathisetal2018,graving2019deepposekit,segalin2020mouse,pereira2020sleap} require human annotations of anatomically defined keypoints, which are expensive and time-consuming to obtain. In addition to the cost, not all data can be crowd-sourced due to the sensitive nature of some experiments. Furthermore, organisms that are translucent (jellyfish) or with complex shapes (tree) can be difficult for non-expert humans to annotate. Our goal is to enable keypoint discovery on videos for behavior analysis, without the need for manual annotations.

After pose estimation, behavior analysis models generally compute trajectory features and train behavior classifiers in a fully supervised fashion~\cite{burgos2012social,hong2015automated,eyjolfsdottir2014detecting,sun2021task,segalin2020mouse}. Some works have also explored using unsupervised methods to discover new motifs and behaviors~\cite{berman2014mapping,wiltschko2015mapping,hsu2021b,luxem2020identifying}. Here, we apply our discovered keypoints to supervised behavior classification and compare against baseline models using supervised keypoints for this task.

\textbf{Keypoint Estimation}.
Keypoint estimation models aim to localize a predefined set of keypoints from visual data, and many works in this area focus on human pose. 
With the success of fully convolutional neural networks~\cite{FCN_segmentation17}, recent methods~\cite{Newell2016StackedHN,Wei_CVPR16,CPN17,Tang_2018_ECCV} employ encoder-decoder networks by predicting high-resolution outputs encoded with 2D Gaussian heatmaps representing each part. To improve model performance,~\cite{Newell2016StackedHN,Wei_CVPR16,Tang_2018_ECCV} propose an iterative refinement approach,~\cite{CPN17,Ryou_2019_ICCV} design efficient learning signals, and~\cite{Cheng_2020_CVPR,Wang2021DeepHR} exploit multi-resolution information. Beyond human pose, there are also works that focus on animal pose estimation, notably~\cite{Mathisetal2018,graving2019deepposekit,pereira2020sleap}. Similar to these works, we also use 2D Gaussian heatmaps to represent parts as keypoints, but instead of using human-defined keypoints, we aim to discover keypoints from video data without manual supervision. 

\begin{figure*}
    \centering
    \includegraphics[width=\linewidth]{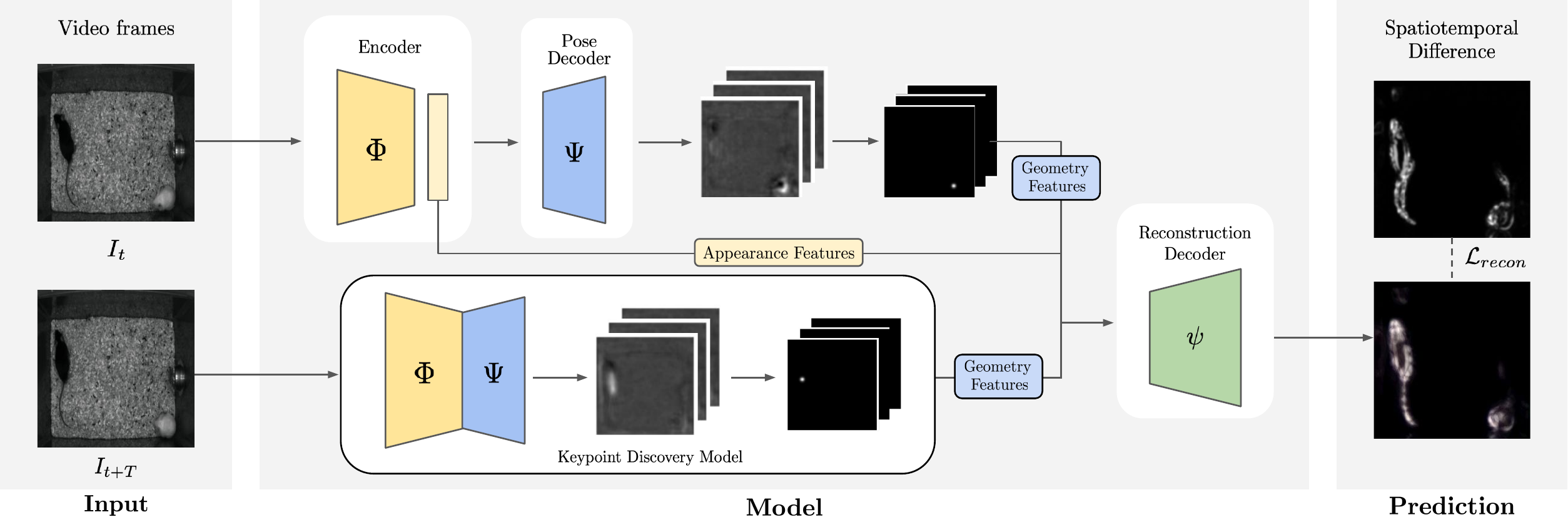}\vspace{-0.05in}
    \caption{\textbf{B-KinD, an approach for keypoint discovery from spatiotemporal difference reconstruction}. $I_t$ and $I_{t+T}$ are video frames at time $t$ and $t+T$. Both frame $I_t$ and frame $I_{t+T}$ are fed to an appearance encoder $\Phi$ and a pose decoder $\Psi$. Given the appearance feature from $I_t$ and geometry features from both $I_t$ and $I_{t+T}$ (Sec~\ref{sec:method_disc}), our model reconstructs the spatiotemporal difference (Sec~\ref{sec:img_diff}) computed from two frames using the reconstruction decoder $\psi$. }
    \vspace{-0.1in}
    \label{fig:pipeline}
\end{figure*}

\textbf{Unsupervised Part Discovery}. 
Though keypoints provide a useful tool for behavior analysis, collecting annotations is time-consuming and labor-intensive especially for new domains that have not been previously studied. Unsupervised keypoint discovery~\cite{JakabNeurips18,ZhangKptDisc18,jakab20self-supervised} has been proposed to reduce keypoint annotation effort and there have been many promising results on centered and/or aligned objects, such as facial images and humans with an upright pose. 
These methods train and evaluate on images where the object of interest is centered in an input bounding box. 
Most of the approaches~\cite{ZhangKptDisc18,JakabNeurips18,Lorenz19} use an autoencoder-based architecture to disentangle the appearance and geometry representation for the image reconstruction task. Our setup is similar in that we also use an encoder-decoder architecture, but crucially, we reconstruct spatiotemporal difference between video frames, instead of the full image as in previous works. We found that this enables our discovered keypoints to track semantically-consistent parts without manual supervision, requiring neither keypoints nor bounding boxes.

There are also works for parts discovery that employ other types of supervision~\cite{jakab20self-supervised,SchmidtkeVELAK21,ryou2021weakly}.
For example, \cite{ryou2021weakly} proposed a weakly-supervised approach using class label to discriminate parts to handle viewpoint changes, \cite{jakab20self-supervised} incorporated pose prior obtained from unpaired data from different datasets in the same domain, and \cite{SchmidtkeVELAK21} proposed a template-based geometry bottleneck based on a pre-defined 2D Gaussian-shaped template. 
Different from these approaches, our method does not require any supervision beyond the behavioral videos.
We chose to focus on this setting since other supervisory sources are not readily available for emerging domains (ex: jellyfish, trees).

In previous works, keypoint discovery has been applied to downstream tasks, such as image and video generation~\cite{minderer2019unsupervised,jakab20self-supervised}, keypoint regression to human-annotated poses~\cite{ZhangKptDisc18,JakabNeurips18}, and video-level action recognition~\cite{kim2019unsupervised,minderer2019unsupervised}. While we also apply keypoint discovery to downstream tasks, we note that our work differs in approach (we discover keypoints directly on behavioral videos using spatiotemporal difference reconstruction), focus (behavioral videos of diverse organisms), and application (real-world behavior analysis tasks~\cite{sun2021multi,cardona2021wind}).


\section{Method}
\label{sec:method}

The goal of B-KinD (Figure~\ref{fig:pipeline}) is to discover semantically meaningful keypoints in behavioral videos of diverse organisms without manual supervision. 
We use an encoder-decoder setup similar to previous methods~\cite{JakabNeurips18,ryou2021weakly}, but instead of image reconstruction, here we study a novel reconstruction target based on spatiotemporal difference.
In behavioral videos, the camera is generally fixed with respect to the world, such that the background is largely stationary and the agents (\eg mice moving in an enclosure) are the only source of motion. Thus spatiotemporal differences provide a strong cue to infer location and movements of agents. 

\subsection{Self-supervised keypoint discovery}
\label{sec:method_disc}
Given a behavioral video, our work aims to reconstruct regions of motion between a reference frame $I_t$ (the video frame at time $t$) and a future frame $I_{t+T}$ (the video frame $T$ timesteps later, for some set value of $T$.) We accomplish this by extracting appearance features from frame $I_t$ and keypoint locations ("geometry features") from both frames $I_t$ and $I_{t+T}$ (Figure~\ref{fig:pipeline}). 
In contrast, previous works~\cite{JakabNeurips18,Lorenz19,jakab20self-supervised,ryou2021weakly,SchmidtkeVELAK21} use appearance features from $I_t$ and geometry features from $I_{t+T}$ to reconstruct the full image $I_{t+T}$ (instead of difference between $I_{t}$ and $I_{t+T}$).

We use an encoder-decoder architecture, with shared appearance encoder $\Phi$, geometry decoder $\Psi$, and reconstruction decoder $\psi$. During training, the pair of frames $I_t$ and $I_{t+T}$ are fed to the appearance encoder $\Phi$ to generate appearance features, and those features are then fed into the geometry decoder $\Psi$ to generate geometry features. In our approach, the reference frame $I_t$ is used to generate both appearance and geometry representations, and the future frame $I_{t+T}$ is only used to generate a geometry representation. The appearance feature $h_a^t$ for frame $I_t$ are defined simply as the output of $\Phi$: $h_a^t = \Phi(I_t)$.

The pose decoder $\Psi$ outputs $K$ raw heatmaps $\textbf{X}_i \in \mathbb{R}^2$, then applies a spatial softmax operation on each heatmap channel. 
Given the extracted $p_i=(u_i, v_i)$ locations for $i=\{1,\dots,K\}$ keypoints from the spatial softmax, we define the geometry features $h_g^t$ to be a concatenation of 2D Gaussians centered at $(u_i,v_i)$ with variance $\sigma$.

Finally, the concatenation of the appearance feature $h_a^t$ and the geometry features $h_g^t$ and $h_g^{t+T}$ is fed to the decoder $\psi$ to reconstruct the learning objective $\hat{S}$ discussed in the next section: $\hat{S} = \psi(h_a^t, h_g^t, h_g^{t+T})$.

\subsection{Learning formulation}
\subsubsection{Spatiotemporal difference}
\label{sec:img_diff}

Our method works with different types of spatiotemporal differences as reconstruction targets. For example:

\textbf{Structural Similarity Index Measure} (SSIM)~\cite{Wang04imagequality}. This is a method for measuring the perceived quality of the two images based on luminance, contrast, and structure features. To compute our reconstruction target based on SSIM, we apply the SSIM measure locally on corresponding patches between $I_{t}$ and $I_{t+T}$ to build a similarity map between frames. Then we compute dissimilarity by taking the negation of the similarity map.

\textbf{Frame differences}.
When the video background is static with little noise, simple frame differences, such as absolute difference ($S_{\left|d\right|} = \left|I_{t+T} - I_t\right|$) or raw difference ($S_{d} = I_{t+T} - I_t$), can also be directly applied as a reconstruction target.

\subsubsection{Reconstruction loss}
We apply perceptual loss~\cite{Johnson2016Perceptual} for reconstructing the spatiotemporal difference $S$. Perceptual loss compares the L2 distance between the features computed from VGG network $\phi$~\cite{VGG14}. The reconstruction $\hat{S}$ and the target $S$ are fed to VGG network, and mean squared error is applied to the features from the intermediate convolutional blocks:
\begin{align}
    \mathcal{L}_{recon} = \left\Vert \phi(S(I_t,I_{t+T})) - \phi(\hat{S}(I_t,I_{t+T})) \right\Vert_2.
\end{align}

\subsubsection{Rotation equivariance loss}
In cases where agents can move in many directions (\eg mice filmed from above can translate and rotate freely), we would like our keypoints to remain semantically consistent. We enforce rotation-equivariance in the discovered keypoints by rotating the image with different angles and imposing that the predicted keypoints should move correspondingly. We apply the rotation equivariance loss (similar to the deformation equivariance in~\cite{Thewlis_2017_ICCV}) on the generated heatmap.

Given reference image $I$ and the corresponding geometry bottleneck $h_g$, we rotate the geometry bottleneck to generate pseudo labels $h_g^{R\degree}$ for rotated input images $I^{R\degree}$ with degree $R=\{90\degree, 180\degree, 270\degree\}$. We apply mean squared error between the predicted geometry bottlenecks $\hat{h_g}$ from the rotated images and the generated pseudo labels $h_g$:
\begin{align}
    \mathcal{L}_{r} = \left\Vert h_g^{R\degree} - \hat{h}_g(I^{R\degree}) \right\Vert_2.
\end{align}

\subsubsection{Separation loss}
Empirical results show that rotation equivariance encourages the discovered keypoints to converge at the center of the image. We apply separation loss to encourage the keypoints to encode unique coordinates, and prevent the discovered keypoints from being centered at the image coordinates~\cite{ZhangKptDisc18}. The separation loss is defined as follows: 
\begin{align}
    \mathcal{L}_{s} = \sum_{i \neq j} \exp{\left( \frac{-(p_i - p_j)^2}{2\sigma_s^2} \right)}.
\end{align}

\subsubsection{Final objective}\label{sec:final_objective}
Our final loss function is composed of three parts: reconstruction loss $\mathcal{L}_{recon}$, rotation equivariance loss $\mathcal{L}_r$, and separation loss $\mathcal{L}_s$:
\begin{align}
    \mathcal{L} = \mathcal{L}_{recon} + \mathbbm{1}_{epoch>n} (w_r \mathcal{L}_r + w_s\mathcal{L}_s).
\end{align}
We adopt curriculum learning~\cite{Bengio2009} and apply $\mathcal{L}_r$ and $\mathcal{L}_s$ once the keypoints are consistently discovered from the semantic parts of the target instance.

\begin{figure}
    \centering
    \includegraphics[width=0.95\linewidth]{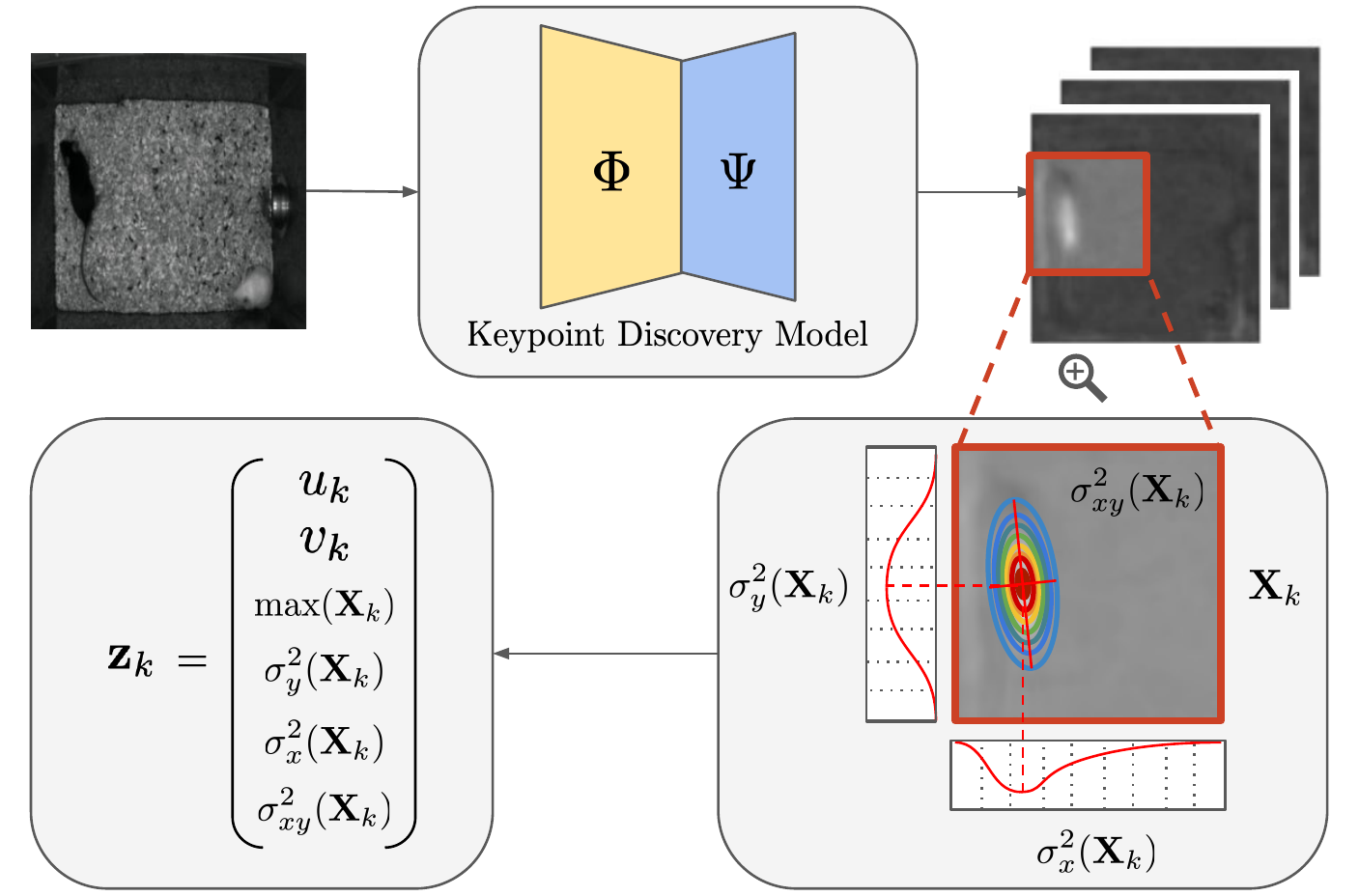}
    \caption{\textbf{Behavior Classification Features}. Extracting information from the raw heatmap (Section~\ref{sec:behavior_quant}): the confidence scores and the covariance matrices are computed from normalized heatmaps. Note that the features are computed for all $x, y$ coordinates. We visualize the zoomed area around the target instance for illustrative purposes.}\vspace{-0.5cm}
    \label{fig:behavior_input}
\end{figure}

\subsection{Feature extraction for behavior analysis}~\label{sec:behavior_quant}
\vspace{-0.3cm}

Following standard approaches~\cite{segalin2020mouse,burgos2012social,hong2015automated}, we use the discovered keypoints from B-KinD as input to a behavior quantification module: either supervised behavior classifiers or a physics-based model. Note that this is a separate process from keypoint discovery; we feed discovered geometry information into a downstream model.

In addition to discovered keypoints, we extracted additional features from the raw heatmap (Figure~\ref{fig:behavior_input}) to be used as input to our downstream modules. 
For instance, we found that the confidence and the shape information from the of the network prediction of keypoint location was informative.
When a target part is well localized, our keypoint discovery network produces a heatmap with a single high peak with low variance; conversely, when a target part is occluded, the raw heatmap contains a blurred shape with lower peak value. This ``confidence" score (heatmap peak value) is also a good indicator for whether keypoints are discovered on the background (blurred over the background with low confidence) or tracking anatomical body parts (peaked with high confidence), visualized in Supplementary materials. The shape of a computed heatmap can also reflect shape information of the target (\eg stretching).

Given a raw heatmap $\textbf{X}_k$ for part $k$, the confidence score is obtained by choosing the maximum value from the heatmap, and the shape information is obtained by computing the covariance matrix from the heatmap. Figure~\ref{fig:behavior_input} visualizes the features we extract from the raw heatmaps. Using the normalized heatmap as the probability distribution, additional geometric features are computed:
\begin{align}
    \sigma^2_x (\textbf{X}_k) &= \sum_{ij} (x_i - u_k)^2 \textbf{X}_k(i,j), \nonumber\\
    \sigma^2_y (\textbf{X}_k) &= \sum_{ij} (y_j - v_k)^2 \textbf{X}_k(i,j), \\
    \sigma^2_{xy} (\textbf{X}_k) &= \sum_{ij} (x_i - u_k) (y_j - v_k) \textbf{X}_k(i,j). \nonumber
\end{align}


\section{Experiments}

We demonstrate that B-KinD is able to discover consistent keypoints in real-world behavioral videos across a range of organisms (Section~\ref{sec:datasets}). We evaluate our keypoints on downstream tasks for behavior classification (Section~\ref{sec:behavior_res}) and pose regression (Section~\ref{sec:pose_res}), then illustrate additional applications of our keypoints (Section~\ref{sec:qualitative_res}). 

\subsection{Experimental setting}

\subsubsection{Datasets}
\label{sec:datasets}

\textbf{CalMS21}. CalMS21~\cite{sun2021multi} is a large-scale dataset for behavior analysis consisting of videos and trajectory data from a pair of interacting mice. Every frame is annotated by an expert for three behaviors: sniff, attack, mount. There are 507k frames in the train split, and 262k frames in the test split (video frame: $1024\times570$, mouse: approx $150\times50$). We use only the train split on videos without miniscope cable to train B-KinD. Following~\cite{sun2021multi}, the downstream behavior classifier is trained on the entire training split, and performance is evaluated on the test split.

\textbf{MARS-Pose}. This dataset consists of a set of videos with similar recording conditions to the CalMS21 dataset. We use a subset of the MARS pose dataset~\cite{segalin2020mouse} with keypoints from manual annotations to evaluate the ability of our model to predict human-annotated keypoints, with $\{10,50,100,500\}$ images for train and 1.5k images for test.

\textbf{Fly vs. Fly}. These videos consists of interactions between a pair of flies, annotated per frame by domain experts. We use the Aggression videos from the Fly vs. Fly dataset~\cite{eyjolfsdottir2014detecting}, with the train and test split having 1229k and 322k frames respectively (video frame: $144 \times 144$, fly: approx $30 \times 10$). Similar to~\cite{sun2021task}, we evaluate on behaviors of interest with more than 1000 frames in the training set (lunge, wing threat, tussle).

\textbf{Human 3.6M}. Human 3.6M~\cite{ionescu2013human3} is a large-scale motion capture dataset, which consists of 3.6 million human poses and images from 4 viewpoints. To quantitatively measure the pose regression performance against baselines, we use the Simplified Human 3.6M dataset, which consists of 800k training and 90k testing images with 6 activities in which the human body is mostly upright. We follow the same evaluation protocol from~\cite{ZhangKptDisc18} to use subjects 1, 5, 6, 7, and 8 for training and 9 and 11 for testing. We note that each subject has different appearance and clothing.

\textbf{Jellyfish}. The jellyfish data is an in-house video dataset containing 30k frames of recorded swimming jellyfish (video frame: $928 \times 1158$, jellyfish: approx $50$ pix in diameter). We use this dataset to qualitatively test the performance of B-KinD on a new organism, and apply our keypoints to detect the pulsing motion of the jellyfish.

\textbf{Vegetation}. This is an in-house dataset acquired over several weeks using a drone to record the motion of swaying trees. The dataset consists of videos of an oak tree and corresponding wind speeds recorded using an anemometer, with a total of 2.41M video frames (video frame: $512 \times 512$, oak tree: varies, approx $\frac{1}{4}$ of the frame). We evaluate this dataset using a physics-based model~\cite{cardona2021wind} that relates the visually observed oscillations to the average wind speeds.

\vspace{-0.2cm}
\subsubsection{Training and evaluation procedure}

We train B-KinD using the full objective in Section~\ref{sec:final_objective}. During training, we rescale images to $256 \times 256$ and use $T$ of around $0.2$ seconds, except Human3.6M, where we use $128 \times 128$. Unless otherwise specified, all experiments are ran with all keypoints discovered from B-KinD with SSIM reconstruction and with 10 keypoints for mouse, fly, and jellyfish, 16 keypoints for Human3.6M, 15 keypoints for Vegetation. We train on the train split of each dataset as specified, except for jellyfish and vegetation, where we use the entire dataset. Additional details are in the Supplementary materials.

After training the keypoint discovery model, we extract the keypoints and use it for different evaluations based on the labels available in the dataset: behavior classification (CalMS21, Fly), keypoint regression (MARS-Pose, Human), and physics-based modeling (Vegetation). 

For keypoint regression, similar to previous works~\cite{JakabNeurips18,jakab20self-supervised}, we compare our regression with a fully supervised 1-stack hourglass network~\cite{Newell2016StackedHN}. We evaluate keypoint regression on Simplified Human 3.6M by using a linear regressor without a bias term, following the same evaluation setup from previous works~\cite{ZhangKptDisc18,Lorenz19}. On MARS-Pose, we train our model in a semi-supervised fashion with $10,50,100,500$ supervised keypoints to test data efficiency.
For behavior classification, we evaluate on CalMS21 and Fly, using available frame-level behavior annotations. To train behavior classifiers, we use the specified train split of each dataset. For CalMS21 and Fly, we train the 1D Convolutional Network benchmark model provided by~\cite{sun2021multi} using B-KinD keypoints. We evaluate using mean average precision (MAP) weighted equally over all behaviors of interest.

\subsection{Behavior classification results}~\label{sec:behavior_res}
\vspace{-0.4cm}

\begin{figure}
    \centering
    \includegraphics[width=\linewidth]{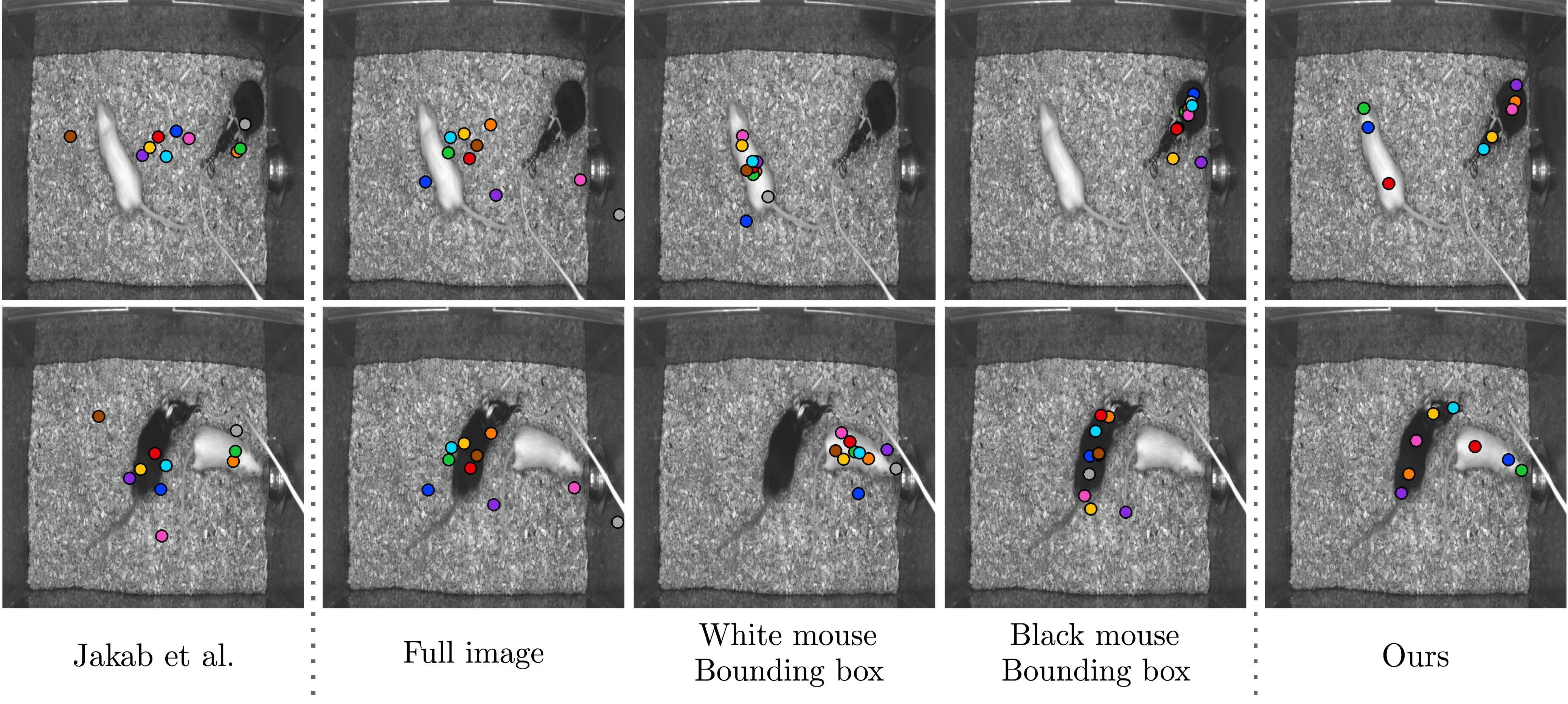}\vspace{-0.3cm}
    \caption{{\bf Comparison with existing methods~\cite{JakabNeurips18}}, full image, bounding box, and SSIM reconstruction (ours). ``Jakab~\etal" and ``full image" results are based on full image reconstruction. ``White mouse bounding box" and ``black mouse bounding box" show the results when the cropped bounding boxes were fed to the network for image reconstruction.}
    \label{fig:comparison}\vspace{-0.35cm}
\end{figure}

\textbf{CalMS21 Behavior Classification}.
We evaluate the effectiveness of B-KinD for behavior classification (Table~\ref{tab:mouse_behavior}). Compared to supervised keypoints trained for this task, our keypoints (without manual supervision) is comparable when using both pose and confidence as input. Compared to other self-supervised methods, even those that use bounding boxes, our discovered keypoints on the full image generally achieve better performance.

Keypoints discovered with image reconstruction, similar to baselines~\cite{JakabNeurips18,ryou2021weakly} cannot track the agents well without using bounding box information (Figure~\ref{fig:comparison}) and does not perform well for behavior classification (Table~\ref{tab:mouse_behavior}). When we provide bounding box information to the model based on image reconstruction, the performance is significantly improved, but this model does not perform as well as B-KinD keypoints from spatiotemporal difference reconstruction.

For the per-class performance (see the Supplementary materials), the biggest gap exists between B-KinD and MARS on the ``attack" behavior. This is likely because during attack, the mice are moving quickly, and there exists a lot of motion blur and occlusion which is difficult to track without supervision. However, once we extract more information from the heatmap, through computing keypoint confidence, our keypoints perform comparably to MARS.

\begin{table}
  \begin{center}
\scalebox{0.9}{
    \begin{tabular}{lcccc}
        \toprule[0.2em]
        CalMS21 & Pose & Conf & Cov & MAP\\
        \toprule[0.2em]
        \multicolumn{5}{c}{\textbf{\textit{Fully supervised}}} \\
        \multirow{3}{*}{MARS $\dagger$ ~\cite{segalin2020mouse}} & \checkmark & & & $.856 \pm .010$ \\
        & \checkmark & \checkmark & & $.874 \pm .003$ \\
        & \checkmark & \checkmark & \checkmark & $.880 \pm .005$ \\
        \bottomrule[0.1em]
        \multicolumn{5}{c}{\textbf{\textit{Self-supervised}}} \\   
        Jakab et al.~\cite{JakabNeurips18} & \checkmark & &  & $.186 \pm .008$\\
        \hline
        \multirow{3}{*}{Image Recon.} & \checkmark & & & $.182 \pm .007$ \\
        & \checkmark & \checkmark & & $.184 \pm .006$ \\
        & \checkmark & \checkmark & \checkmark & $.165 \pm .012$ \\
        \hline
        \multirow{3}{*}{Image Recon. bbox$\dagger$} & \checkmark & & & $.819 \pm .008$ \\
        & \checkmark & \checkmark & & $.812 \pm .006$ \\
        & \checkmark & \checkmark & \checkmark & $.812 \pm .010$ \\ \hline
        \multirow{3}{*}{Ours} & \checkmark & & & $.814 \pm .007$ \\
        & \checkmark & \checkmark & & $.857 \pm .005$ \\
        & \checkmark & \checkmark & \checkmark & $.852 \pm .013$ \\   
        \bottomrule[0.1em]
    \end{tabular}}
  \caption{\textbf{Behavior Classification Results on CalMS21}. ``Ours" represents classifiers using input keypoints from our discovered keypoints. ``conf" represents using the confidence score, and ``cov" represents values from the covariance matrix of the heatmap. $\dagger$ refers to models that require bounding box inputs before keypoint estimation. Mean and std dev from 5 classifier runs are shown. }
  \label{tab:mouse_behavior}
  \end{center}\vspace{-0.4cm}
\end{table}

\begin{table}
  \begin{center}
  \vspace{-0.05in}
\scalebox{0.9}{
    \begin{tabular}{lc}
        \toprule[0.2em]
        Fly &  MAP\\
        \toprule[0.2em]
        \multicolumn{2}{c}{\textbf{\textit{Hand-crafted features}}} \\
        FlyTracker~\cite{eyjolfsdottir2014detecting} & $.809 \pm .013$ \\
        \bottomrule[0.1em]
        \multicolumn{2}{c}{\textbf{\textit{Self-supervised + generic features}}} \\    
        Image Recon. & $.500 \pm .024$  \\
        Image Recon. bbox$\dagger$ & $.750 \pm .020$ \\
        Ours & $.727 \pm .022$ \\
        \bottomrule[0.1em]
    \end{tabular}}
  \vspace{-0.05in}    
  \caption{\textbf{Behavior Classification Results on Fly}. ``FlyTracker" represents classifiers using  hand-crafted inputs from~\cite{eyjolfsdottir2014detecting}. The self-supervised keypoints all use the same ``generic features" computed on all keypoints: speed, acceleration, distance, and angle. $\dagger$ refers to models that require bounding box inputs before keypoint estimation. Mean and std dev from 5 classifier runs are shown. }\vspace{-0.6cm}
  \label{tab:fly_behavior}
  \end{center}
\end{table}

\textbf{Fly Behavior Classification}.
The FlyTracker~\cite{eyjolfsdottir2014detecting} uses hand-crafted features computed from the image, such as contrast, as well as features from tracked fly body parts, such as wing angle or distance between flies. 
Using discovered keypoints, we compute comparable features without assuming keypoint identity, by computing speed and acceleration of every keypoint, distance between every pair, and angle between every triplet. For all self-supervised methods, we use keypoints, confidence, and covariance for behavior classification. Results demonstrate that while there is a small gap in performance to the supervised estimator, our discovered keypoints perform much better than image reconstruction, and is comparable to models that require bounding box inputs (Table~\ref{tab:fly_behavior}).

\vspace{-0.1cm}
\subsection{Pose regression results}~\label{sec:pose_res}
\vspace{-0.4cm}

\begin{figure}
    \centering
    \includegraphics[width=0.9\linewidth]{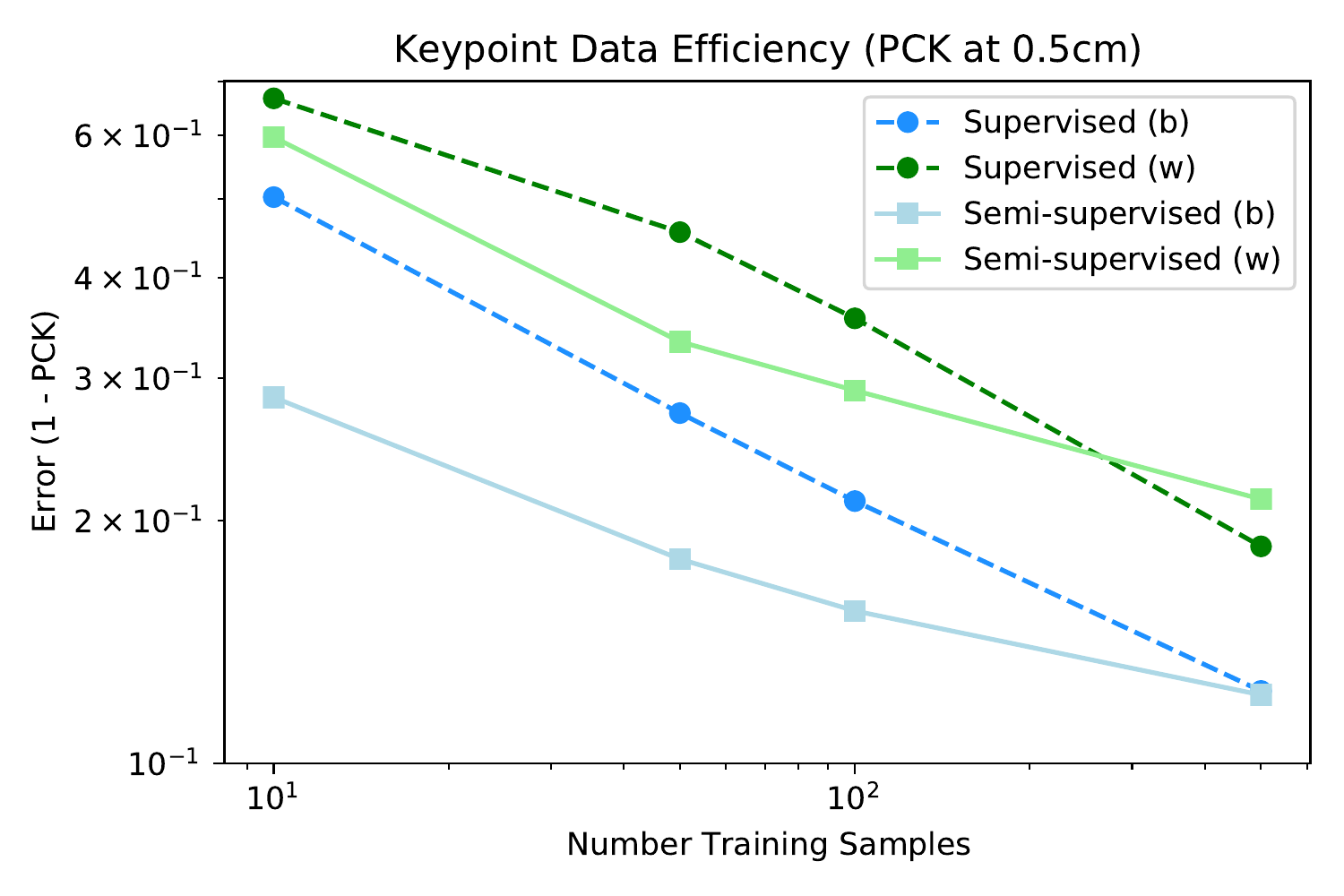}\vspace{-0.3cm}
    \caption{\textbf{Keypoint data efficiency on MARS-Pose}. The supervised model is based on~\cite{segalin2020mouse} using stacked hourglass~\cite{Newell2016StackedHN}, while the semi-supervised model uses both our self-supervised loss and supervision. PCK is computed at $0.5cm$ threshold, averaged across nose, ears, and tail keypoints, over 3 runs. ``b" and ``w" indicates the black and white mouse respectively. }\vspace{-0.4cm}
    \label{fig:mouse_kp}
\end{figure}

\textbf{MARS Pose Regression}.
We evaluate the pose estimation performance of our method in the setting where some human annotated keypoints exist (Figure~\ref{fig:mouse_kp}). For this experiment, we train B-KinD in a semi-supervised fashion, where the loss is a sum of both our keypoint discovery objective (Section~\ref{sec:final_objective}) as well as standard keypoint estimation objectives based on MSE~\cite{segalin2020mouse}.
For both black and white mouse, when using our keypoint discovery objective in a semi-supervised way during training, we are able to track keypoints more accurately compared to the supervised method~\cite{segalin2020mouse} alone. We note that the performance of both methods converge at around 500 annotated examples.

\textbf{Simplified Human 3.6M Pose Regression}.
To compare with existing keypoint discovery methods, we evaluate our discovered keypoints on Simplified Human3.6M (a standard benchmarking dataset) by regressing to annotated keypoints (Table~\ref{tab:simple_h36m_regression}). Though our method is directly applicable to full images, we train the discovery model using cropped bounding box for a fair comparison with baselines, which all use cropped bounding boxes centered on the subject. 
Compared to both self-supervised + prior information and self-supervised + regression, our method shows state-of-the-art performance on the keypoint regression task, suggesting spatiotemporal difference is an effective reconstruction target for keypoint discovery.

\begin{table}
  \begin{center}
\scalebox{0.75}{
    \begin{tabular}{lccccccc}
        \toprule[0.2em]
        Simplified H36M & all & wait & pose & greet & direct & discuss & walk \\
        \toprule[0.2em]
        \multicolumn{8}{c}{\textbf{\textit{Fully supervised:}}} \\
        Newell~\cite{Newell2016StackedHN} & 2.16 & 1.88 & 1.92 & 2.15 & 1.62 & 1.88 & 2.21 \\
        \bottomrule[0.1em]
        \multicolumn{8}{c}{\textbf{\textit{Self-supervised + unpaired labels}}} \\    
        Jakab~\cite{jakab20self-supervised}$\ddagger$ & 2.73 & 2.66 & 2.27 & 2.73 & 2.35 & 2.35 & 4.00 \\
        \bottomrule[0.1em]
        \multicolumn{8}{c}{\textbf{\textit{Self-supervised + template}}} \\    
        Schmidtke~\cite{SchmidtkeVELAK21} & 3.31 & 3.51 & 3.28 & 3.50 & 3.03 & 2.97 & 3.55 \\
        \bottomrule[0.1em]     
        \multicolumn{8}{c}{\textbf{\textit{Self-supervised + regression}}} \\
        Thewlis~\cite{Thewlis_2017_ICCV} & 7.51 & 7.54 & 8.56 & 7.26 & 6.47 & 7.93 & 5.40 \\
        Zhang~\cite{ZhangKptDisc18} & 4.14 & 5.01 & 4.61 & 4.76 & 4.45 & 4.91 & 4.61 \\
        Lorenz~\cite{Lorenz19} & 2.79 & – & – & – & – & – & – \\
        \midrule[0.1em]
        Ours (best) & 2.44 & 2.50 & 2.22 & 2.47 & 2.22 & 2.77 & 2.50 \\
        Ours (mean) & 2.53 & 2.58 & 2.31 & 2.56 & 2.34 & 2.83 & 2.58 \\
        Ours (std) & .056 & .047 & .062 & .048 & .066 & .048 & .063 \\
        \bottomrule[0.1em]
    \end{tabular}}
  \caption{\textbf{Comparison with state-of-the-art methods for landmark prediction on Simplified Human 3.6M}. The error is in \%-MSE normalized by image size. All methods predict 16 keypoints except for~\cite{jakab20self-supervised}$\ddagger$, which uses 32 keypoints for training a prior model from the Human 3.6M dataset. B-Kind results are computed from 5 runs.} 
  \label{tab:simple_h36m_regression}
  \end{center}\vspace{-0.5cm}
\end{table}

\textbf{Learning Objective Ablation Study} We report the pose regression performance on Simplified Human3.6M (Table~\ref{tab:ablation_target_h36m}) by varying the spatiotemporal difference reconstruction target for training B-KinD. Here, image reconstruction also performs well since cropped bounding box is used as an input to the network. Overall, spatiotemporal difference reconstruction yield better performance over image reconstruction, and performance can be further improved by extracting additional confidence and covariance information from the discovered heatmaps.

\begin{table}
  \begin{center}
  \scalebox{0.8}{
    \begin{tabular}{lc}
        \toprule[0.2em]
        Learning Objective &  \%-MSE \\
        \toprule[0.2em]
        Image Recon. & 2.918 $\pm$ 0.139 \\
        Abs. Difference & 2.642 $\pm$ 0.174 \\
        Difference & 2.770 $\pm$ 0.158 \\
        SSIM & \textbf{2.534 $\pm$ 0.056}  \\
        \bottomrule[0.1em]
        \multicolumn{2}{c}{\textbf{\textit{Self-supervised + extracted features}}} \\
        SSIM & \textbf{2.494 $\pm$ 0.047} \\
        \bottomrule[0.1em]
    \end{tabular}}
  \caption{\textbf{Learning objective ablation on Simplified Human3.6M}. \%-MSE error is reported by changing the reconstruction target. Extracted features correspond to keypoint locations, confidence, and covariance. Results are from 5 B-KinD runs.}
  \label{tab:ablation_target_h36m}
  \end{center}\vspace{-0.6cm}
\end{table}


\def\figsize{0.15}
\def\fighspace{-2mm}
\def\fighspacer{-2mm}
\begin{figure*}
\centering
\begin{tabular}{cccccc}
\centering
\hspace{\fighspace}\includegraphics[width=\figsize\textwidth,height=\figsize\textwidth]{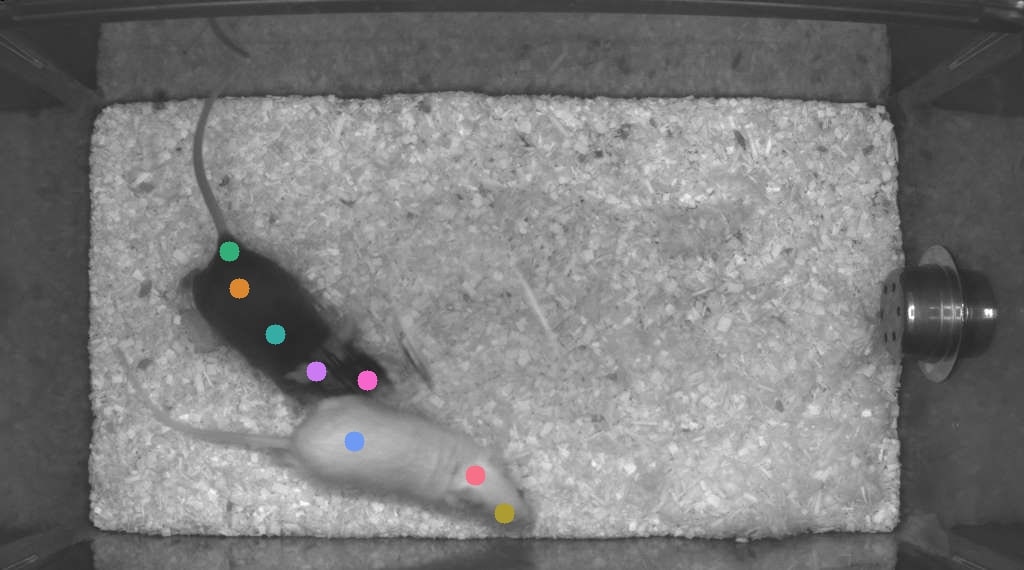}\hspace{\fighspace} & \includegraphics[width=\figsize\textwidth,height=\figsize\textwidth]{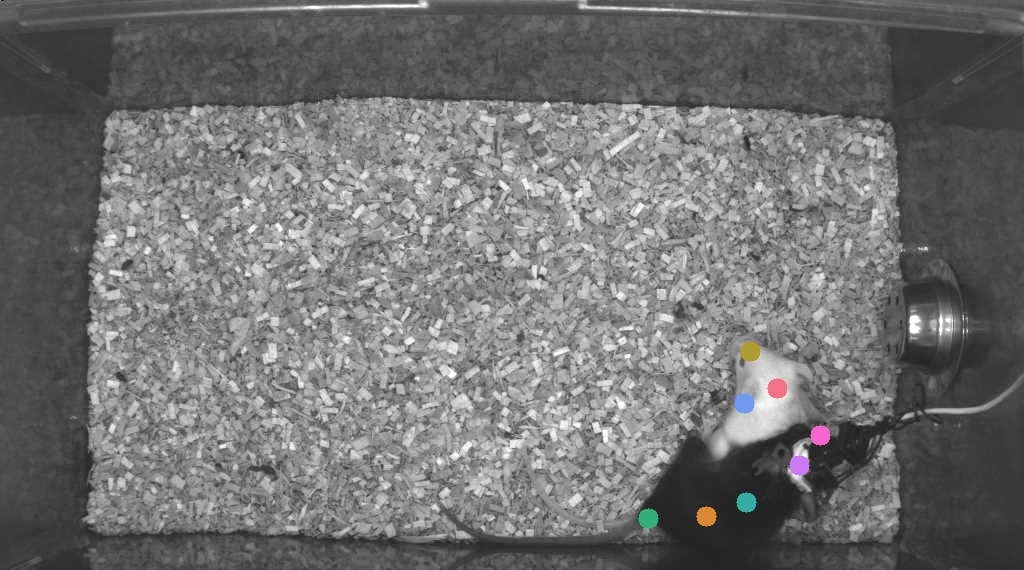}\hspace{\fighspacer} & \includegraphics[width=\figsize\textwidth,height=\figsize\textwidth]{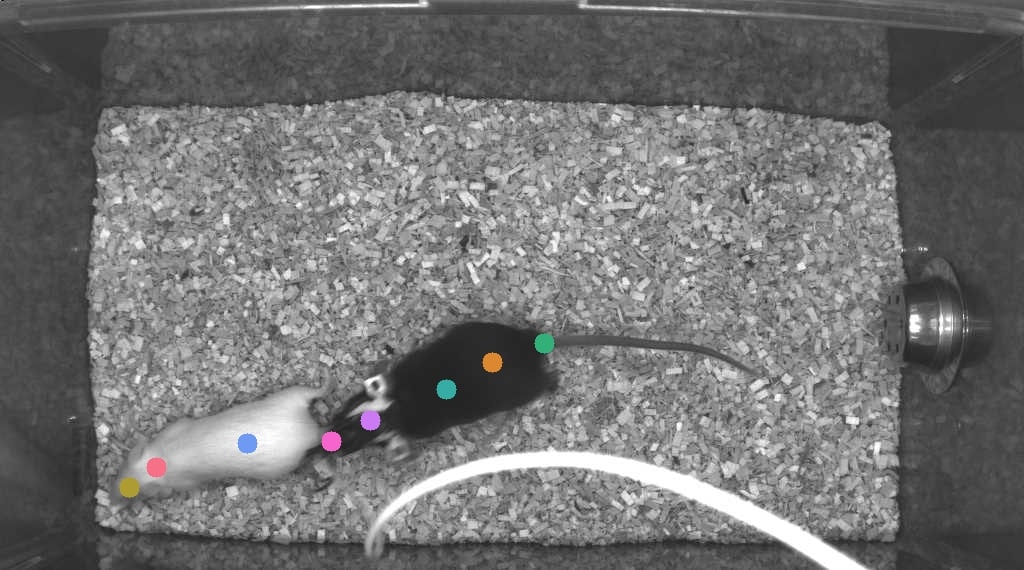}\hspace{\fighspace} & 
\includegraphics[width=\figsize\textwidth]{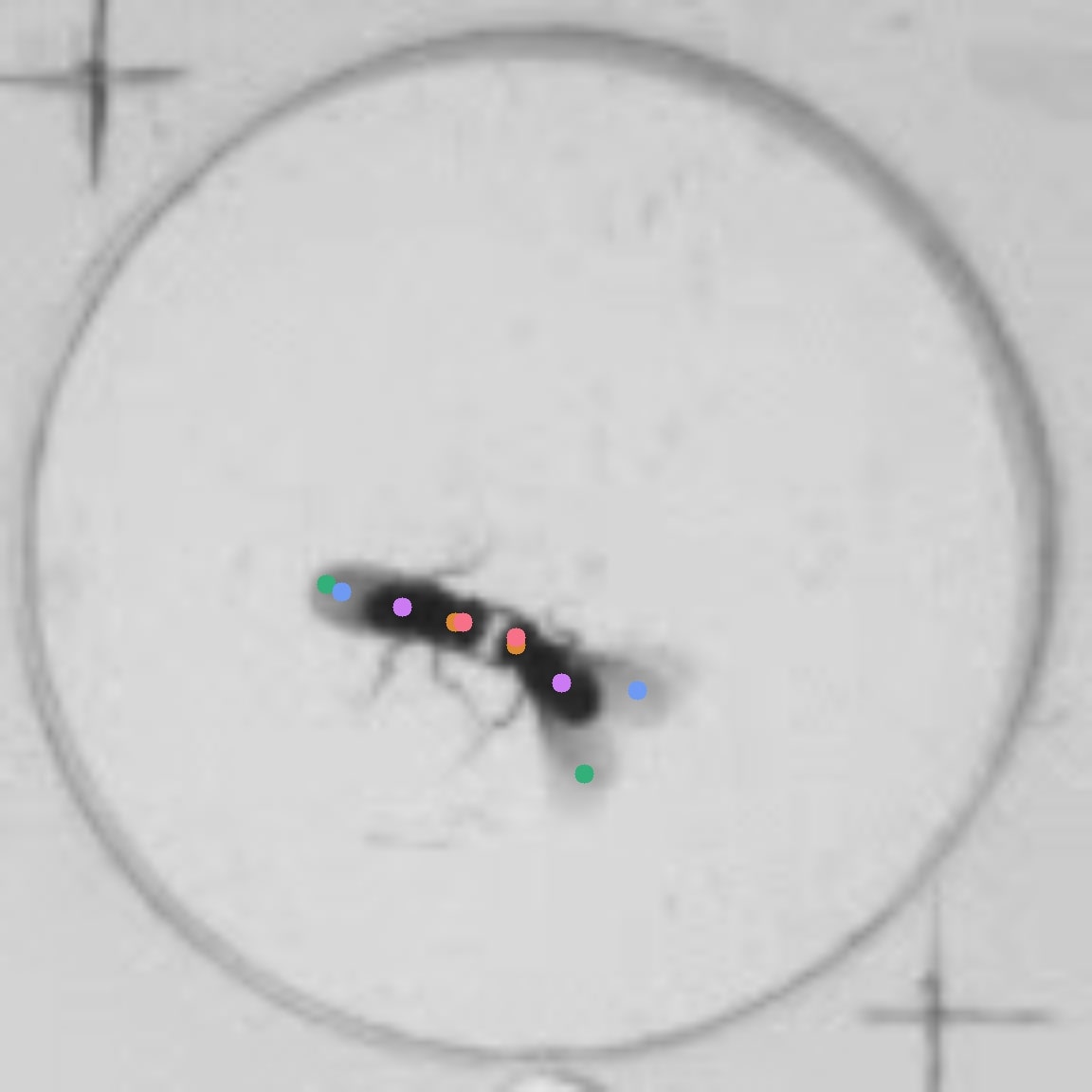}\hspace{\fighspace} & \includegraphics[width=\figsize\textwidth]{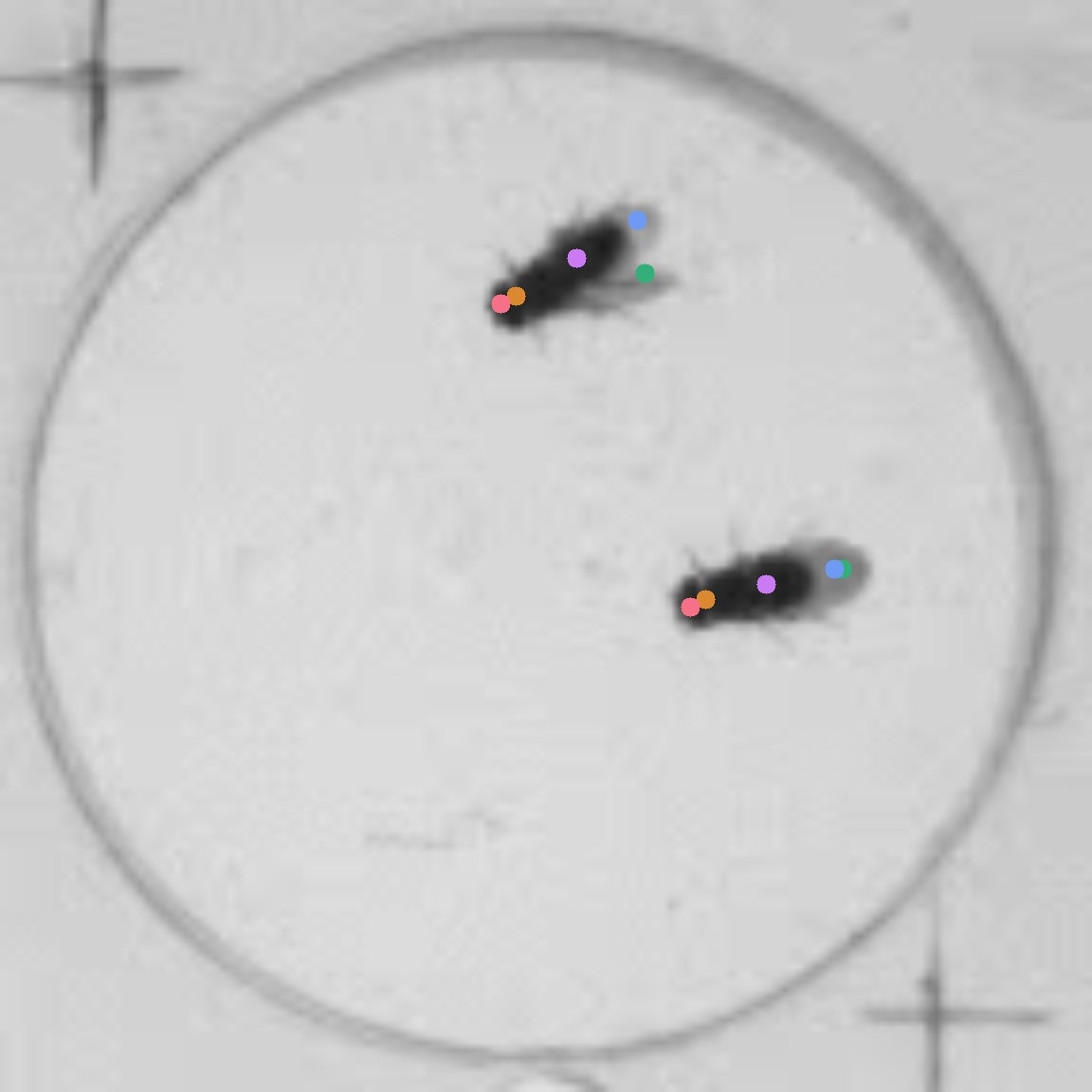}\hspace{\fighspacer} & \includegraphics[width=\figsize\textwidth]{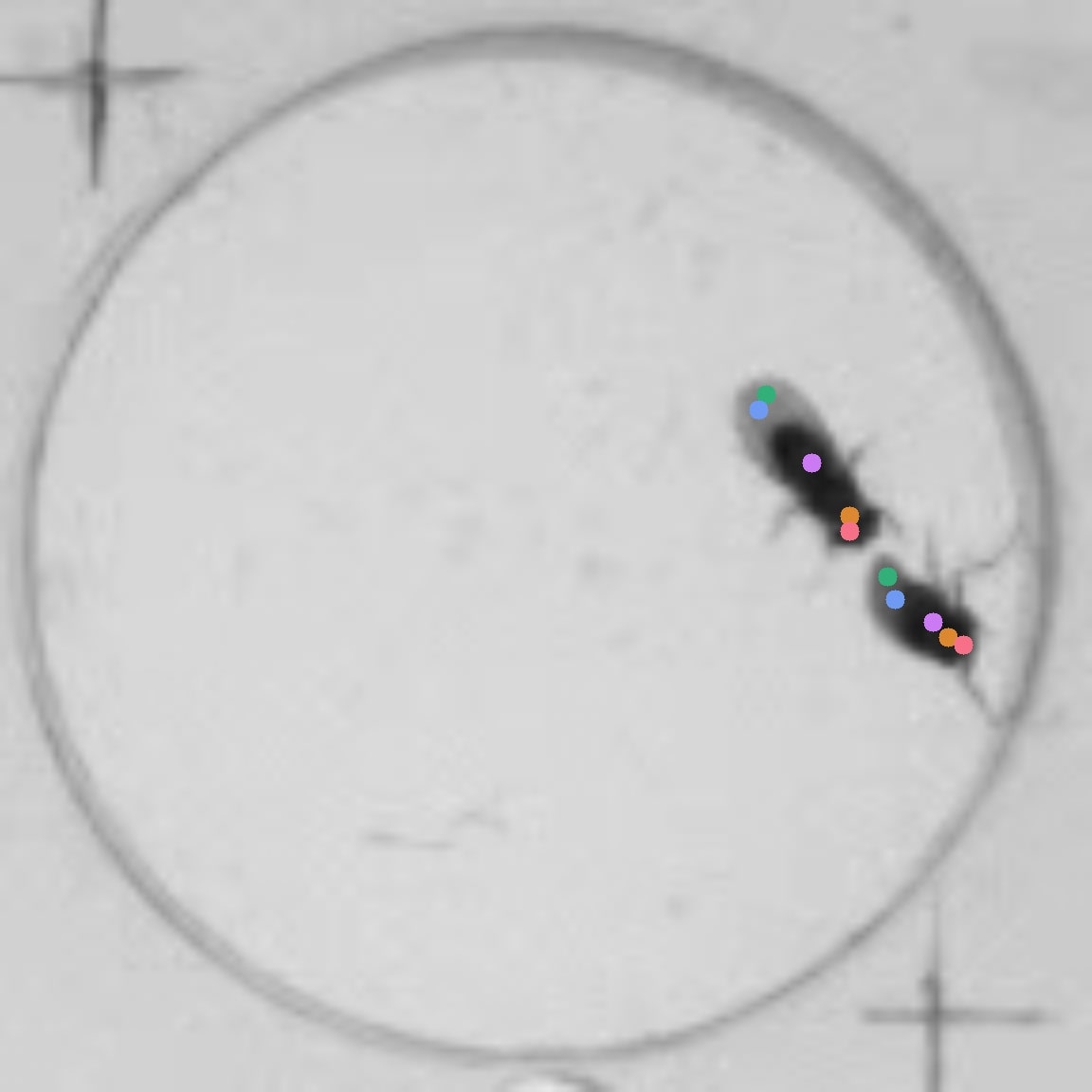}\hspace{\fighspace}  \\
\hspace{\fighspace}\includegraphics[width=\figsize\textwidth]{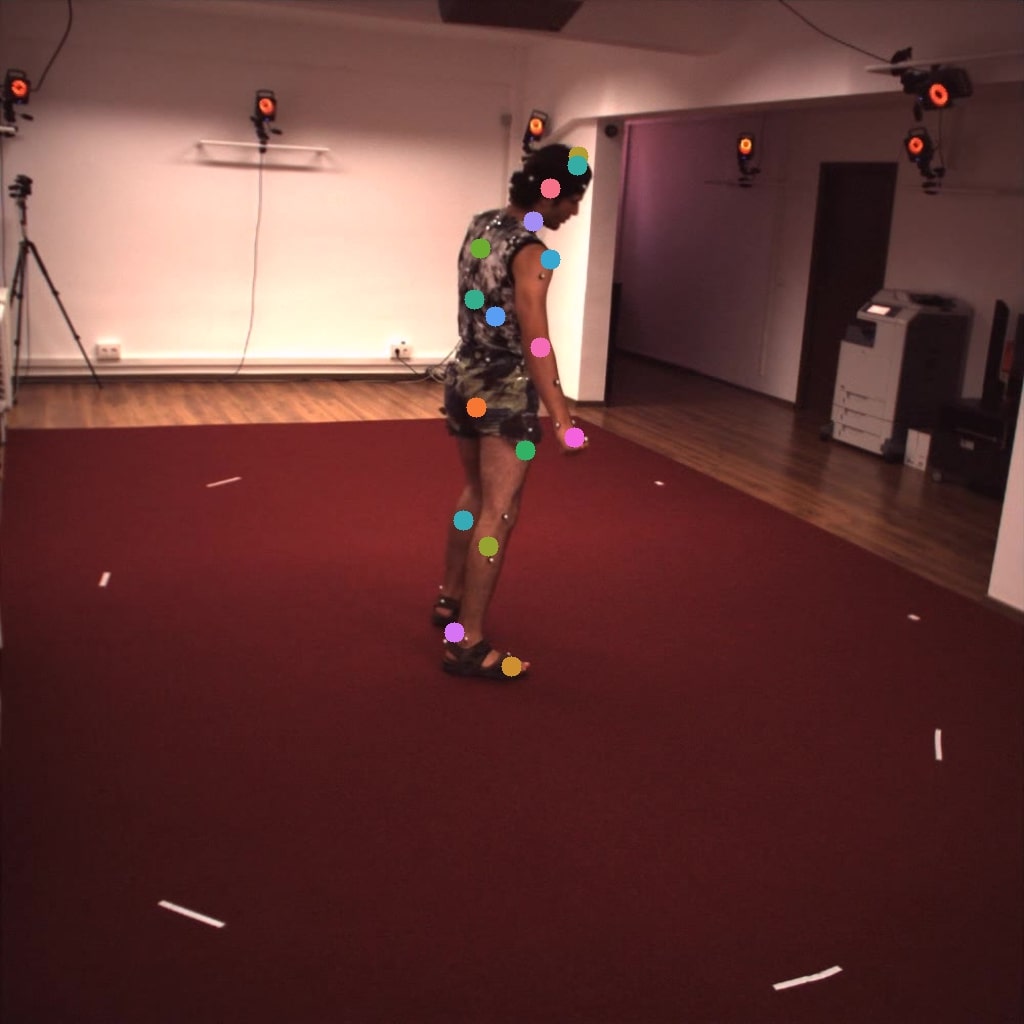}\hspace{\fighspace} & \includegraphics[width=\figsize\textwidth]{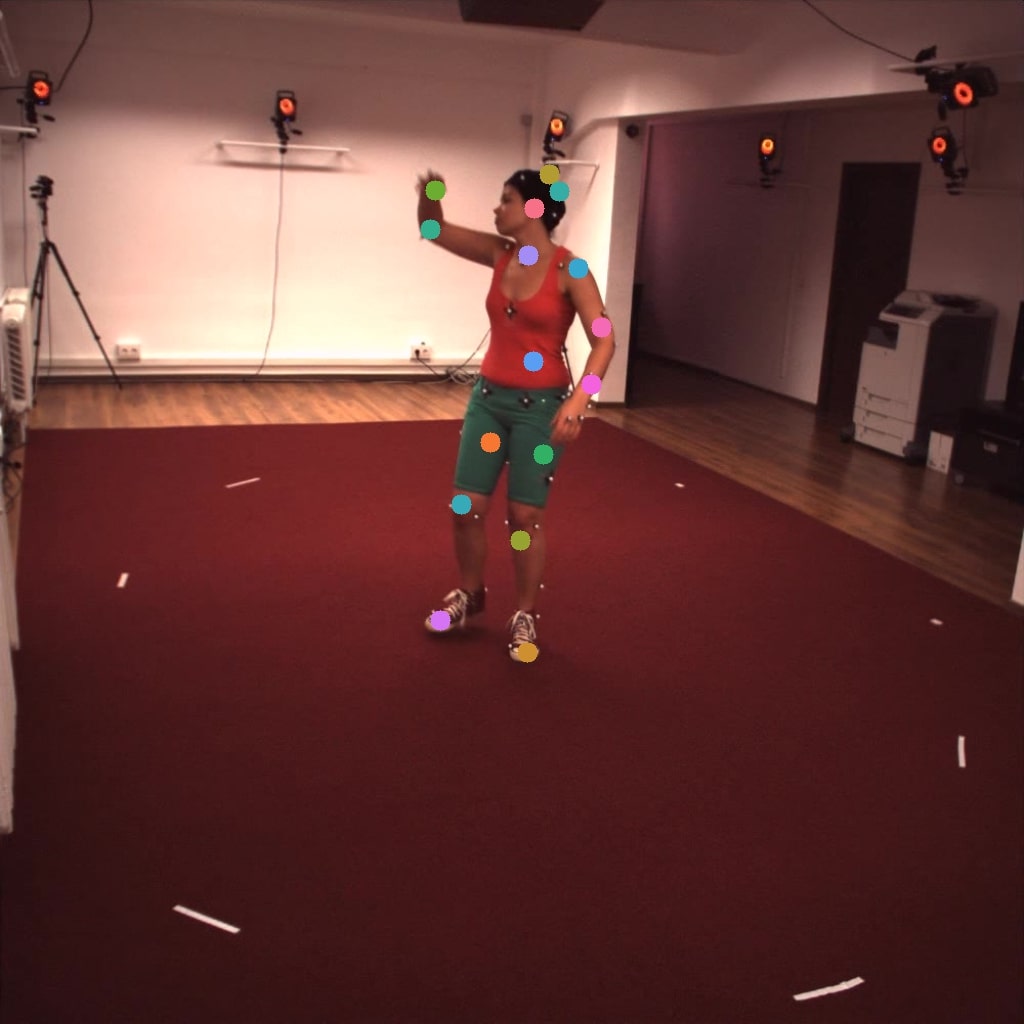}\hspace{\fighspacer} & \includegraphics[width=\figsize\textwidth]{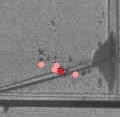}\hspace{\fighspace} & 
\includegraphics[width=\figsize\textwidth]{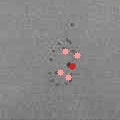}\hspace{\fighspace} & \includegraphics[width=\figsize\textwidth]{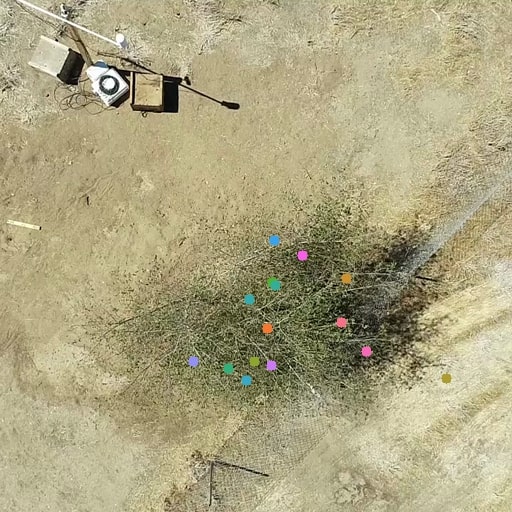}\hspace{\fighspacer} & \includegraphics[width=\figsize\textwidth]{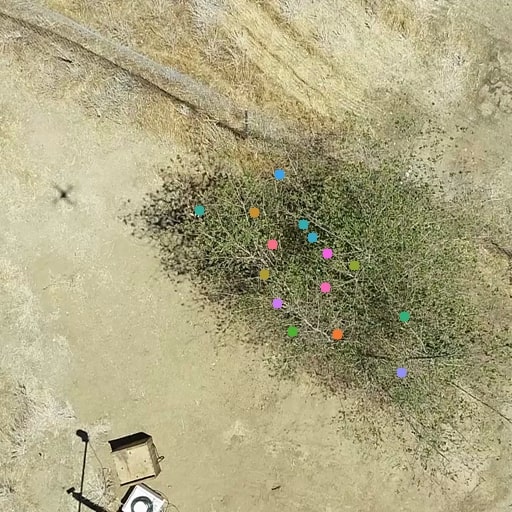}\hspace{\fighspace}  \\
\end{tabular}
\vspace{-0.2cm}
\caption{\textbf{Qualitative Results of B-KinD}. Qualitative results for B-KinD trained on CalMS21 (mouse), Fly vs. Fly (fly), Human3.6M (human), jellyfish and Vegetation (tree). Additional visualizations are in the Supplementary materials.}
\label{fig:qualitative_res}\vspace{-0.3cm}
\end{figure*}

\subsection{Additional applications}~\label{sec:qualitative_res}
\vspace{-0.3cm}

We show qualitative performance and demonstrate additional downstream tasks using our discovered keypoints, on pulse detection for Jellyfish and on wind speed regression for Vegetation. 

\textbf{Qualitative Results}. Qualitative results (Figure~\ref{fig:qualitative_res}) demonstrates that B-KinD is able to track some body parts consistently, such as the nose of both mice and keypoints along the spine; the body and wings of the flies; the mouth and gonads of the jellyfish; and points on the arms and legs of the human. For visualization only, we show only keypoints discovered with high confidence values (Section~\ref{sec:behavior_quant}); for all other experiments, we use all discovered keypoints. 

\textbf{Pulse Detection}. Jellyfish swimming is among the most energetically efficient forms of transport, and its control and mechanics are studied in hydrodynamics research~\cite{costello2021hydrodynamics}. Of key interest is the relationship between body plan and swim pulse frequency across diverse jellyfish species. By computing distance between B-KinD keypoints, we are able to extract a frequency spectrogram to study jellyfish pulsing, with a visible band at the swimming frequency (Supplementary materials). This provides a way to automatically annotate swimming behavior, which could be quickly applied to video from multiple species to characterize the relationship between swimming dynamics and body plan.

\textbf{Wind Speed Modeling}. Measuring local wind speed is useful for tasks such as tracking air pollution and weather forecasting~\cite{NEURIPS2019_a9ad5f28}. Oscillations of trees encode information on wind conditions, and as such, videos of moving trees could function as wind speed sensors~\cite{NEURIPS2019_a9ad5f28,cardona2021wind}.
Using the Vegetation dataset, we evaluate the ability of our keypoints to predict wind speed using a physics-based model~\cite{cardona2021wind}. This model defines the relationship between the mean wind speed and the structural oscillations of the tree, and requires tracking these oscillations from video, which was previously done manually. We show that B-KinD can accomplish this task automatically. Using our keypoints, we are able to regress the measured ground truth wind speed with an $R^2 =0.79$, suggesting there is a good agreement between the proportionality assumption from~\cite{cardona2021wind} and the experimental results using the keypoint discovery model. 

\textbf{Limitations}. One issue we did not explore in detail, and which will require further work, is keypoint discovery for agents that may be partially or completely occluded at some point during observation, including self-occlusion. 
Additionally, similar to other keypoint discovery models~\cite{ZhangKptDisc18,Lorenz19,SchmidtkeVELAK21}, we observe left/right swapping of some body parts, such as the legs in a walking human. One approach that might overcome these issues would be to extend our model to discover the 3D structure of the organism, for instance by using data from multiple cameras. 
Despite these challenges, our model performs comparably to supervised keypoints for behavior classification.


\section{Discussion and conclusion}

We propose B-KinD, a self-supervised method to discover meaningful keypoints from unlabelled videos for behavior analysis. 
We observe that in many settings, behavioral videos have fixed cameras recording agents moving against a (quasi) stationary background.
Our proposed method is based on reconstructing spatiotemporal difference between video frames, which enables B-KinD to focus on keypoints on the moving agents. 
Our approach is general, and is applicable to behavior analysis across a range of organisms without requiring manual annotations.

Results show that our discovered keypoints are semantically meaningful, informative, and enable performance comparable to supervised keypoints on the downstream task of behavior classification. Our method will reduce the time and cost dramatically for video-based behavior analysis, thus accelerating scientific progress in fields such as ethology and neuroscience. 

\vspace{-0.1in}
\section{Acknowledgements}

This work was generously supported by the Simons Collaboration on the Global Brain grant 543025 (to PP and DJA), NIH Award \#R00MH117264 (to AK), NSF Award \#1918839 (to YY), NSF Award \#2019712 (to JOD and RHG), NINDS Award \#K99NS119749 (to BW), NIH Award \#R01MH123612 (to DJA and PP), NSERC Award \#PGSD3-532647-2019 (to JJS), as well as a gift from Charles and Lily Trimble (to PP).

{\small
\bibliographystyle{ieee_fullname}
\bibliography{egbib}
}

\clearpage

\onecolumn

\appendix

\section*{Supplementary Material}

We present additional experimental results (Section~\ref{sec:addtional_results}), additional implementation details (Section~\ref{sec:additional_implementation}), and visualizations (Section~\ref{sec:visualizations}). Additional video visualizations and code are available in the project website: \url{https://sites.google.com/view/b-kind}.

{\bf Benefits and risks of this technology.} Automating the analysis of behavior is useful across many fields: in neuroscience, to study the neural control of behavior; in ethology and conservation, to study animal behavior and their response to human encroachment; in rehabilitation, to track patients' recovery of motor function; and in helping improve safety in the workplace. Risks are inherent in any application where humans behavior is analyzed, and care must be taken to respect privacy and human rights. Responsible use in research requires following all applicable rules and policies, including filing for permission with the relevant internal review board (IRB), and obtaining written informed consent from human subjects being filmed.

\section{Additional Experimental Results}~\label{sec:addtional_results}
\vspace{-0.2in}

\subsection{CalMS21 Ablation Study}~\label{sec:mouse_ablation}

Similar to the main paper, we evaluate CalMS21 on the behavior classification train/test split described in task 1~\cite{sun2021multi}, and show results using Mean Average Precision (MAP) across the annotated behavior classes. We use B-KinD keypoints as input to behavior classification to compare against supervised and other self-supervised baselines.

\begin{table}[b]
    \begin{center}
        \scalebox{1.0}{
        \begin{tabular}{cc|c || cc | c}
            \toprule[0.2em]
            Hyperparam. & Value & MAP & Hyperparam. & Value & MAP \\
            \toprule[0.2em]
             & 6 & $.852 \pm .013$ & & 6 & $.850 \pm .017$ \\
            Frame Gap & 12 & $.862 \pm .012$ & \# keypoints & 10 & $.852 \pm .013$\\
             & 30 & $.839 \pm .003$ & & 20* & $.868 \pm .008$ \\
            \bottomrule[0.1em]
        \end{tabular}
        }
    \end{center}
    \caption{\textbf{Effect of Hyperparameters on CalMS21}. For frame gap experiments, the number of keypoints is set to 10. For experiments with varying number of keypoints, frame gap is set to 6. All keypoints, confidence, and covariance are used as inputs, except (*) for the experiments with 20 keypoints, where only high-confidence keypoints are used (11 keypoints) since a high proportion of keypoints are discovered on the background. Mean and standard dev from 5 classifier runs are shown.}
    \label{tab:ablation_mouse}
\end{table}

\textbf{Effect of Hyperparameters}. For all experiments on CalMS21, we use a frame gap of 6 with 10 discovered keypoints.
Here, we vary the number of discovered keypoints and frame gap for our model, and apply the learned keypoints to behavior classification (Table~\ref{tab:ablation_mouse}). There are small variations in performance, in particular, the downstream performance generally improves with increasing the number of keypoints, and a frame gap of 6 or 12 works better than larger frame gaps. We note that the number of low confidence background keypoints also increases with the number of discovered keypoints (Figure~\ref{fig:qual_kps}), and due to the large proportion of background keypoints, we do not use background keypoints in the 20 keypoints case for the classification task. In all cases, we note that we do better than other self-supervised baselines even with bounding box information (MAP = $.819$) for this task.

\def\figsize{0.2}
\def\fighspace{-2mm}
\def\fighspacer{-2mm}
\begin{figure*}[h]
\centering
\begin{tabular}{m{0.08\textwidth}m{\figsize\textwidth}m{\figsize\textwidth}m{\figsize\textwidth}m{\figsize\textwidth}}
\centering
\small{6 keypoints} & \includegraphics[width=\figsize\textwidth]{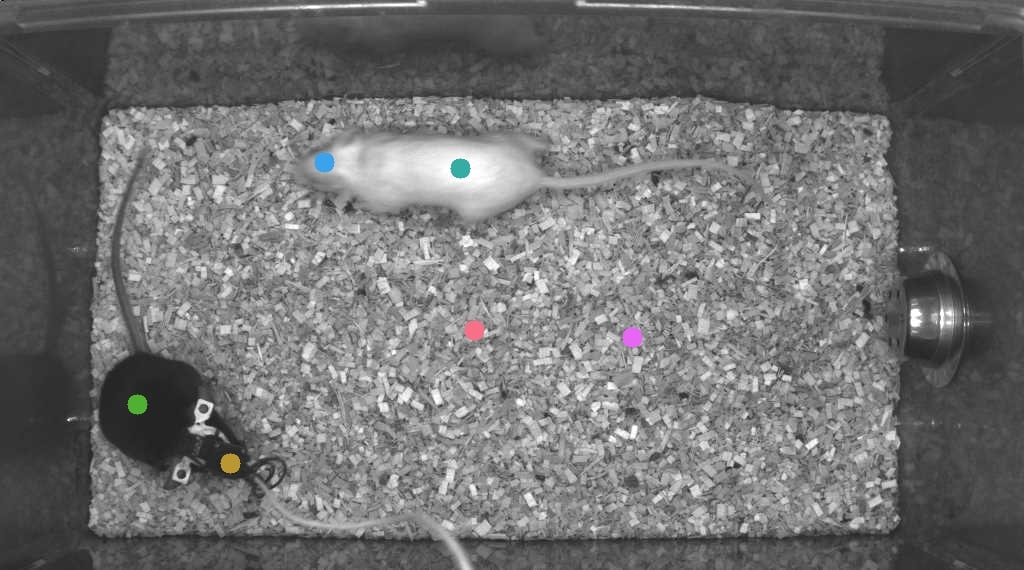}\hspace{\fighspacer} & \includegraphics[width=\figsize\textwidth]{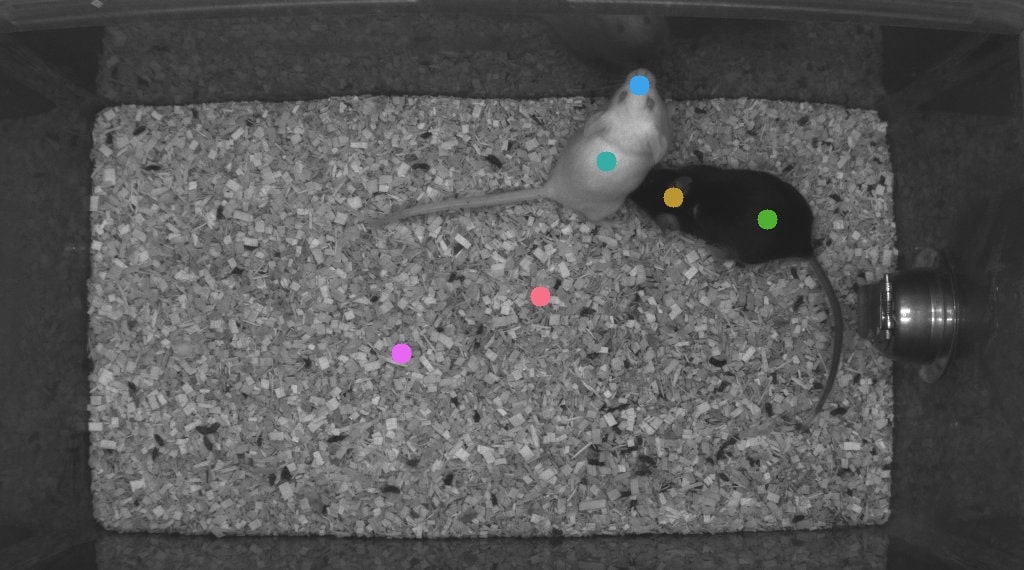}\hspace{\fighspace} & 
\includegraphics[width=\figsize\textwidth]{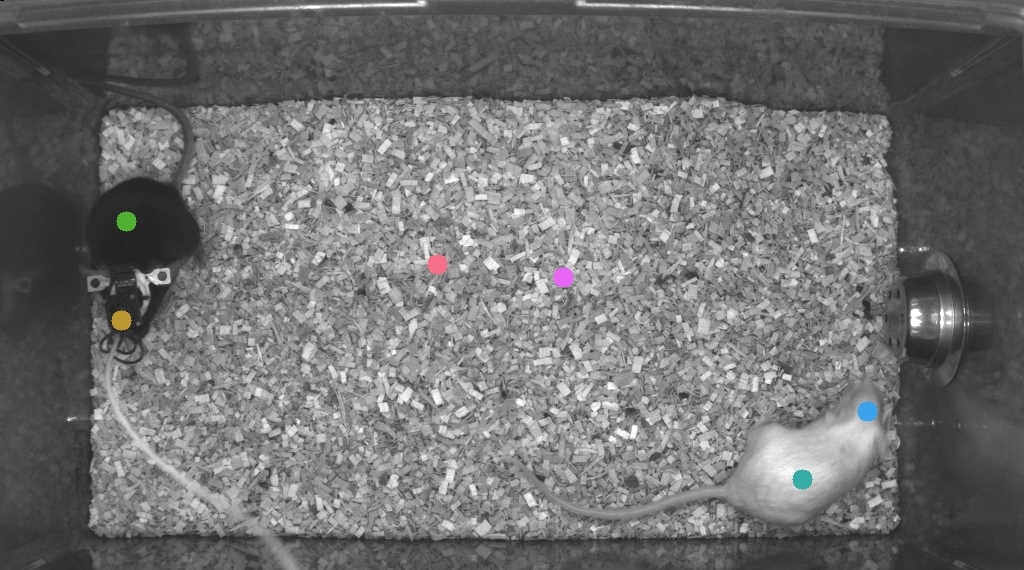}\hspace{\fighspace} & \includegraphics[width=\figsize\textwidth]{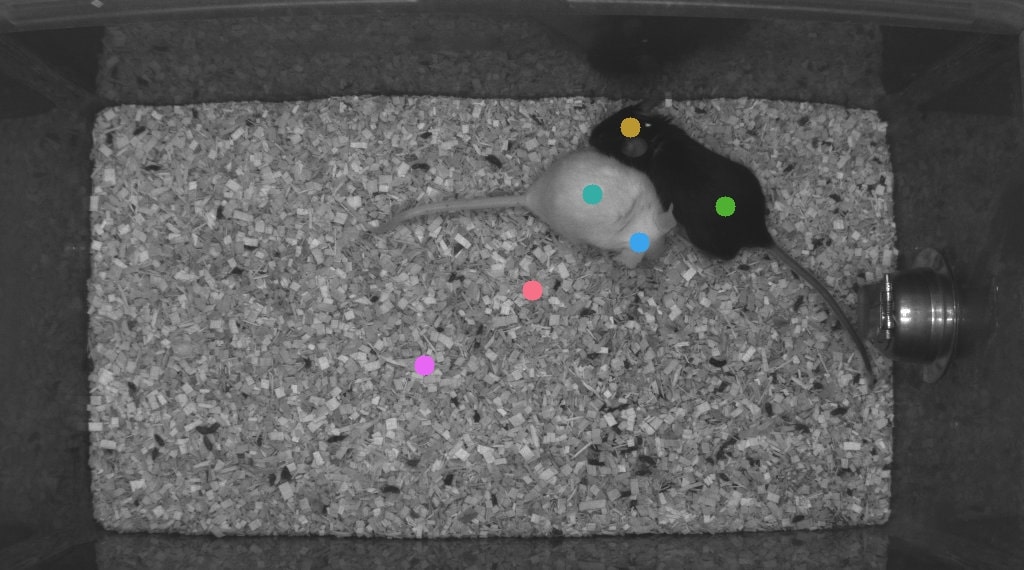}\hspace{\fighspacer} \\
\centering
\small{10 keypoints} & \includegraphics[width=\figsize\textwidth]{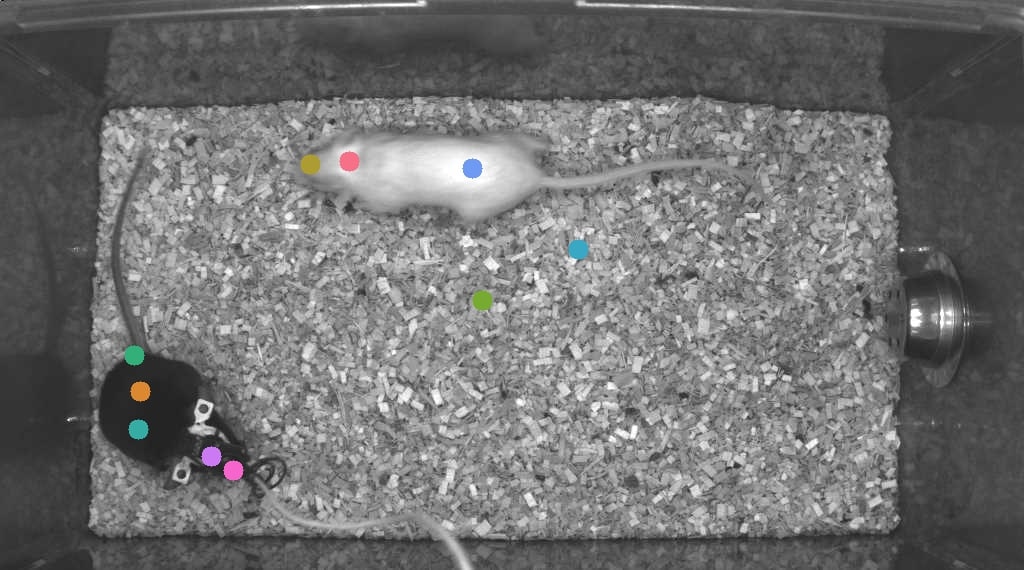}\hspace{\fighspacer} & \includegraphics[width=\figsize\textwidth]{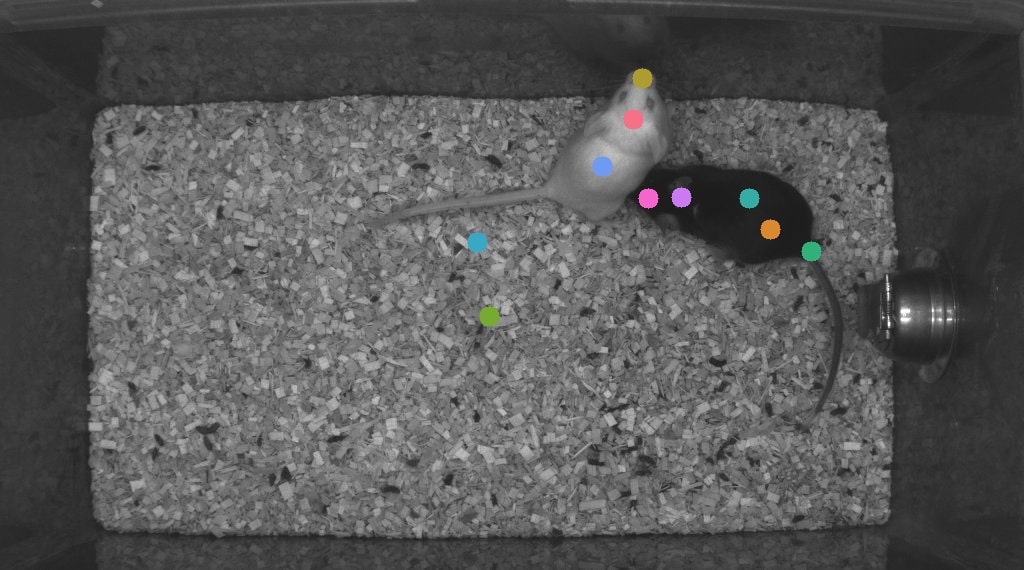}\hspace{\fighspace} & \includegraphics[width=\figsize\textwidth]{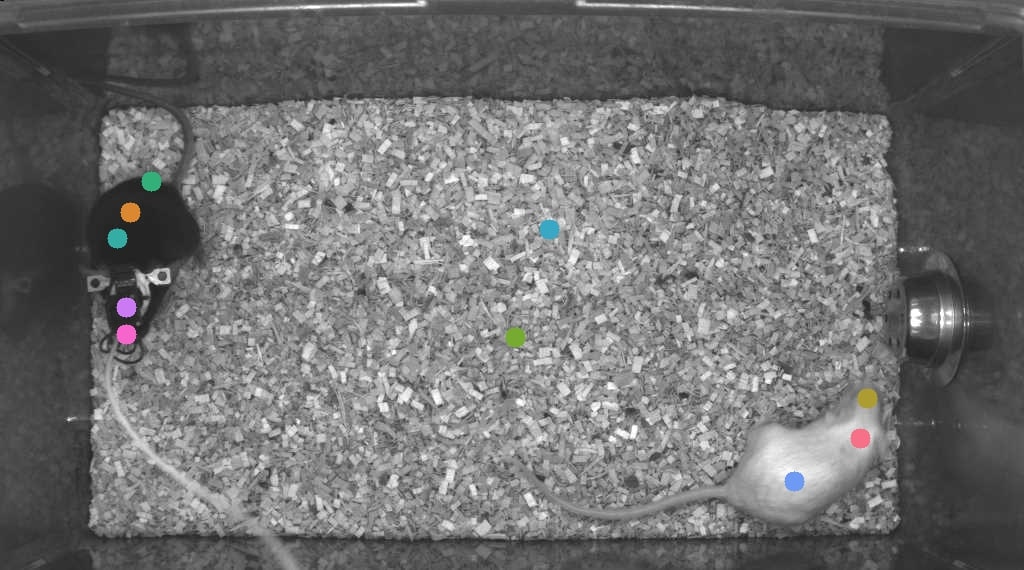}\hspace{\fighspacer} & \includegraphics[width=\figsize\textwidth]{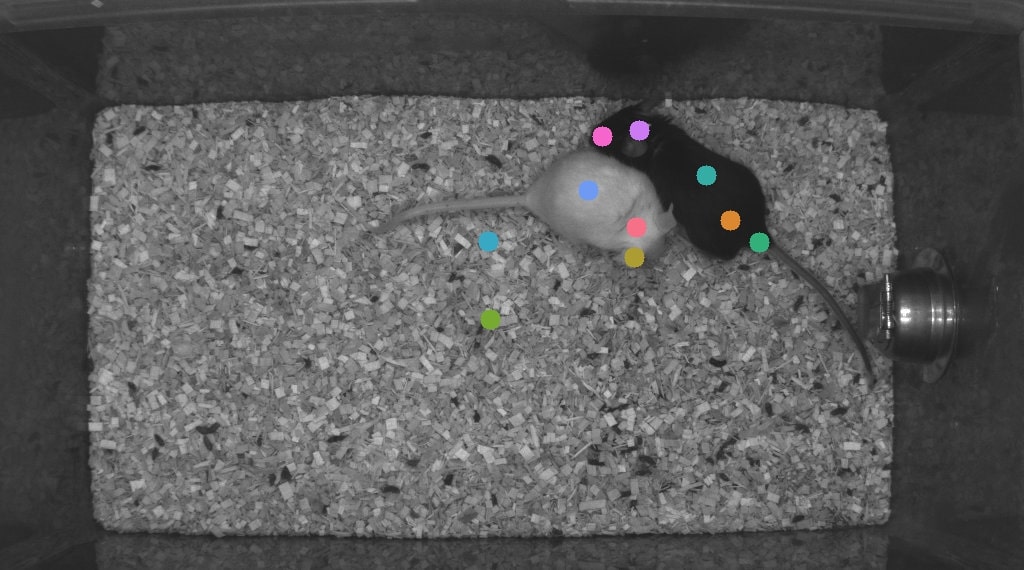}\hspace{\fighspacer} \\
\centering
\small{20 keypoints} & \includegraphics[width=\figsize\textwidth]{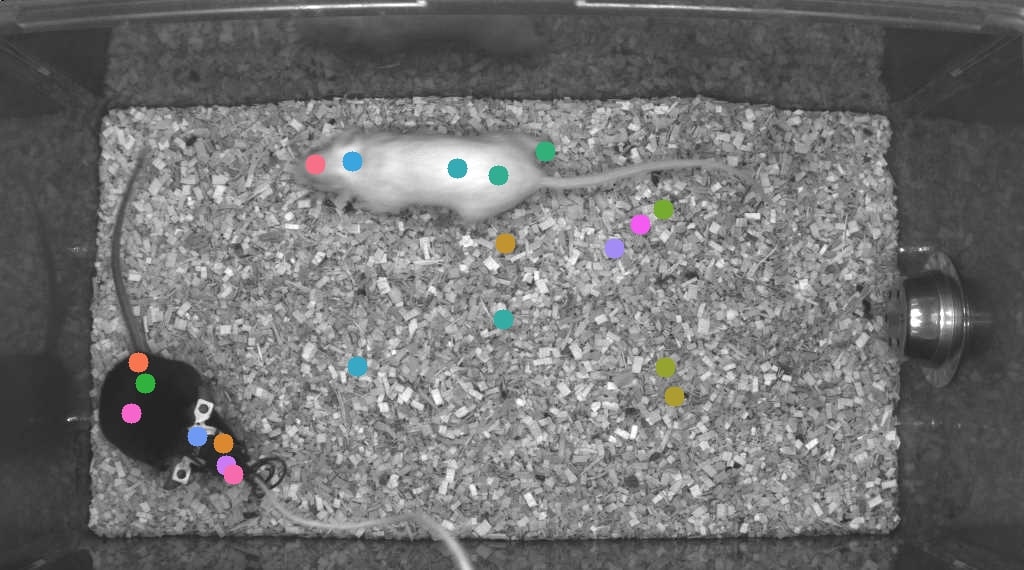}\hspace{\fighspacer} & \includegraphics[width=\figsize\textwidth]{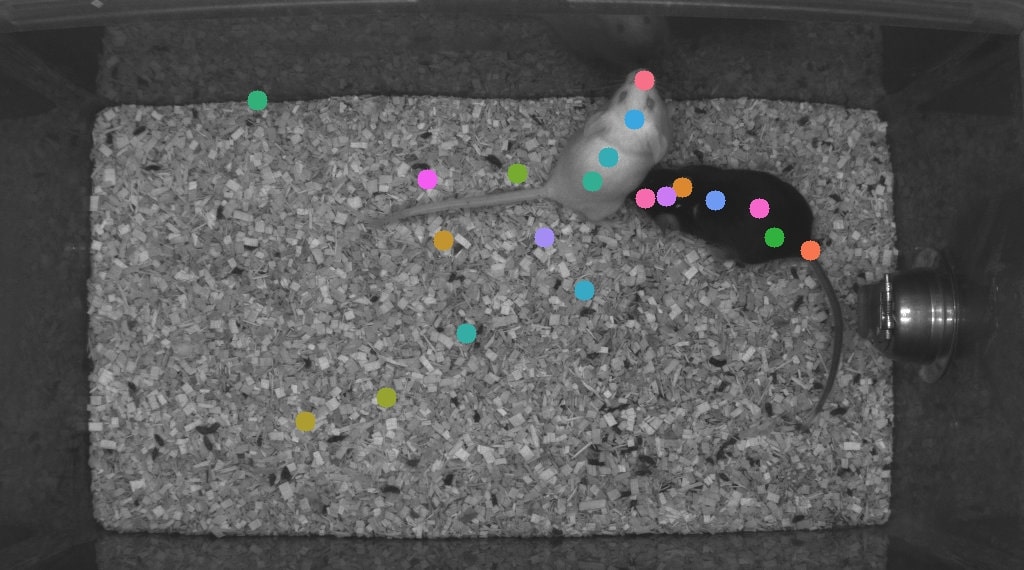}\hspace{\fighspace} & \includegraphics[width=\figsize\textwidth]{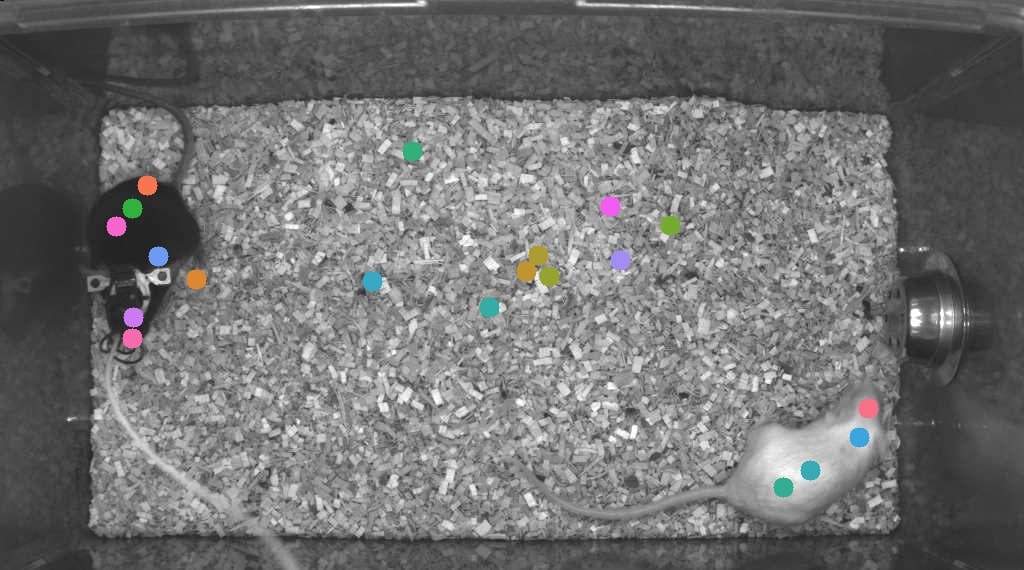}\hspace{\fighspacer} & \includegraphics[width=\figsize\textwidth]{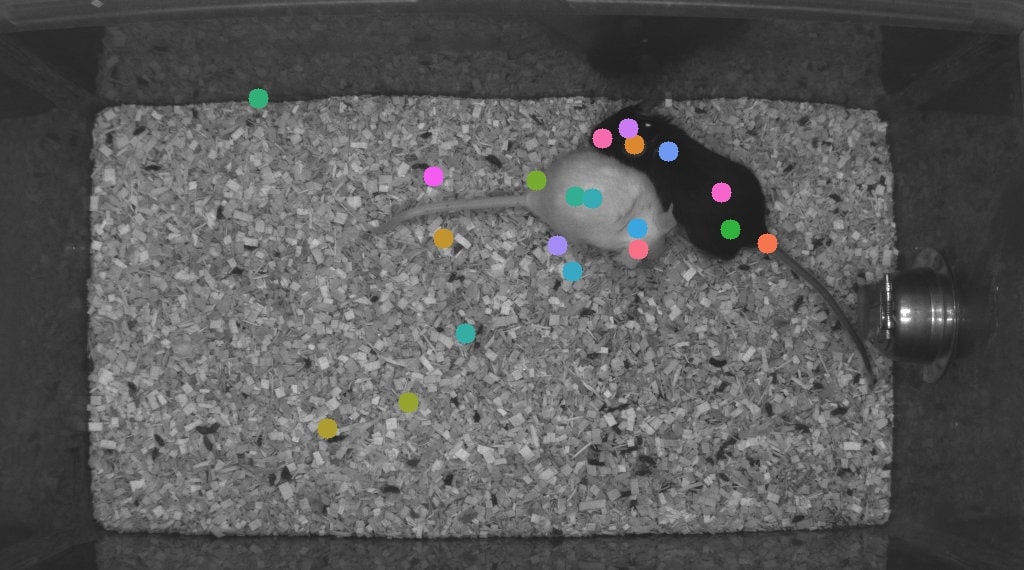}\hspace{\fighspacer} \\
\end{tabular}
\caption{\textbf{Qualitative Results on CalMS21 by varying the number of keypoints}. We train the keypoint discovery model with different numbers of discovered keypoints. Each row shows qualitative results with all the keypoints including the background keypoints. We note that there are 2 background (low-confidence) keypoints for 6 and 10 discovered keypoints, and 9 background keypoints for 20 discovered keypoints.}
\label{fig:qual_kps}
\end{figure*}

\textbf{Varying Amount of Unlabeled Video Data}. We vary the amount of input data (unlabelled image pairs) used to train B-KinD, and observe comparable performance at different amounts of data availability (Table~\ref{tab:ablation_train}). In particular, we are able to achieve comparable performance on behavior classification to supervised keypoints (Table~\ref{tab:mouse_behavior_class}) by using only $7.8k$ input training pairs in our model (approximately 4 minutes of video recorded at 30Hz; approximately 30 minutes of video considering no overlaps on selected image pairs). We note that this experiment is varying the amount of unlabelled data for training the keypoint discovery model, while the train/test split for evaluating the behavior classifier stays the same.

\begin{table}
    \begin{center}
        \scalebox{1.0}{
        \begin{tabular}{ccc }
            \toprule[0.2em]
            \# Training Pairs &  Corresponding Video Length (30Hz) & MAP \\
            \toprule[0.2em]
             7.8k & $4.3$ min & $.867 \pm .003$\\
             18k & $10$ min & $.840 \pm .016$\\
             26k & $14$ min & $.852 \pm .013$\\
            \bottomrule[0.1em]
        \end{tabular}
        }
    \end{center}
    \caption{\textbf{Effect of Varying Training Data Amount for Keypoint Discovery}. We train the keypoint discovery model with different amounts of input training frame pairs from video. Different training amounts are selected by choosing random video subsets from the full set of CalMS21 training videos. Image pairs are sampled from videos with a gap of 6 frames, and between pairs to ensure no overlaps, there is a gap of 7 frames. All keypoints, confidence, and covariance values on 10 discovered keypoints are used. Mean and standard dev from 5 classifier runs are shown.}
    \label{tab:ablation_train}
\end{table}

\textbf{Loss Ablation Study}. 
We compare B-KinD trained with the full objective (reconstruction, rotation equivariance, separation) to one trained only on spatiotemporal difference reconstruction (Table~\ref{tab:mouse_loss}).
The rotation equivariance loss is qualitatively important for tracking semantically consistent parts of the mouse (Figure~\ref{fig:loss_qual}) and the separation loss prevents the model from predicting keypoints at the center of the image, which are rotationally consistent but do not track semantic body parts. The full objective is important to achieving comparable performance to supervised baselines. We would like to note that the image reconstruction baselines in our main results are also trained with the full objective, except the reconstruction is based on image reconstruction. Additionally, since keypoint locations are not consistent for spatiotemporal difference reconstruction only, we note that adding confidence and covariance significantly improves the performance of the reconstruction loss only model (Table~\ref{tab:mouse_loss}).

\textbf{Single Geometry Branch}. Our proposed model extracts appearance features from the frame at time $t$, and two geometry features (the keypoints), where one is for frame at time $t$ and the other is for frame at time $t+k$. It is also possible to train the model using only one geometry branch only for the frame at time $t+k$, without the geometry branch for time $t$. On CalMS21, training B-KinD with one geometry branch reduced the classification performance from MAP of $0.852 \pm 0.013$ (full model) to $0.835 \pm 0.013$ (single branch). 

\begin{table}
  \begin{center}
\scalebox{0.9}{
    \begin{tabular}{lccccc}
        \toprule[0.2em]
        CalMS21 & Pose & Conf & Cov & Ours (MAP) & Reconstruction (MAP)\\
        \toprule[0.2em]
        \multirow{3}{*}{Loss Variation} & \checkmark & & & $.814 \pm .007$ & $.695 \pm .022$ \\
        & \checkmark & \checkmark & & $.857 \pm .005$ & $.776 \pm .012$ \\
        & \checkmark & \checkmark & \checkmark & $.852 \pm .013$ & $.794 \pm .008 $ \\   
        \bottomrule[0.1em]
    \end{tabular}}
  \caption{\textbf{Loss Variations on CalMS21}. ``Ours" represents training B-KinD keypoints with the full objective (reconstruction, rotation equivariance, separation) and ``Reconstruction" indicates training with spatiotemporal difference reconstruction only. Mean and standard dev from 5 classifier runs are shown. }
  \label{tab:mouse_loss}
  \end{center}
\end{table}

\def\figsize{0.2}
\def\fighspace{-2mm}
\def\fighspacer{-2mm}
\begin{figure*}[h]
\centering
\begin{tabular}{m{0.08\textwidth}m{\figsize\textwidth}m{\figsize\textwidth}m{\figsize\textwidth}m{\figsize\textwidth}}
\centering
\small{No rotation} & \includegraphics[width=\figsize\textwidth]{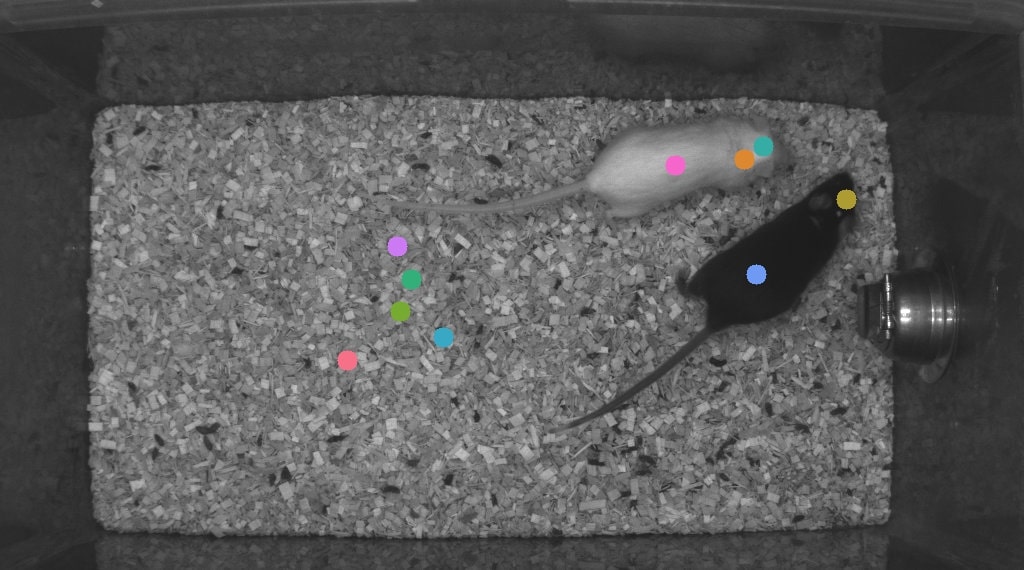}\hspace{\fighspacer} & \includegraphics[width=\figsize\textwidth]{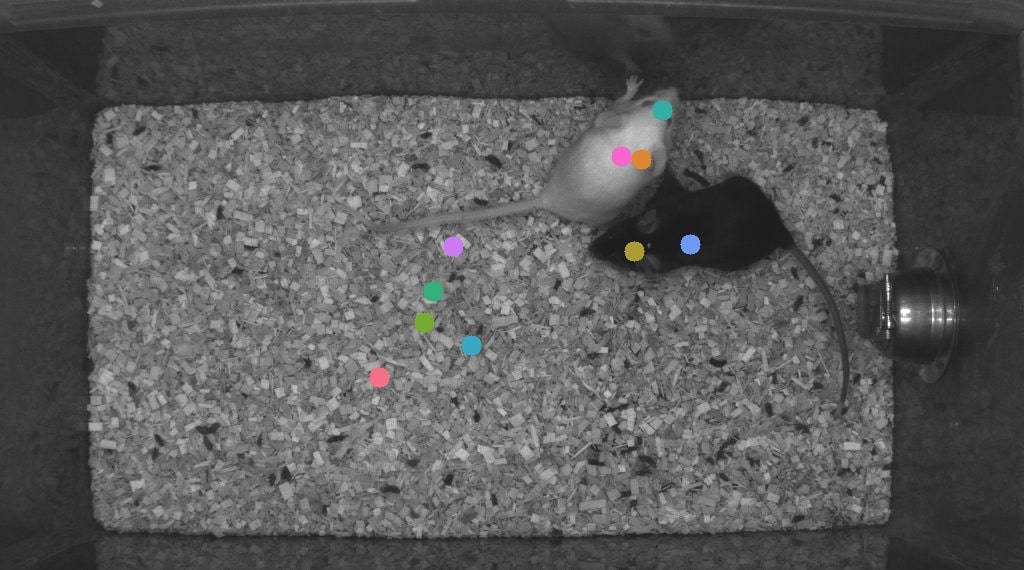}\hspace{\fighspace} & 
\includegraphics[width=\figsize\textwidth]{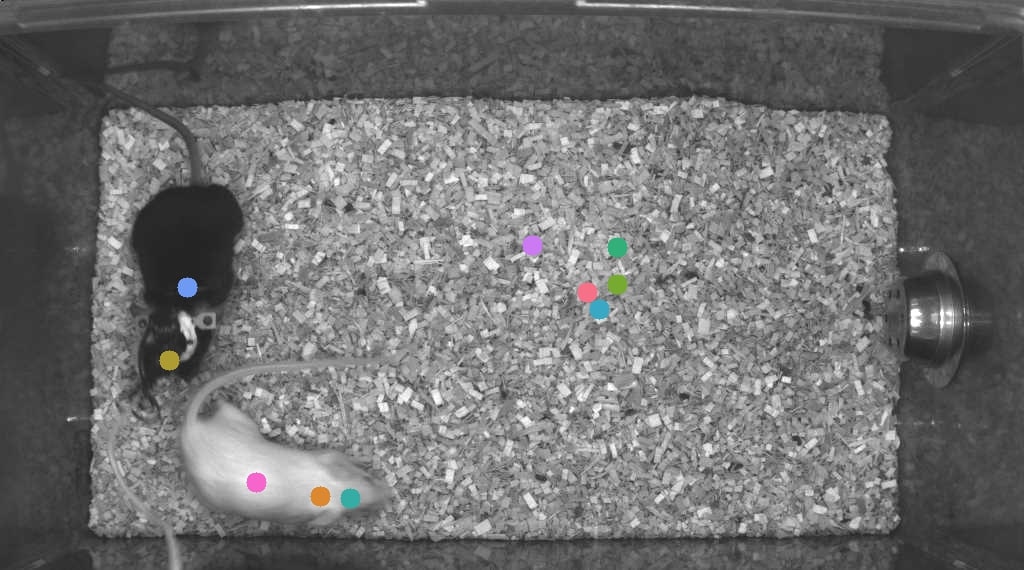}\hspace{\fighspace} & \includegraphics[width=\figsize\textwidth]{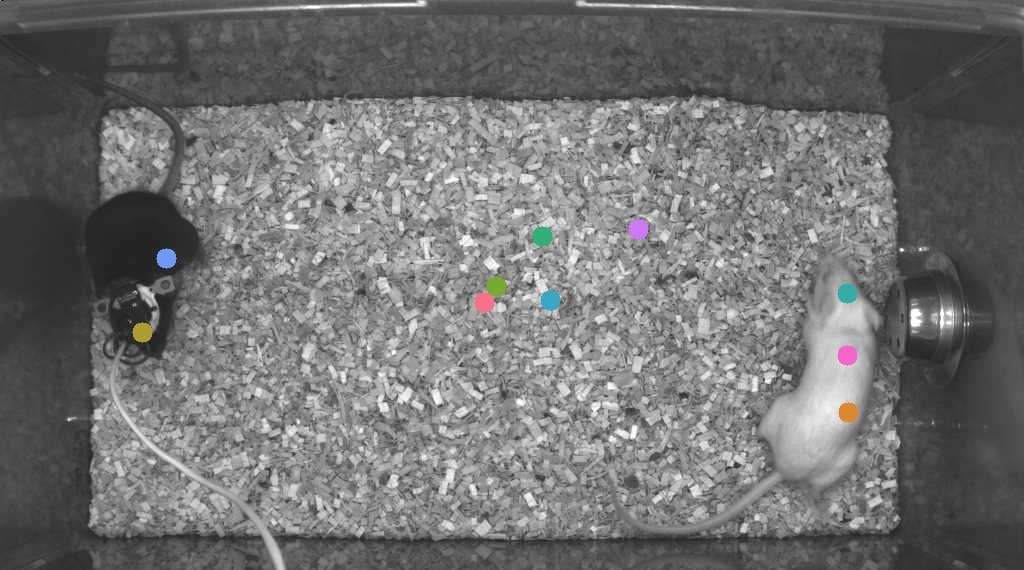}\hspace{\fighspacer} \\
\centering
\small{With rotation} & \includegraphics[width=\figsize\textwidth]{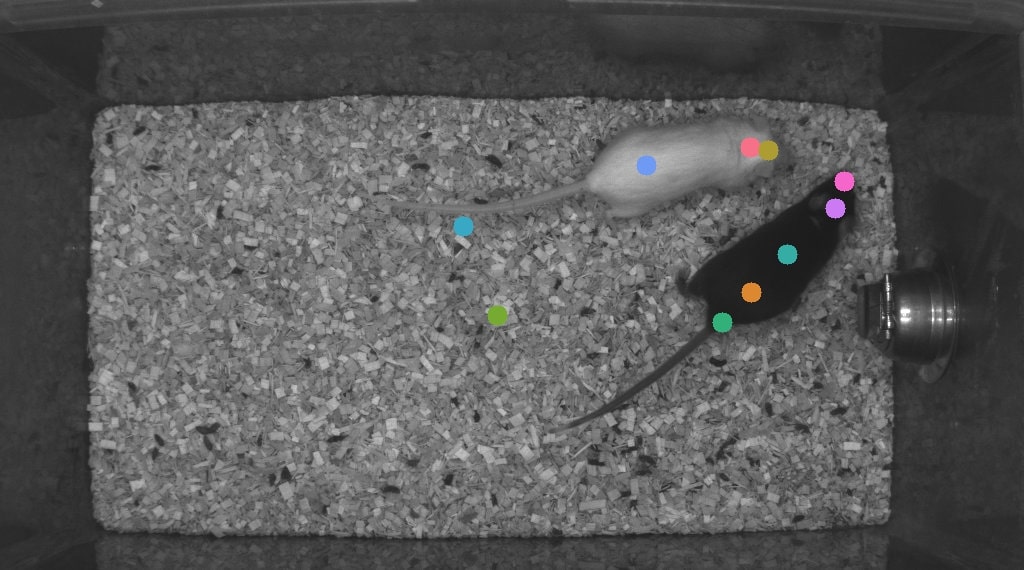}\hspace{\fighspacer} & \includegraphics[width=\figsize\textwidth]{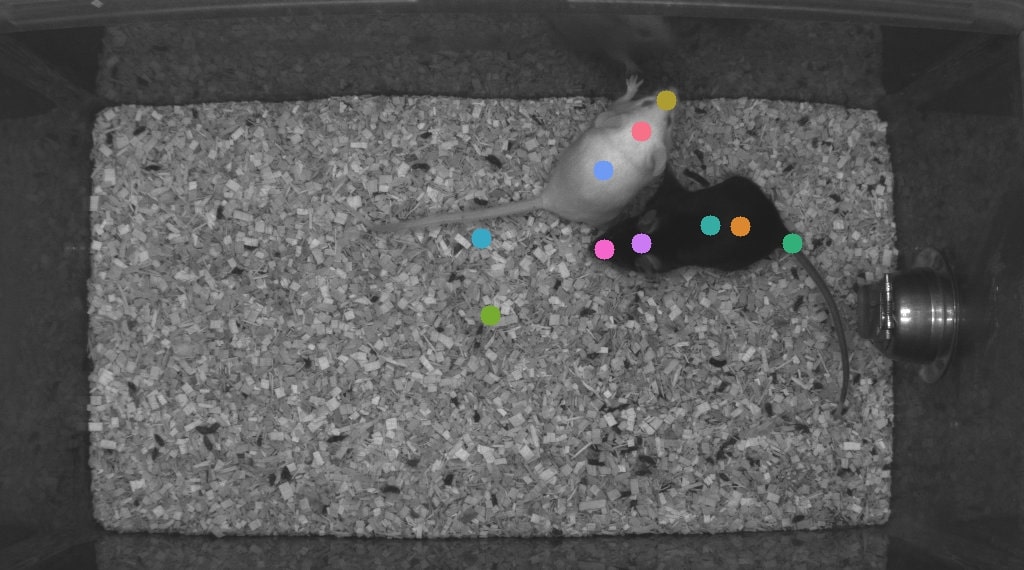}\hspace{\fighspace} & \includegraphics[width=\figsize\textwidth]{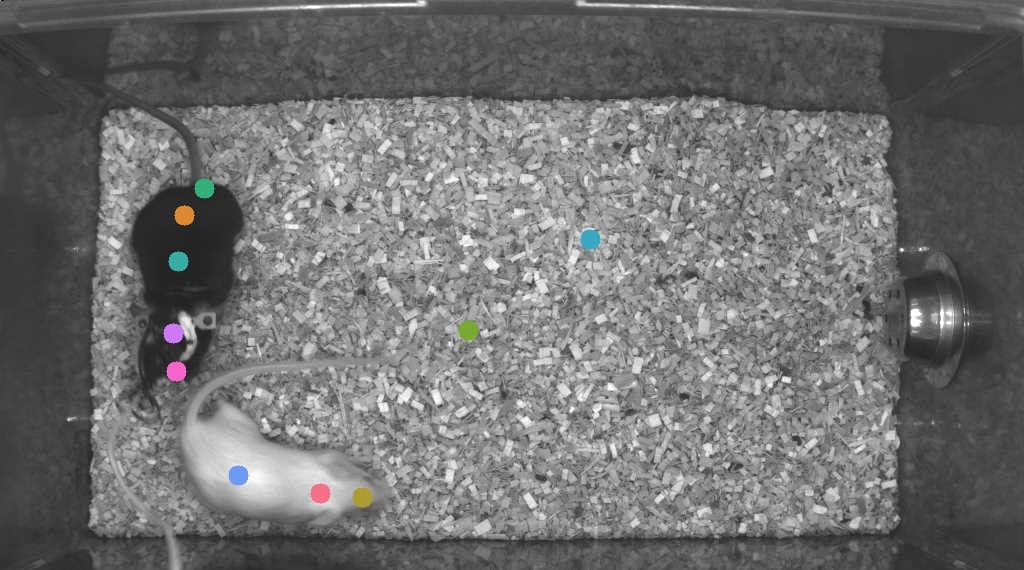}\hspace{\fighspacer} & \includegraphics[width=\figsize\textwidth]{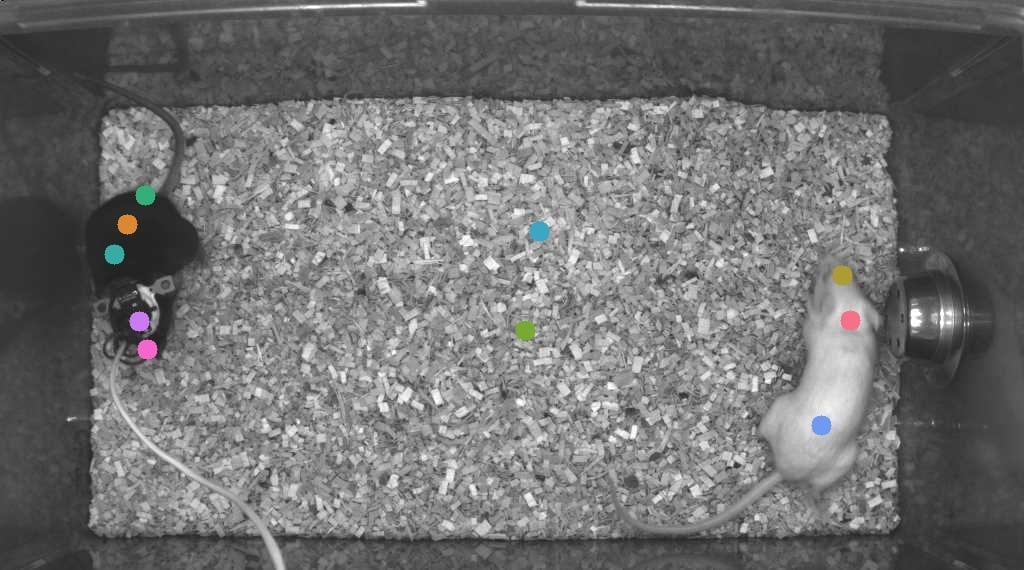}\hspace{\fighspacer} \\
\end{tabular}
\caption{\textbf{Qualitative Results on CalMS21 for loss ablation study}. With the full training objective for our discovered keypoints, we are able to track 8/10 keypoints consistently, while without rotation loss, there are only 5/10 tracked keypoints on both mice. Additionally, some of the discovered keypoints without rotation are not semantically consistent (for example, the pink and orange keypoints, two keypoints on the body of the white mouse, shift in order as the white mouse moves around). See quantitative results in Table~\ref{tab:mouse_loss}.}
\label{fig:loss_qual}
\end{figure*}

\subsection{CalMS21 Per-Class Performance}

B-KinD keypoints achieve comparable performance to supervised keypoints when using pose and confidence features from the heatmap (Table~\ref{tab:mouse_behavior_class}). For both supervised keypoints and our keypoints, the behavior classes with the biggest improvement when adding confidence features is on the ``Attack" class, which contains frames with occlusion and motion blur since the mice are moving quickly and chasing/tussling. Heatmap confidence and covariance values provides more information about the detected part (Figure~\ref{fig:mouse_confidence}). For example, when a part is well localized (ex: visible nose of mouse), our keypoint discovery network produces a heatmap with a single high peak with low variance; conversely, when a target part is occluded, the heatmap contains a blurred shape with lower peak value. We note that performance is similar for the supervised keypoints and our keypoints on the ``Investigation" and ``Mount" classes.

\begin{table}
  \begin{center}
\scalebox{0.9}{
    \begin{tabular}{lccccccc}
        \toprule[0.2em]
        CalMS21 & Pose & Conf & Cov & MAP & Attack AP & Investigation AP & Mount AP\\
        \toprule[0.2em]
        \multicolumn{8}{c}{\textbf{\textit{Fully supervised}}} \\
        \multirow{3}{*}{MARS $\dagger$ ~\cite{segalin2020mouse}} & \checkmark & & & $.856 \pm .010$ & $.724 \pm .023$ & $.893 \pm .005$ & $.950 \pm .004$\\
        & \checkmark & \checkmark & & $.874 \pm .003$ & $.790 \pm .004$ & $.890 \pm .006$ & $.943 \pm .004$\\
        & \checkmark & \checkmark & \checkmark & $.880 \pm .005$ & $.804 \pm .012$ & $.902 \pm .004$ & $.934 \pm .006$\\
        \bottomrule[0.1em]
        \multicolumn{8}{c}{\textbf{\textit{Self-supervised}}} \\   
        Jakab et al.~\cite{JakabNeurips18} & \checkmark & &  & $.186 \pm .008$ & $.135 \pm .019$ & $.254 \pm .019$ & $.170 \pm .029$\\
        \hline
        \multirow{3}{*}{Image Recon.} & \checkmark & & & $.182 \pm .007$ & $.111 \pm .016$ & $.217 \pm .011$ & $.219 \pm .021$\\
        & \checkmark & \checkmark & & $.184 \pm .006$ & $.114 \pm .006$ & $.209 \pm .012$ & $.229 \pm .021$\\
        & \checkmark & \checkmark & \checkmark & $.165 \pm .012$ & $.110 \pm .016$ & $.218 \pm .013$ & $.167 \pm .038$ \\
        \hline
        \multirow{3}{*}{Image Recon. bbox$\dagger$} & \checkmark & & & $.819 \pm .008$ & $.680 \pm .028$ & $.861 \pm .007$ & $.918 \pm .007$\\
        & \checkmark & \checkmark & & $.812 \pm .006$ & $.694 \pm .011$ & $.818 \pm .016$ & $.923 \pm .013$\\
        & \checkmark & \checkmark & \checkmark & $.812 \pm .010$ & $.709 \pm .008$ & $.806 \pm .019$ & $.922 \pm .013$\\ \hline
        \multirow{3}{*}{Ours} & \checkmark & & & $.814 \pm .007$ & $.654 \pm .025$ & $.861 \pm .003$ & $.925 \pm .014$\\
        & \checkmark & \checkmark & & $.857 \pm .005$ & $.763 \pm .015$ & $.879 \pm .009$ & $.928 \pm .006$\\
        & \checkmark & \checkmark & \checkmark & $.852 \pm .013$ & $.751 \pm .025$ & $.870 \pm .009$ & $.935 \pm .010$\\   
        \bottomrule[0.1em]
    \end{tabular}}
  \caption{\textbf{Per-Class Behavior Classification Results on CalMS21}. ``Ours" represents classifiers using input keypoints from B-KinD. ``conf" represents using the confidence score, and ``cov" represents values from the covariance matrix of the heatmap. $\dagger$ refers to models that require bounding box inputs before keypoint estimation. Mean and standard dev from 5 classifier runs are shown. }
  \label{tab:mouse_behavior_class}
  \end{center}
\end{table}

\subsection{Human3.6M Ablation Study}\label{sec:human_ablation}

\begin{table}
    \begin{center}
        \scalebox{0.8}{
        \begin{tabular}{cc|c || cc | c}
            \toprule[0.2em]
            Hyperparam. & Value & \%-MSE & Hyperparam. & Value & \%-MSE  \\
            \toprule[0.2em]
             & 10 & 2.81 & & 10 & 2.96 \\
            Frame Gap & 20 & \textbf{2.57} & \# keypoints & 16 & \textbf{2.57}\\
             & 30 & 2.64 & & 30 & 2.63 \\
            \bottomrule[0.1em]
        \end{tabular}
        }
    \end{center}\vspace{-0.3cm}
    \caption{\textbf{Hyperparameters Study on Simplified Human 3.6M}. \%-MSE error from a single run is shown. For frame gap experiments, the number of keypoints is set to 16. Frame gap is set to 20 for experiments with a varying number of keypoints. We use frame difference here as a reconstruction target for studying the effect of hyperparameters.}
    \label{tab:ablation_KT_h36m}\vspace{-0.3cm}
\end{table}

We evaluate the effect of number of keypoints and frame gaps on simplified Human 3.6M  (Table~\ref{tab:ablation_KT_h36m}). 
Note that we use frame difference, instead of SSIM, as a reconstruction target for studying the effect of hyperparameters. When the frame gap is too small, the region of motion becomes too narrow, which results in slightly lower performance. 
Also, discovering more keypoints does not always guarantee better performance. Empirical results show that informative keypoints are discoverable with 16 keypoints.

We also perform an ablation study using a single geometry branch at time $t+k$ for two different reconstruction targets (image and SSIM), using the same \%-MSE error metric as the main paper. Training with one geometry branch reduced the pose regression performance from 2.534 $\pm$ 0.056 to 2.596 $\pm$ 0.1089 (image) and 2.556 $\pm$ 0.0320 (SSIM) where the standard deviation is computed over 5 runs. As a loss ablation study, we train our model without the rotation equivariance loss on simplified Human 3.6M, and the pose regression performance is reduced to 2.61.

\subsection{Jellyfish Pulse Detection}~\label{sec:pulse}

\begin{figure}
    \centering
    \includegraphics[width=0.7\linewidth]{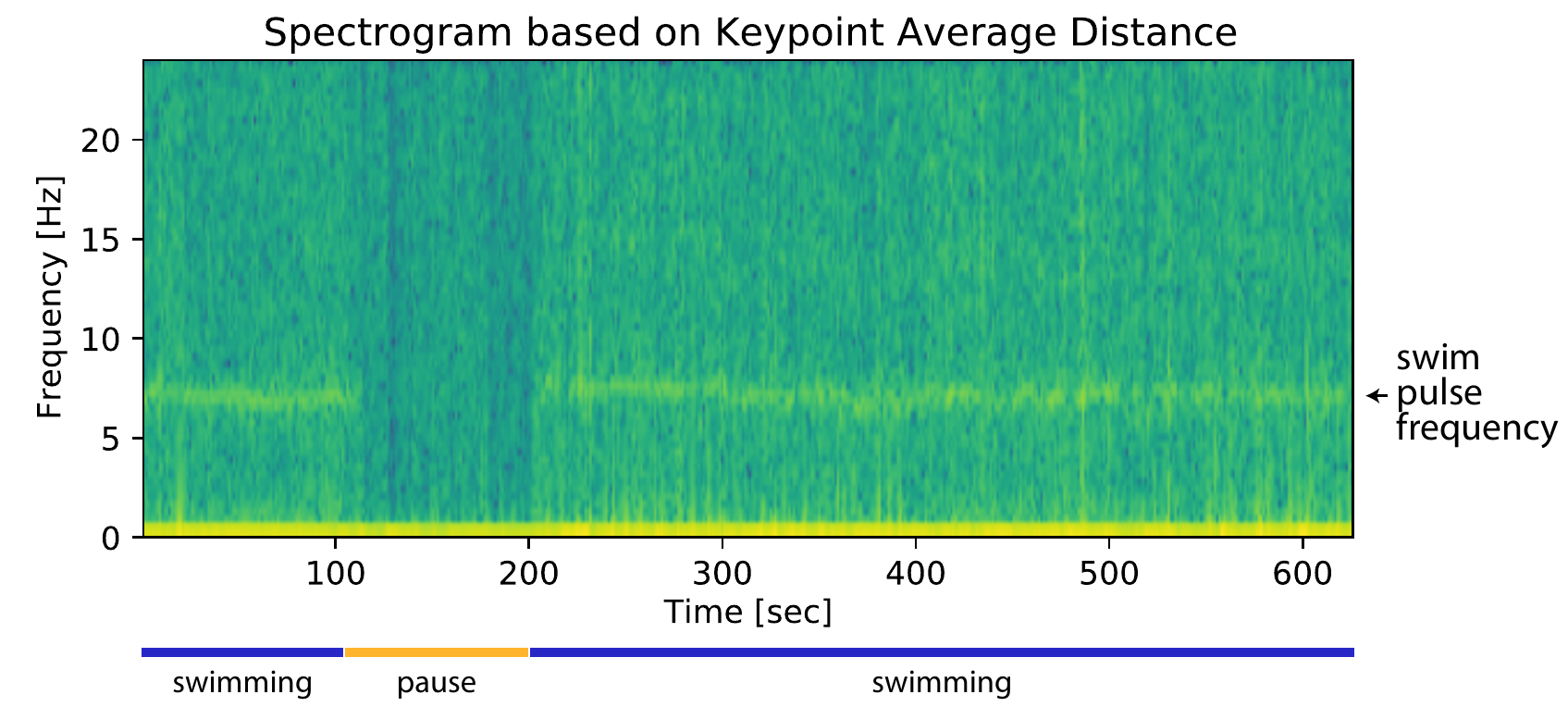}
    \caption{\textbf{Spectrogram from Distance of Discovered Keypoints}. From a recorded video of jellyfish swimming at 48Hz, we discover keypoints at each frame using our model and compute a spectrogram based on the average distance between discovered keypoints on the jellyfish.}
    \label{fig:spectrogram}
\end{figure}

The energy efficiency of swimming jellyfish combined with their structural simplicity makes them a good organism for understanding the hydrodynamics of animal propulsion~\cite{costello2021hydrodynamics}.
In particular, researchers would like to study the relationship between body plan and swim pulse frequency across jellyfish species.
This has applications in ethology, hydrodynamics, as well as bio-inspired vehicles.
Here, we use Clytia hemisphaerica as our jellyfish species to study jellyfish pulsing during swimming using our discovered keypoints.
After videos are recorded from a swimming jellyfish from in a tank, we apply our keypoint discovery model to track keypoints automatically on the jellyfish (visualization provided in project website). We also compute the swim pulse frequency by computing the distance between all pairs of our discovered keypoints with high confidence (5 keypoints) and extracting a frequency spectrogram based on average keypoint distance (Figure~\ref{fig:spectrogram}). We observe a visible band at the swimming frequency around $7$Hz, and we note that between $110$ to $200$ seconds, the jellyfish is not swimming (floating), and thus the swimming frequency band is not visible in that duration.
Since our discovered keypoints are able to detect pulsing, this provides a way to automatically annotate swimming behavior. 
This method can be applied to videos from other jellyfish species to study the relationship between swimming dynamics and body plan.

\subsection{Vegetations Wind Speed Regression}~\label{sec:tree}

\begin{figure}
    \centering
    \includegraphics[width=0.5\linewidth]{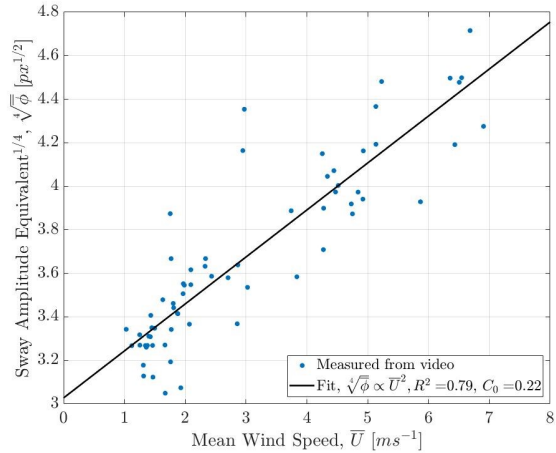}
    \caption{\textbf{Wind Speed Regression from Discovered Keypoints}. Mean wind speed, $\Bar{U}$, vs. the fourth root of the sway amplitude equivalent measured from the standard deviation of the convex hull area of the 15 discovered keypoints in each clip, based on model from~\cite{cardona2021wind}. The scatter represents 10-mintute averages of the same data used for training the keypoint model. The black lines represent the best linear regression fit for the proportionality assumption. The proportionality coefficient and the $R^2$ values are presented in the legend.  }
    \label{fig:wind_regression}
\end{figure}

Videos of oscillation of tree branches and leaves encode information on local wind conditions, and could function as wind speed sensors. Local wind speed measurements are useful for a variety of tasks, including air pollution monitoring, weather forecasting, and predicting movement of forest fires~\cite{NEURIPS2019_a9ad5f28,cardona2021wind}. 
We use the Vegetation dataset to study the effectiveness of our discovered keypoints for capturing oscillating movement of trees.
This dataset consists of videos of swaying trees recorded from an overhead camera from a drone, while the wind speed is measured using an anemometer.
We observe that the discovered keypoints from our approach are of different parts of the tree in separate views but are consistent within a single clip, as to capture oscillations of branches/leaves (visualization provided in project website).

We use a physics-based model~\cite{cardona2021wind} to study the relationship between oscillations of trees and wind speed. This model defines the relationship between structural oscillation and wind speed as:
$$    \sigma \sim I_u \Bar{U} $$
where $\sigma$ is the standard deviation of the amplitude of the structural oscillations, $\Bar{U}$ is the mean wind speed, and $I_u$ is the measure of the turbulence intensity of the streamwise component, defined as the standard deviation of the streamwise velocity fluctuations normalized to the mean wind speed. The model requires tracing of the structural oscillations of the branches/leaves, which was previously done manually and we show that the keypoint discovery model can do this automatically. The 15 detected keypoints track these oscillations in a 2D space and a representative measure of these oscillations in both coordinates is calculated using the convex hull area, or the sway amplitude equivalent, $\phi$. 
The average sway amplitude equivalent of the keypoints, $\Bar{\phi}$, provides the following proportionality relationship:
$$C_0 \sqrt{\Bar{\phi}} \sim \Bar{U}$$
where $C_0$ is the coefficient of proportionality. The best regression fit of the experimental data calculated using the least squares method has $R^2 = 0.79$ suggesting there is a good agreement between the proportionality assumption and the experimental results using the keypoint detection model (Figure~\ref{fig:wind_regression}).

\section{Additional Implementation Details}~\label{sec:additional_implementation}

\textbf{Architecture Details} 
Our method uses ResNet-50~\cite{He2016DeepRL} as an encoder $\Phi$, GlobalNet~\cite{CPN17} as a pose decoder $\Psi$, and a series of convolution blocks as a reconstruction decoder $\psi$, following the unsupervised keypoint discovery model from~\cite{ryou2021weakly}. Architecture details about reconstruction decoder is shown in Table~\ref{tab:decoder_details}. For more implementation details, the code is available on our project website: \url{https://sites.google.com/view/b-kind}.

\begin{table}
    \caption{\textbf{Architecture details of the reconstruction decoder.} ``Conv\_block" refers to a basic convolution block which is composed of 3$\times$3 convolution, batch normalization, and ReLU activation. Note that output size for Human3.6M experiments is downsampled by a factor of 2 for all the layers. }
    \label{tab:decoder_details}
    \begin{center}
        \begin{tabular}{lccc}
        \toprule
            Type & Input dimension & Output dimension & Output size  \\
            \midrule
            Upsampling & - & - &  16x16 \\
            Conv\_block & 2048 + \# keypoints $\times$ 2 & 1024 & 16x16 \\
            Upsampling & - & - & 32x32 \\
            Conv\_block & 1024 + \# keypoints $\times$ 2 & 512 & 32x32  \\
            Upsampling & - & - & 64x64 \\
            Conv\_block & 512 + \# keypoints $\times$ 2 & 256 & 64x64  \\
            Upsampling & - & - & 128x128 \\
            Conv\_block & 256 + \# keypoints $\times$ 2 & 128 & 128x128  \\
            Upsampling & - & - & 256x256 \\
            Conv\_block & 128 + \# keypoints $\times$ 2 & 64 & 256x256 \\
            Convolution & 64 & 3 & 256x256 \\
            \bottomrule\\[-1em]
        \end{tabular}
    \end{center}
\end{table}

The hyperparameters for the keypoint discovery model is included in Table~\ref{tab:hyperparameter_1}. All models use SSIM image as the reconstruction target, unless stated otherwise. All keypoint discovery models are trained until convergence of the training loss on a NVIDIA V100 Tensor Core GPU. Below, we include a additional details on the keypoint discovery model and downstream task used to evaluate each dataset. 

\textbf{CalMS21}. The CalMS21 dataset~\cite{sun2021multi} consists of videos and trajectory data from a pair of interacting mice, annotated with behavior labels at each frame by neuroscientists. There is one black mouse and one white mouse engaging in social behaviors, recorded at $1024 \times 570$ at 30 Hz. The supervised keypoints provided with CalMS21 are from the MARS detector~\cite{segalin2020mouse} developed for this dataset, which detects 7 anatomically-defined keypoints for each mouse. For training keypoint discovery, we use a subset of the training split without miniscope cable (26k images), and we use the full train/test split defined by \cite{sun2021multi} on Task 1 for evaluating behavior classification. For behavior classification, we use the same setup (1D Conv Net architecture, hyperparameters, random seeds, data split, etc.) as the CalMS21 dataset benchmarks, except we replace the supervised input keypoints with our discovered keypoints for evaluation. We additionally experiment with adding heatmap confidence and convariance during classification by appending these additional features to input keypoints during classifier training. This dataset is available under the CC-BY-NC-SA license.

\textbf{MARS-Pose}. MARS-Pose is a set of mouse interaction images with human keypoint annotations~\cite{segalin2020mouse} and these images are recorded in similar recording conditions to CalMS21~\cite{sun2021multi}. We use a subset of the images for training (10,50,100,500) and test on the full 1.5k images test set. We evaluate this dataset based on pose estimation performance to the human-annotated keypoints. For the supervised model, we use the stacked hourglass model~\cite{Newell2016StackedHN} and for the semi-supervised model, we add a supervised keypoint estimation loss based on MSE to our keypoint discovery framework. 

\textbf{Fly vs. Fly}. This dataset consists of videos of two interacting flies~\cite{eyjolfsdottir2014detecting} with frame-level behavior annotations. We use the ``Aggression" videos from this dataset ($144 \times 144$ at 30 Hz) and use the behaviors with more than 1000 annotated training samples, with the same setup as~\cite{sun2021task}. The provided FlyTracker with this dataset computes hand-crafted behavioral features directly from video for behavior classification. Since keypoints may be discovered from any body part, we compute corresponding generic features not based on keypoint identity: speed of every keypoint, acceleration of every keypoint, distance between every pair, and angle between every triplet. Additionally, since the flies are similar in appearance, when extracting keypoint locations from the B-KinD heatmaps, we detect 2 max locations for the 2 peaks. We then take the spatial softmax over the region around each max location, instead of taking the spatial softmax over the whole heatmap. In terms of identity, we always use the fly with smaller y values at centroid as the first fly, and the fly with larger y values as the second. For the classifier model, we use the same setup (1D Conv Net architecture (except frame gap in the Conv Net is 1 instead of 2 since flies have faster behaviors), hyperparameters, random seeds, data split, etc.) as the CalMS21 dataset benchmarks, except using the fly features as input to classify annotated behavior at each frame. This dataset is available under the CC0 1.0 Universal license.

\textbf{Human3.6M}.
Human 3.6M dataset~\cite{ionescu2013human3} is a large-scale dataset containing 3.6 million 3D and 2D human poses with corresponding images. The videos are taken from 4 different viewpoints for 17 scenarios (discussion, taking photo, walking, ...) with the same background. This dataset is available for academic use, and the dataset license is provided by the Human 3.6M authors on the dataset website, link available within~\cite{ionescu2013human3}. Simplified Human 3.6M dataset, introduced by~\cite{ZhangKptDisc18}, consists of 6 different activities with mostly upright poses by cropping the full image using bounding box. Since our method requires static background assumption, we crop a pair of full images using the same bounding box for training a keypoint discovery model. The final image has 128$\times$128 resolution. We evaluate the pose regression performance on the same testing set from the Simplified Human 3.6M dataset.

\textbf{Jellyfish}. This is an in-house video dataset consisting of a freely swimming Clytia hemisphaerica in a water tank. We train and run our keypoint discovery model on the same 30k frames, recorded at 48Hz, to demonstrate our keypoints on new organisms and on detecting swimming frequency. Since the jellyfish is very small ($\sim 50$ pix) relative to the size of the image ($928 \times 1158$), we first use the SSIM image to identify a rough bounding box around the jellyfish ($150 \times 150$) before re-scaling the input to the keypoint discovery model to $256 \times 256$. We note that this step would not be necessary given a GPU with more memory, since the jellyfish would still be visible at higher resolutions. More details on the pulse detection is in Section~\ref{sec:pulse}.

\textbf{Vegetations}. This is an in-house video dataset captured from a drone flying overhead of an Oak tree as the tree is swaying in the wind, and local wind speed is recorded using an anemometer. The video frames are processed at $512 \times 512$ and 120 Hz, and re-scaled to be $256 \times 256$ for the keypoint discovery model. The drone may shift slightly over the video recording, and we use existing image alignment methods~\cite{thevenaz1998stackreg} to align video frames before computing the spatiotemporal difference reconstruction target for our method. More details on the wind speed regression is in Section~\ref{sec:tree}.

\begin{table}[!t]
  \centering
  \small
  \scalebox{1.0}{
   \begin{tabular}{c | c | c| c | c| c  } 
   \hline
   Dataset & \# Keypoints & Batch size & Resolution & Frame Gap & Learning Rate  \\
   \hline
    CalMS21 & 10 &  5 & 256 & 6 & 0.001 \\
    \hline
    Fly & 10 & 5 & 256 & 3 & 0.001\\
   \hline
    Human & 16 & 36 & 128 & 20 & 0.001\\
   \hline
    Jellyfish & 10 & 5 & 256 & 20 & 0.001\\
   \hline
    Vegetations & 15 & 5 & 256 & 60 & 0.001\\
   \hline   
\end{tabular}
}
\caption{{\bf Hyperparameters for Keypoint Discovery.}} \label{tab:hyperparameter_1}
\end{table}

\section{Visualizations}~\label{sec:visualizations}

We present additional visualization results on mouse (Figure~\ref{fig:mouse_qual}), fly (Figure~\ref{fig:fly_qual}), tree (Figure~\ref{fig:tree_qual}), and human (Figure~\ref{fig:human_qual}).
Additional videos are available on our project website: \url{https://sites.google.com/view/b-kind}.

\begin{figure}
    \centering
    \includegraphics[width=0.9\linewidth]{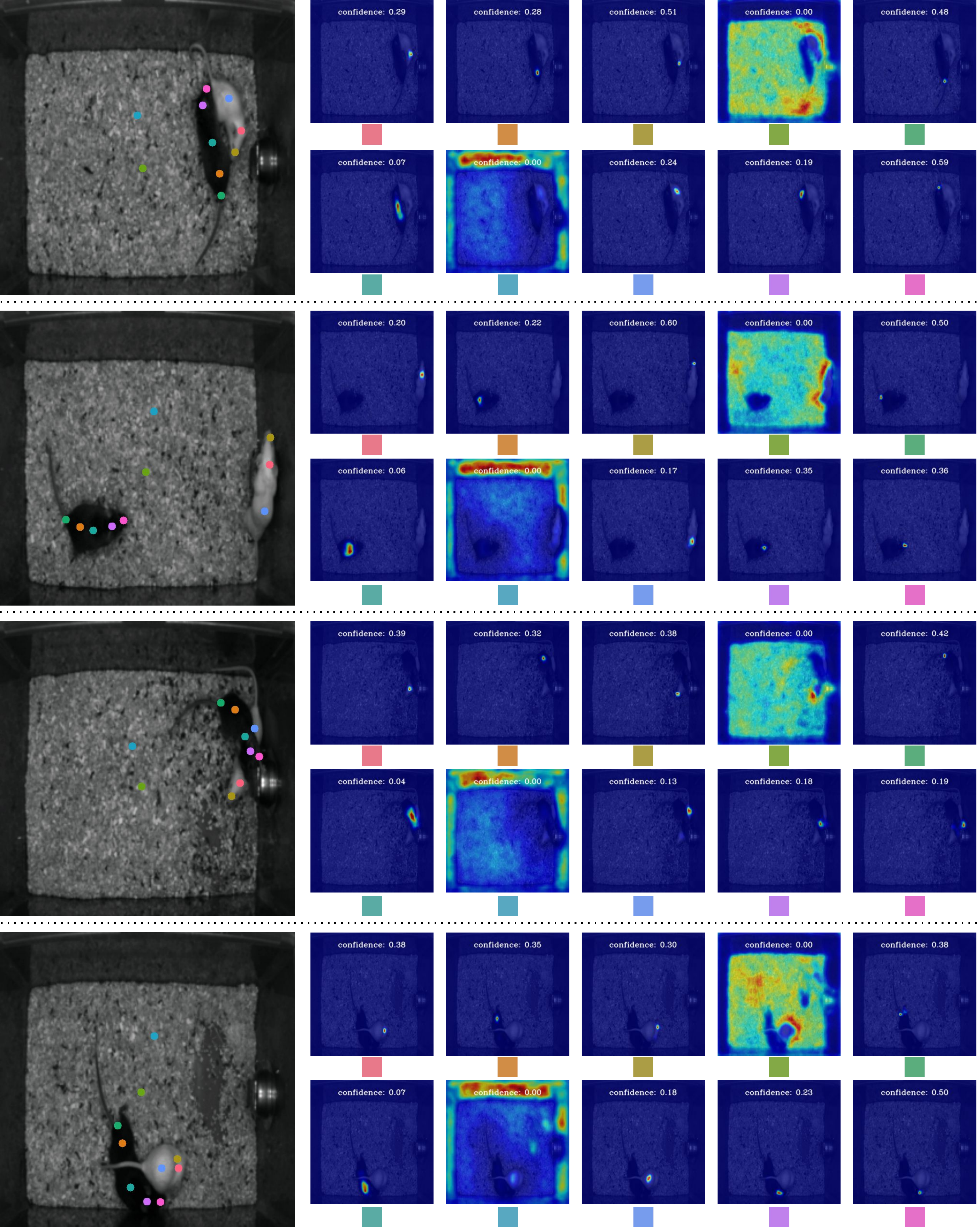}
    \caption{\textbf{Confidence visualization on CalMS21}. Confidence score (maximum prediction value) is shown with the normalized heatmap. Background keypoints (fourth on row 1 and second on row 2) have very low confidence.}
    \label{fig:mouse_confidence}
\end{figure}

\textbf{Confidence Visualizations}. We observe that keypoints discovered on the background and not tracking agent parts generally have very low confidence (Figure~\ref{fig:mouse_confidence}).
This is because heatmaps of background keypoints are not well-localized, and is spread over the image, thus have a low peak value (low confidence). In comparison, discovered keypoints on body parts (such as the nose), is localized to a specific part of the image and has higher peak values. Additionally, confidence values can provide information on occluded parts. For example, for the nose of the white mouse (third column, first row, Figure~\ref{fig:mouse_confidence}), the confidence varies from $0.5 \sim 0.6$ when the nose is visible in the first two examples to $0.3 \sim 0.4$ when the nose is harder to see in the last two examples.

\textbf{Challenges}. Difficult examples for our model are visualized in Figure~\ref{fig:failure_qual}. When there is occlusion, such as in the mouse examples, the keypoint is generally discovered on the visible parts, and when there is heavy occlusion, such as from the miniscope cable, discovered keypoint location may be shifted. This is likely why including additional information from the heatmap, such as confidence (Figure~\ref{fig:mouse_confidence}) is helpful for behavior classification. We can see similar effects on self-occlusion for humans, and also left-right swapping of some keypoints for when humans are facing towards or away from the camera (this has also been observed with other keypoint discovery models~\cite{ZhangKptDisc18,Lorenz19,SchmidtkeVELAK21}). Unusual poses may also be difficult, such as when the fly is completely tilted towards the camera in the last column of row 1. Future directions to integrate 3D structure, for instance by using multi-view videos, could help address these issues. 
Despite this, we note that our current discovery model achieves state-of-the-art results among other self-supervised methods for behavior classification and keypoint regression.


\def\figsize{0.23}
\def\fighspace{-2mm}
\def\fighspacer{-2mm}
\begin{figure*}[h]
\centering
\begin{tabular}{cccc}
\centering
\hspace{\fighspace}\includegraphics[width=\figsize\textwidth]{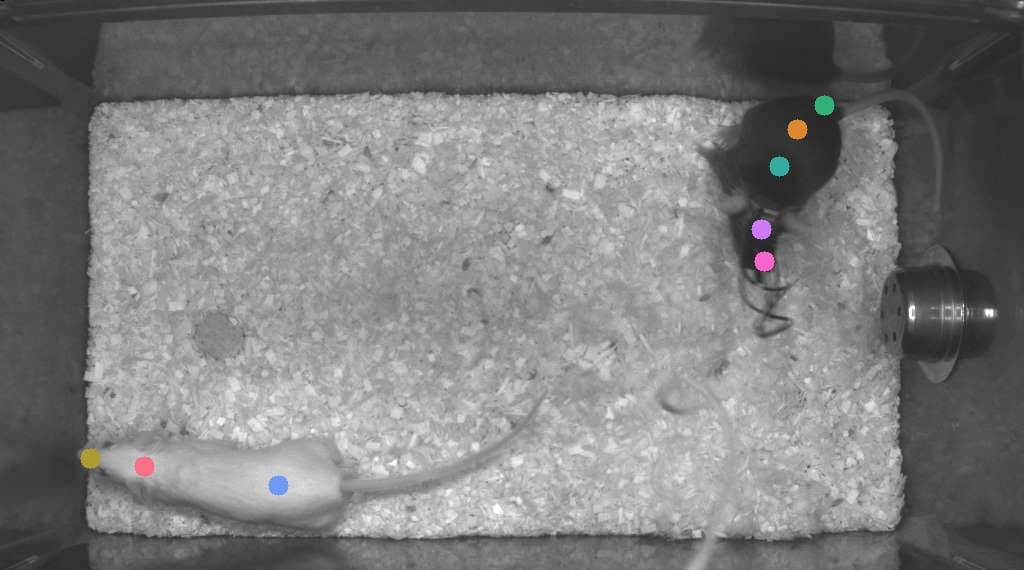}\hspace{\fighspace} & \includegraphics[width=\figsize\textwidth]{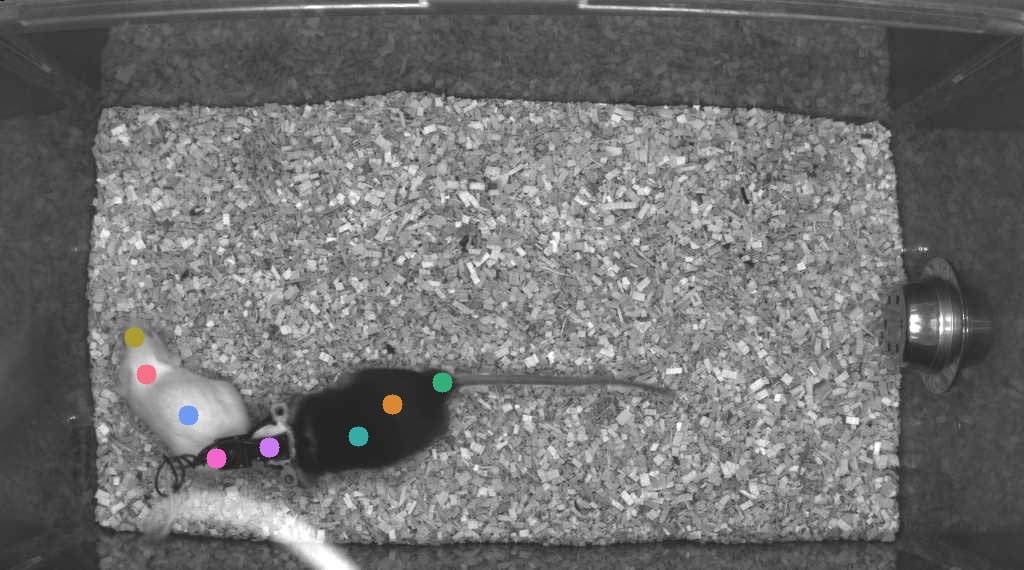}\hspace{\fighspacer} & \includegraphics[width=\figsize\textwidth]{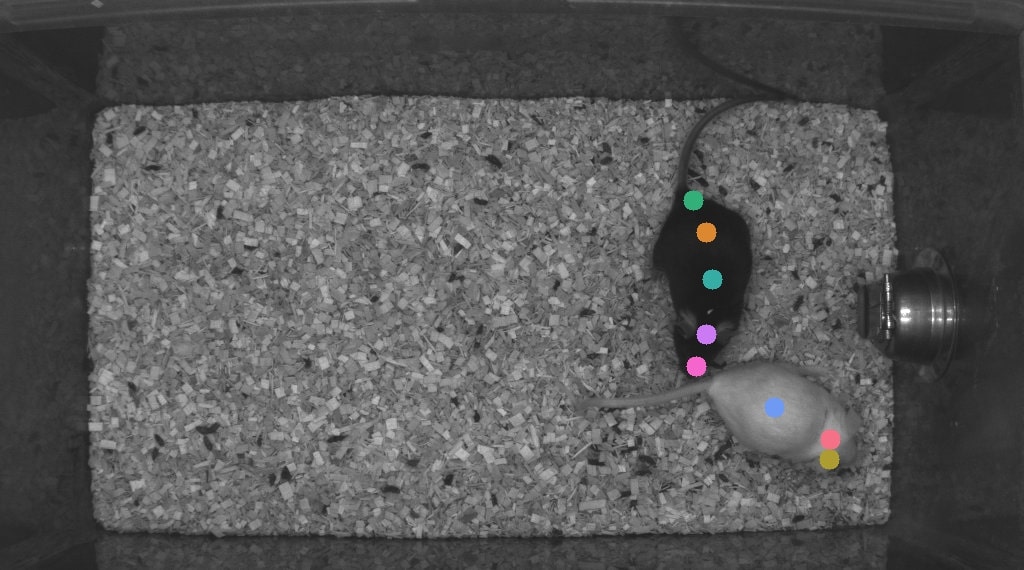}\hspace{\fighspace} & 
\includegraphics[width=\figsize\textwidth]{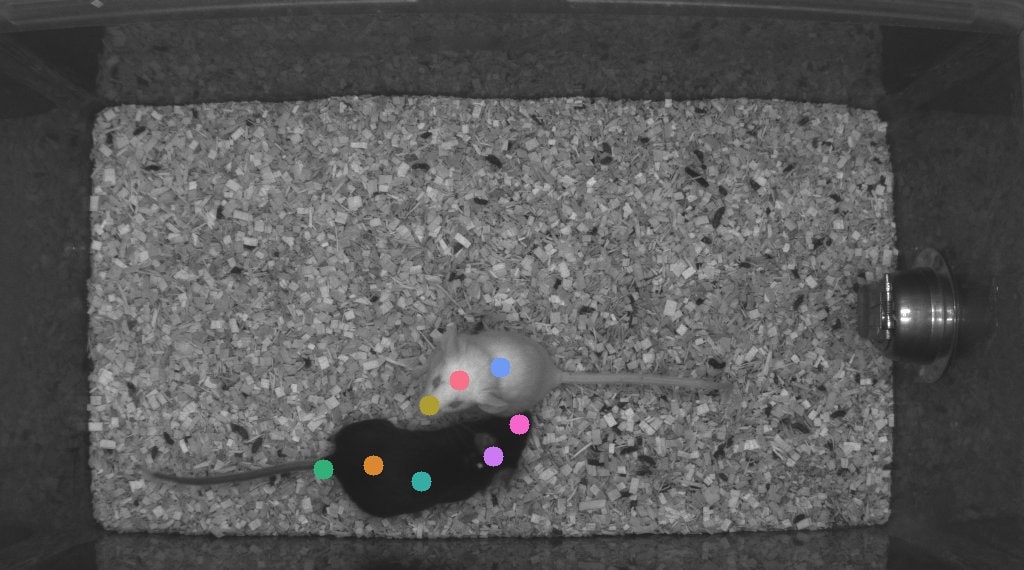}\hspace{\fighspace}  \\
\hspace{\fighspace}\includegraphics[width=\figsize\textwidth]{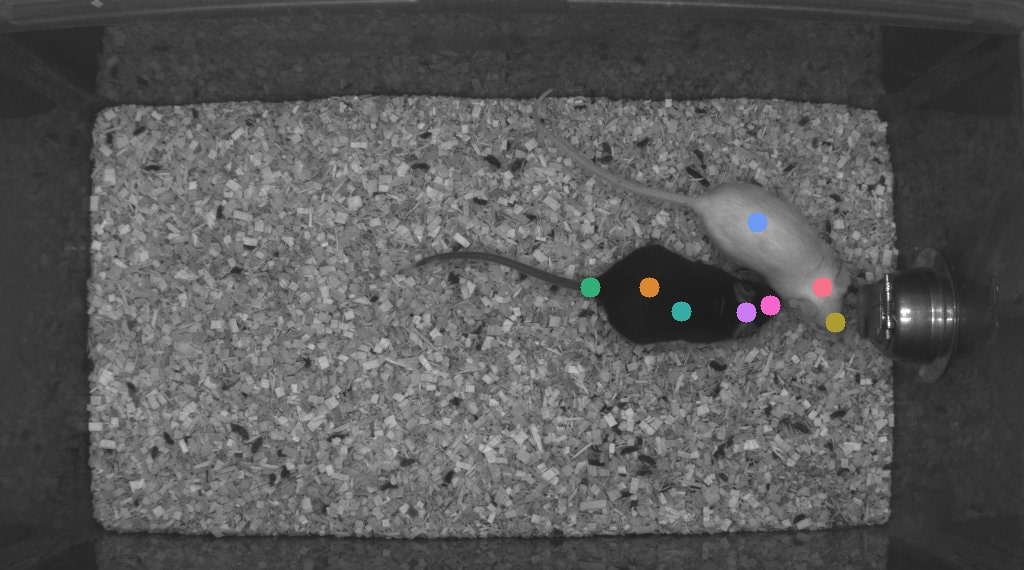}\hspace{\fighspace} & \includegraphics[width=\figsize\textwidth]{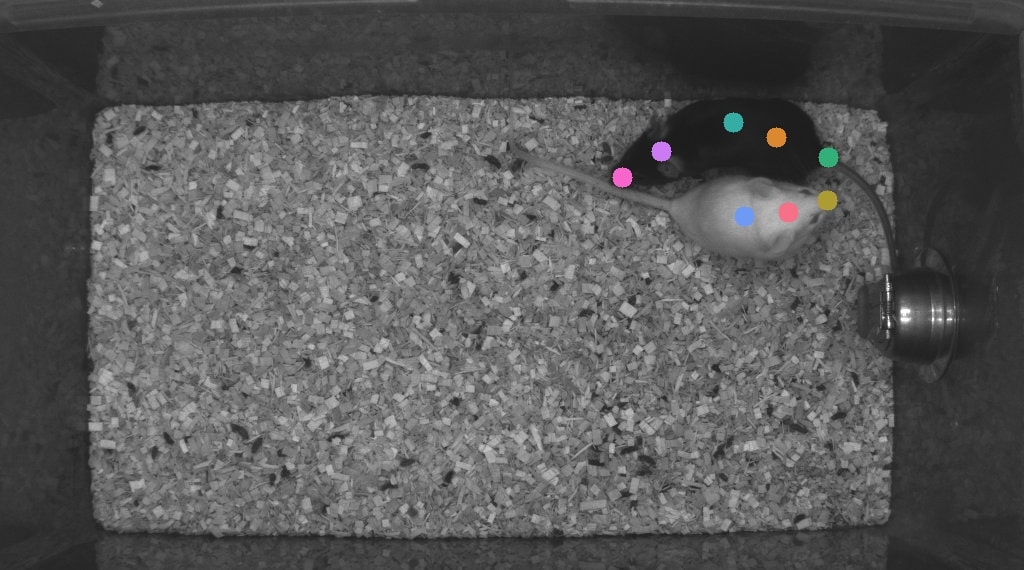}\hspace{\fighspacer} & \includegraphics[width=\figsize\textwidth]{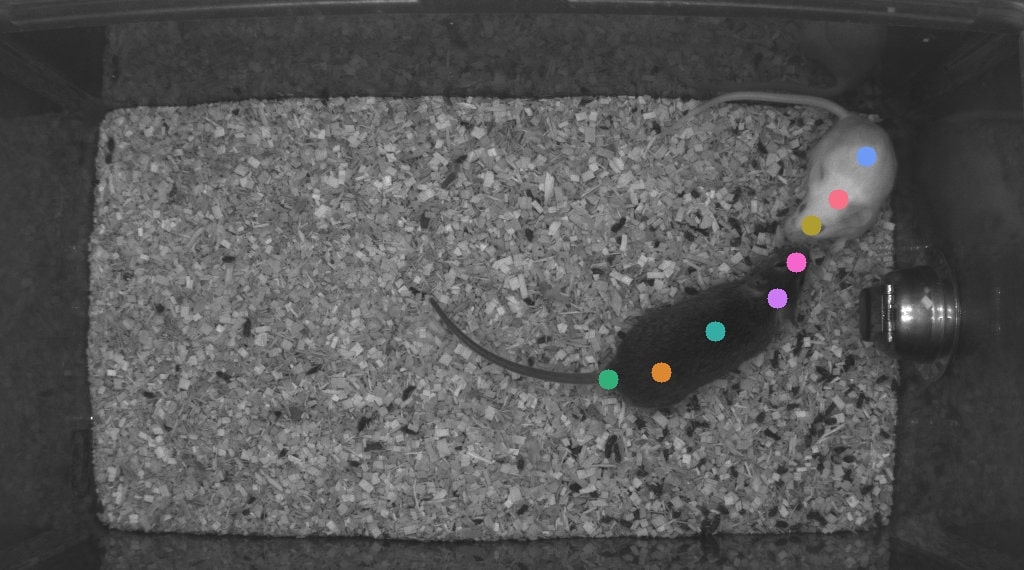}\hspace{\fighspace} & \includegraphics[width=\figsize\textwidth]{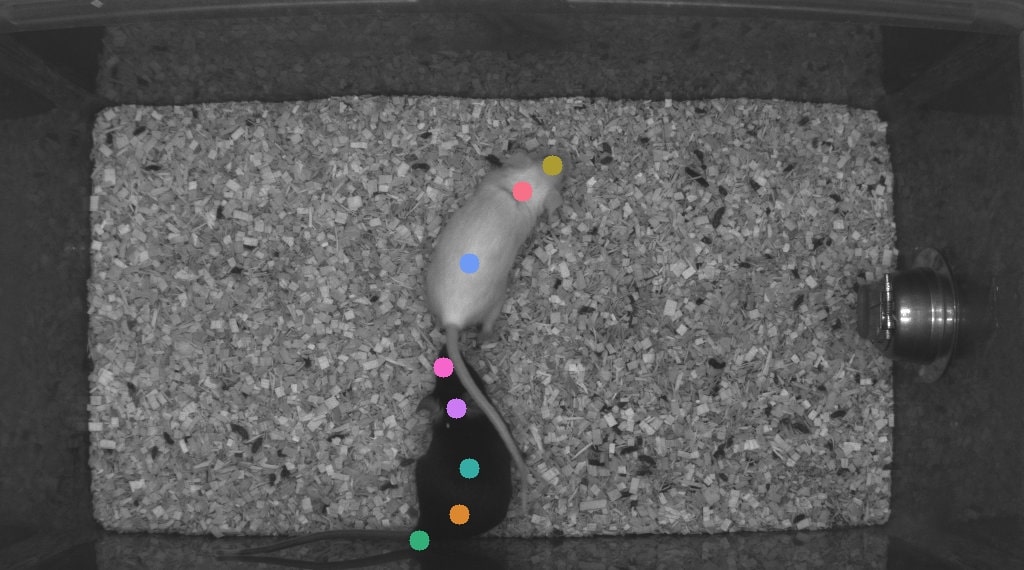}\hspace{\fighspacer} \\
\end{tabular}
\caption{\textbf{Qualitative Results on CalMS21}. We observe that keypoints are discovered for noses of both mice and generally along the spine of the mice.}
\label{fig:mouse_qual}
\end{figure*}

\def\figsize{0.15}
\def\fighspace{-2mm}
\def\fighspacer{-2mm}
\begin{figure*}[h]
\centering
\begin{tabular}{cccccc}
\centering
\hspace{\fighspace}\includegraphics[width=\figsize\textwidth]{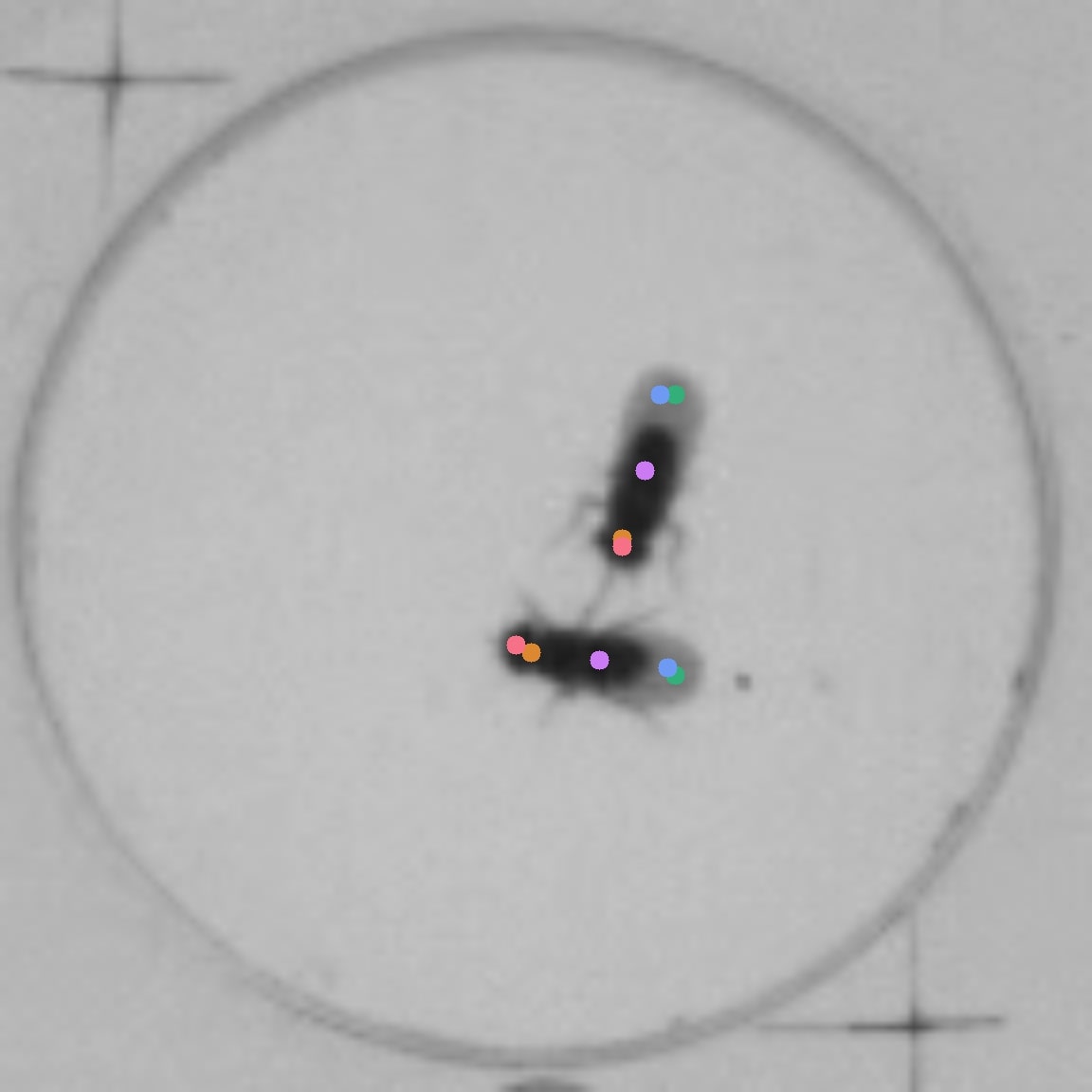}\hspace{\fighspace} & \includegraphics[width=\figsize\textwidth]{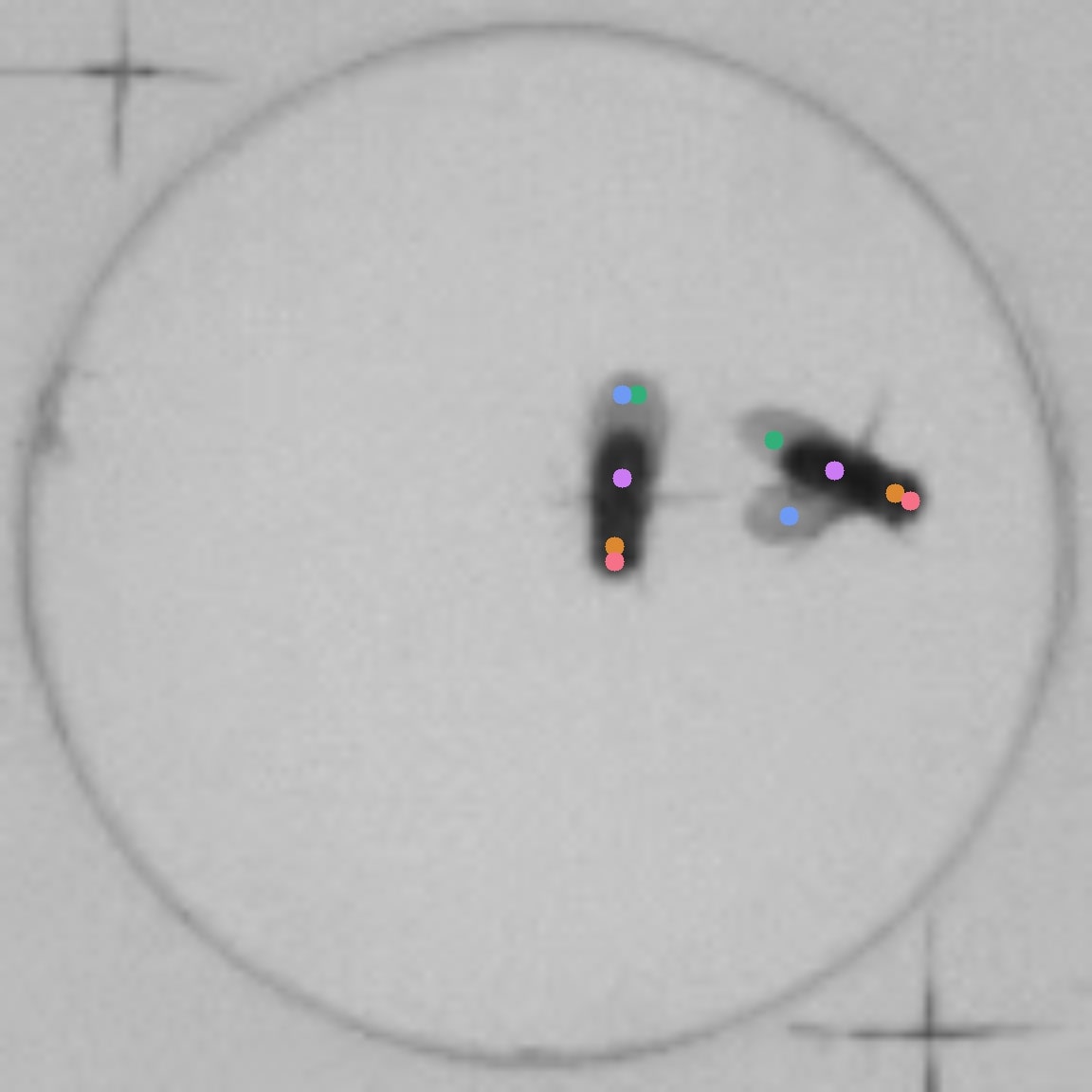}\hspace{\fighspacer} & \includegraphics[width=\figsize\textwidth]{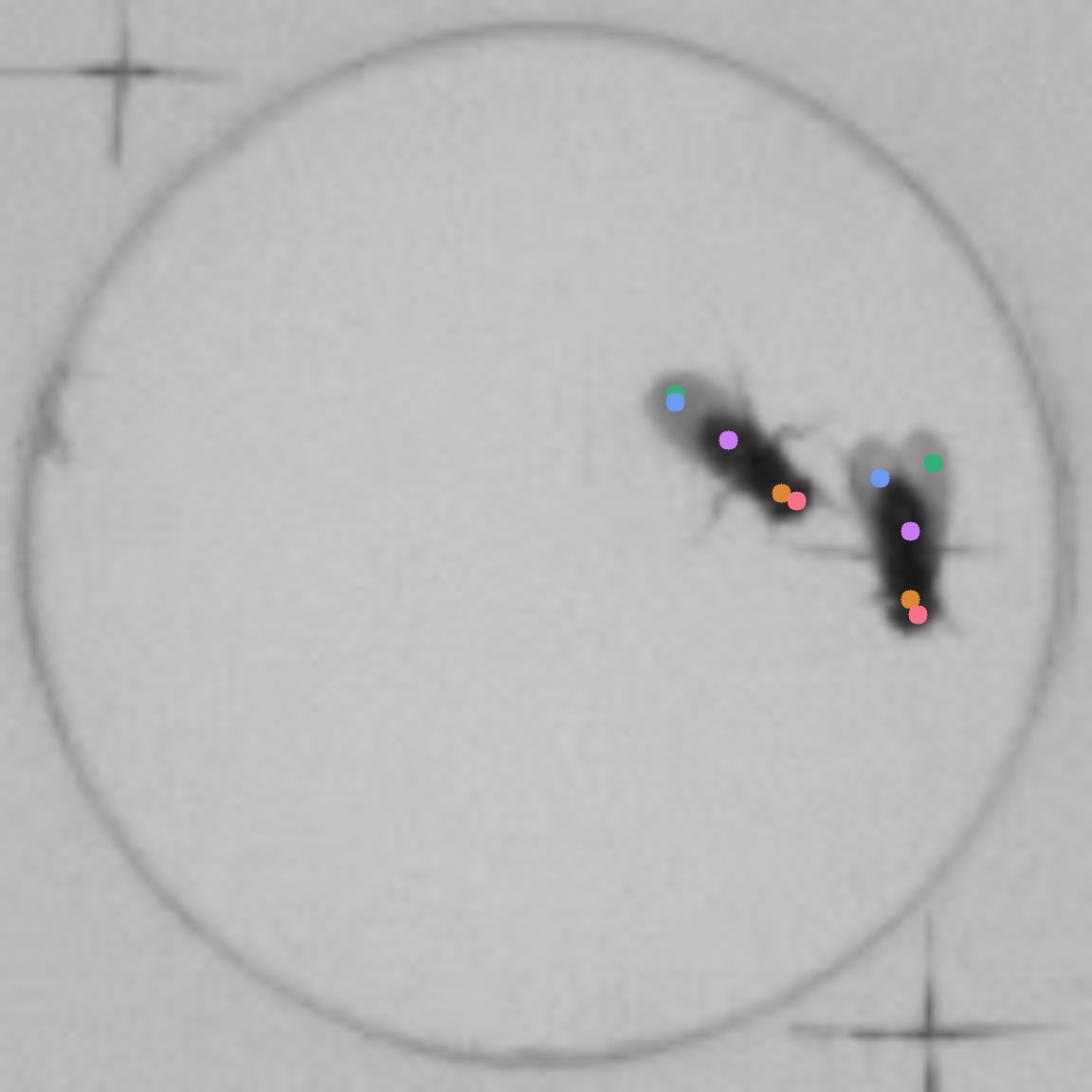}\hspace{\fighspace} & 
\includegraphics[width=\figsize\textwidth]{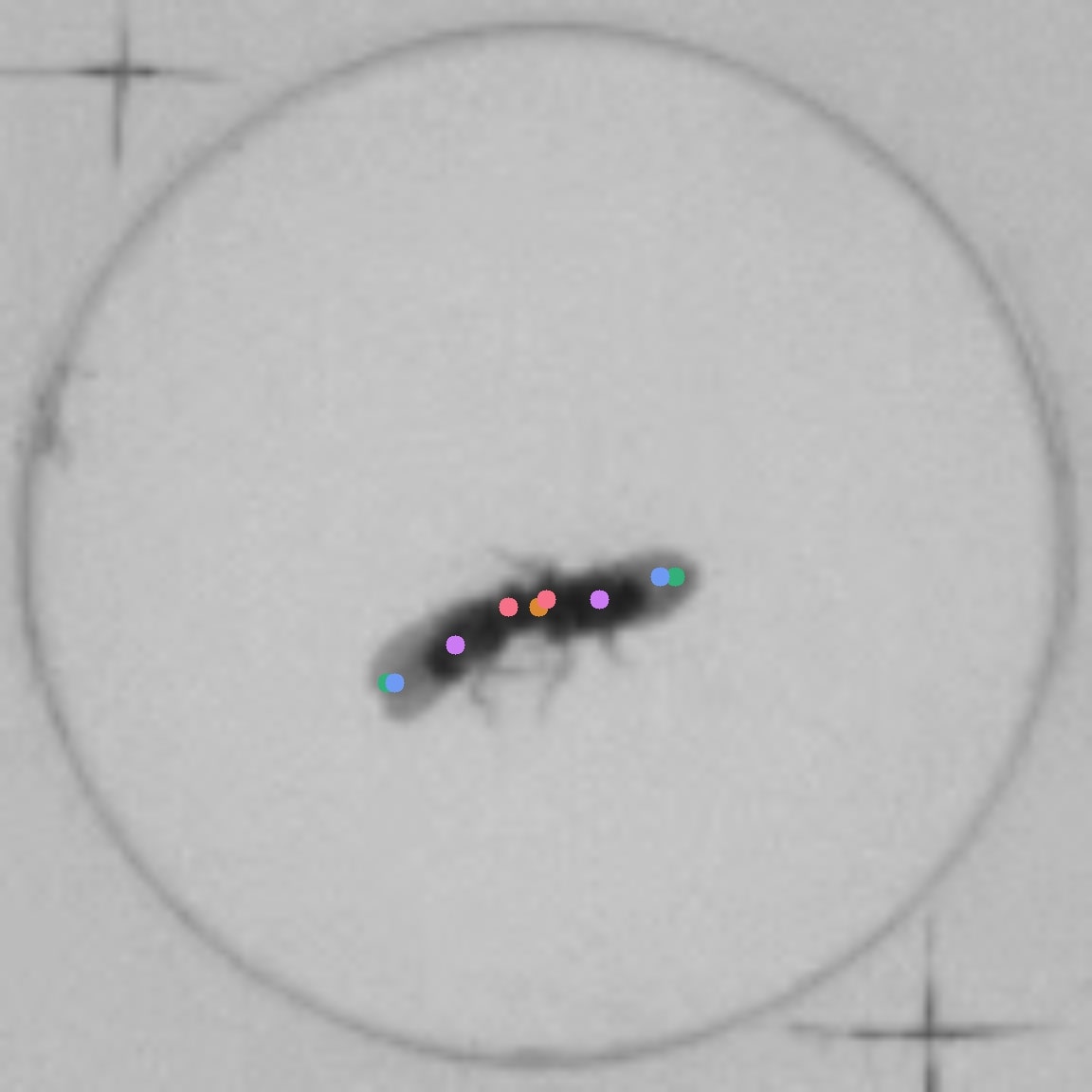}\hspace{\fighspace} & \includegraphics[width=\figsize\textwidth]{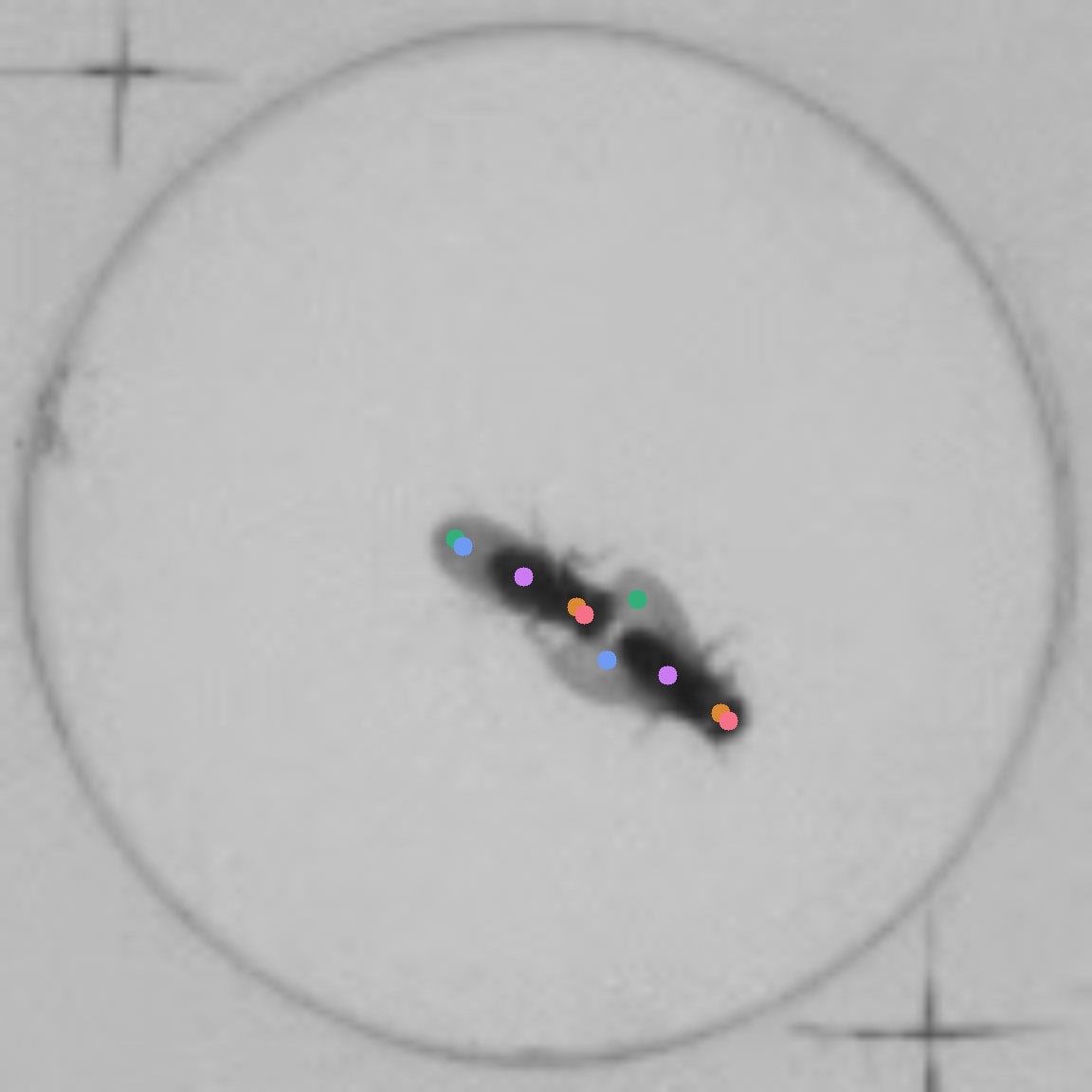}\hspace{\fighspacer} & \includegraphics[width=\figsize\textwidth]{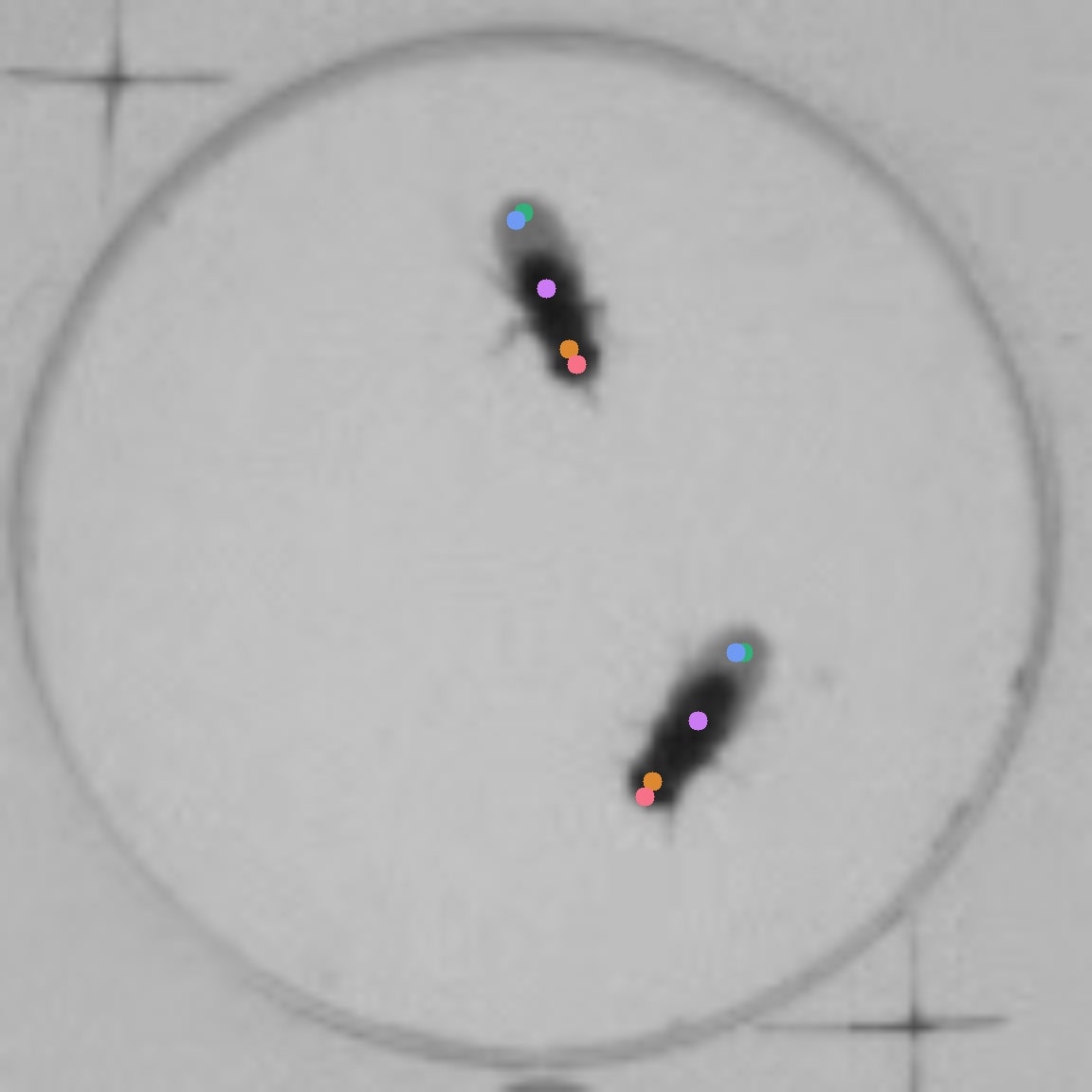}\hspace{\fighspacer}  \\
\hspace{\fighspace}\includegraphics[width=\figsize\textwidth]{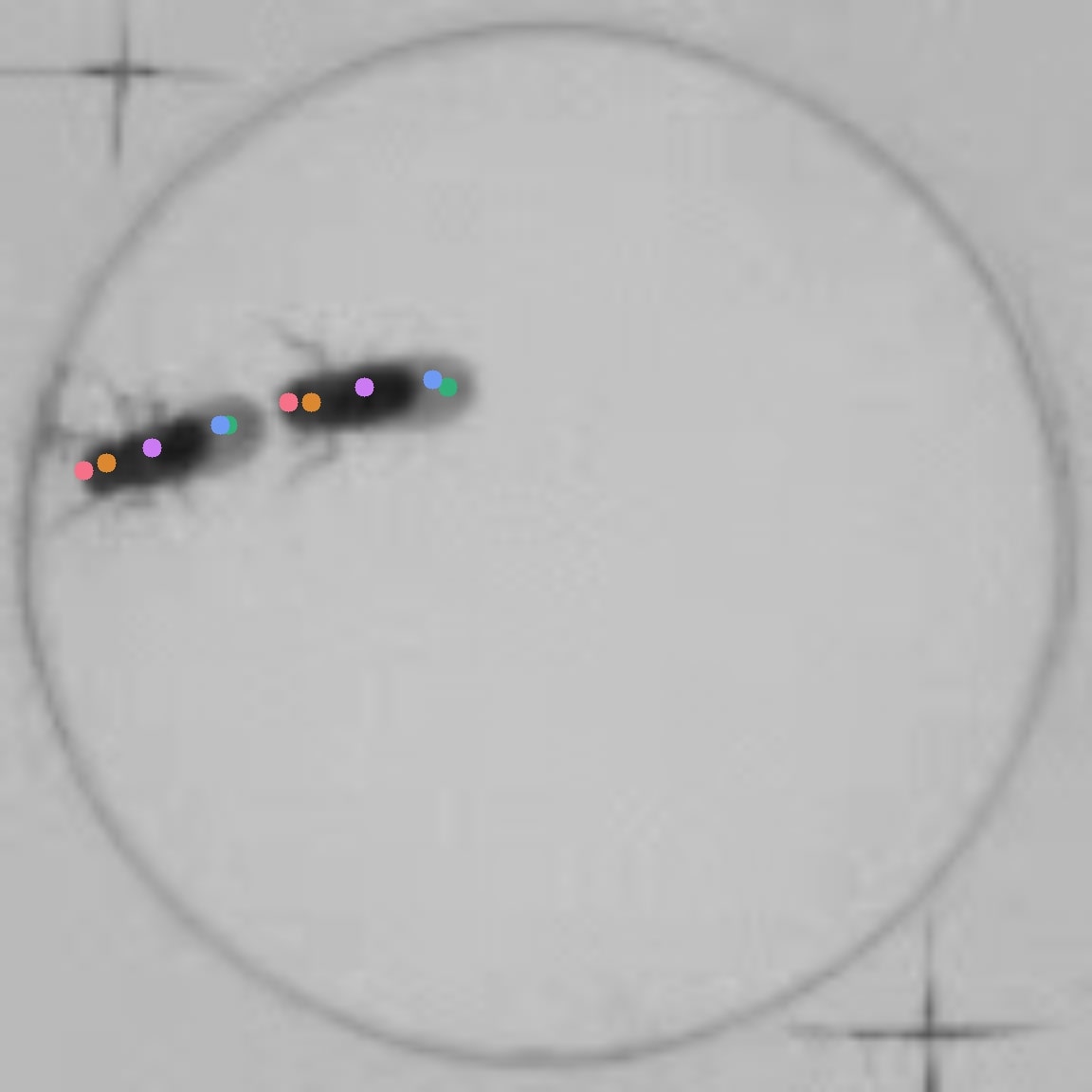}\hspace{\fighspace} & \includegraphics[width=\figsize\textwidth]{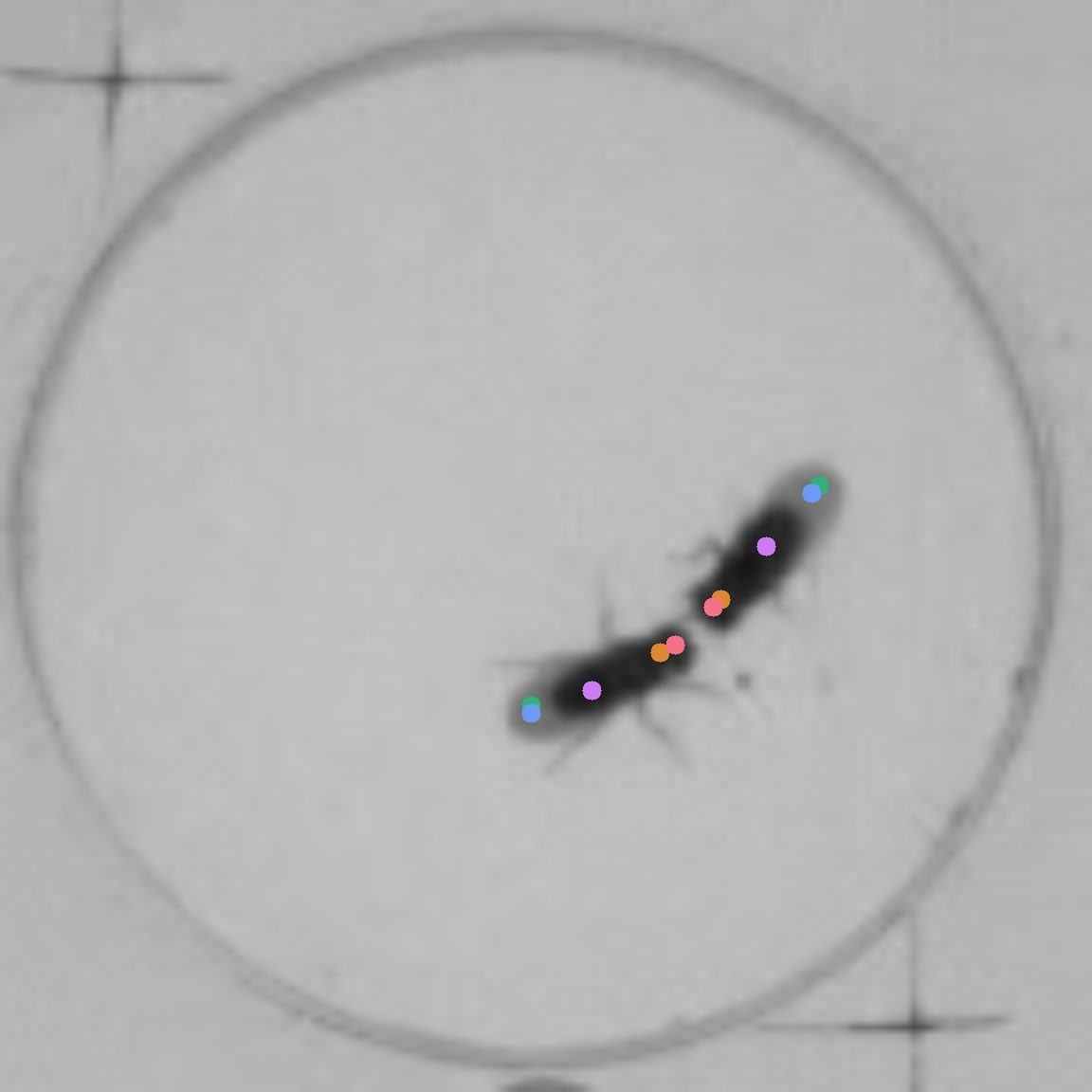}\hspace{\fighspacer} & \includegraphics[width=\figsize\textwidth]{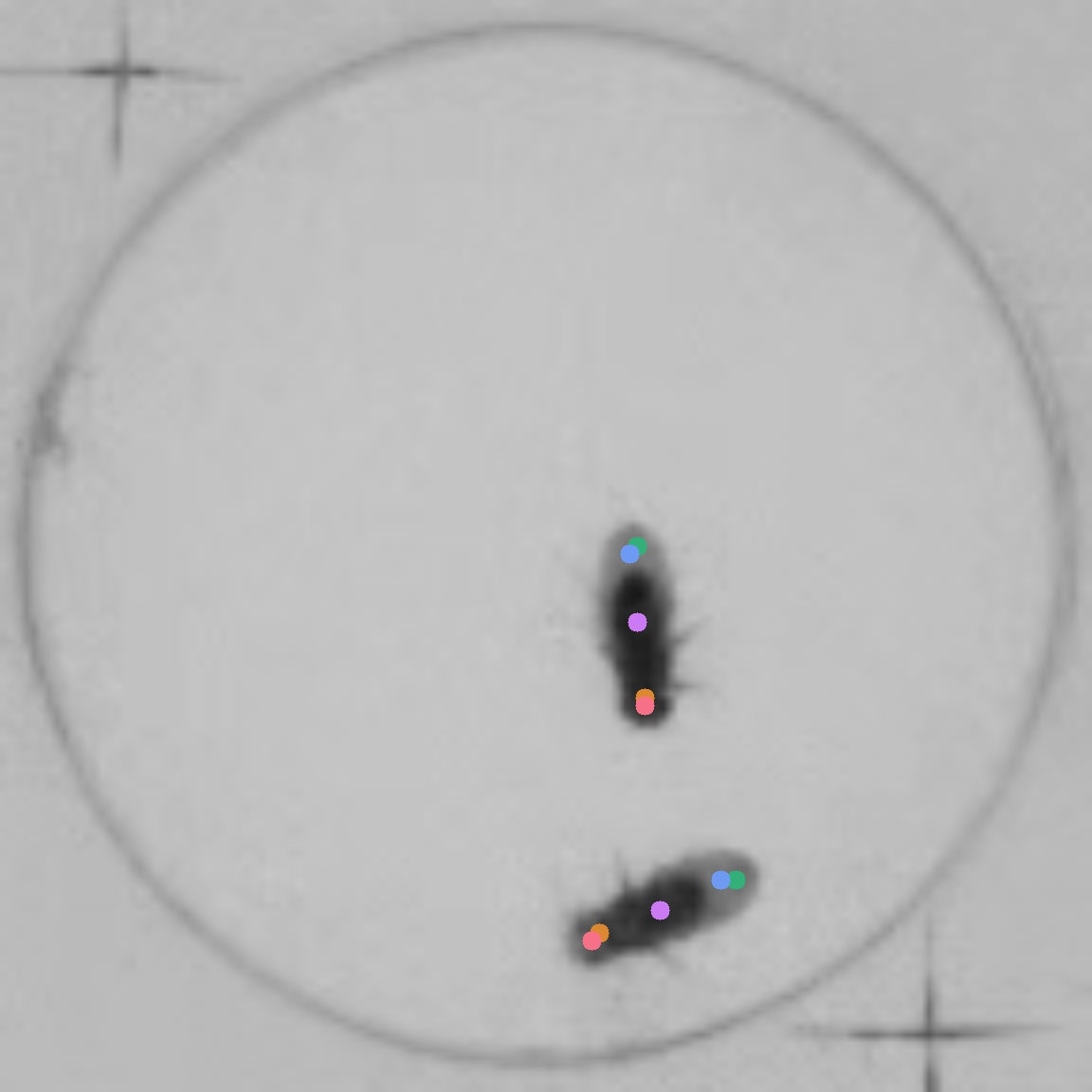}\hspace{\fighspace} & \includegraphics[width=\figsize\textwidth]{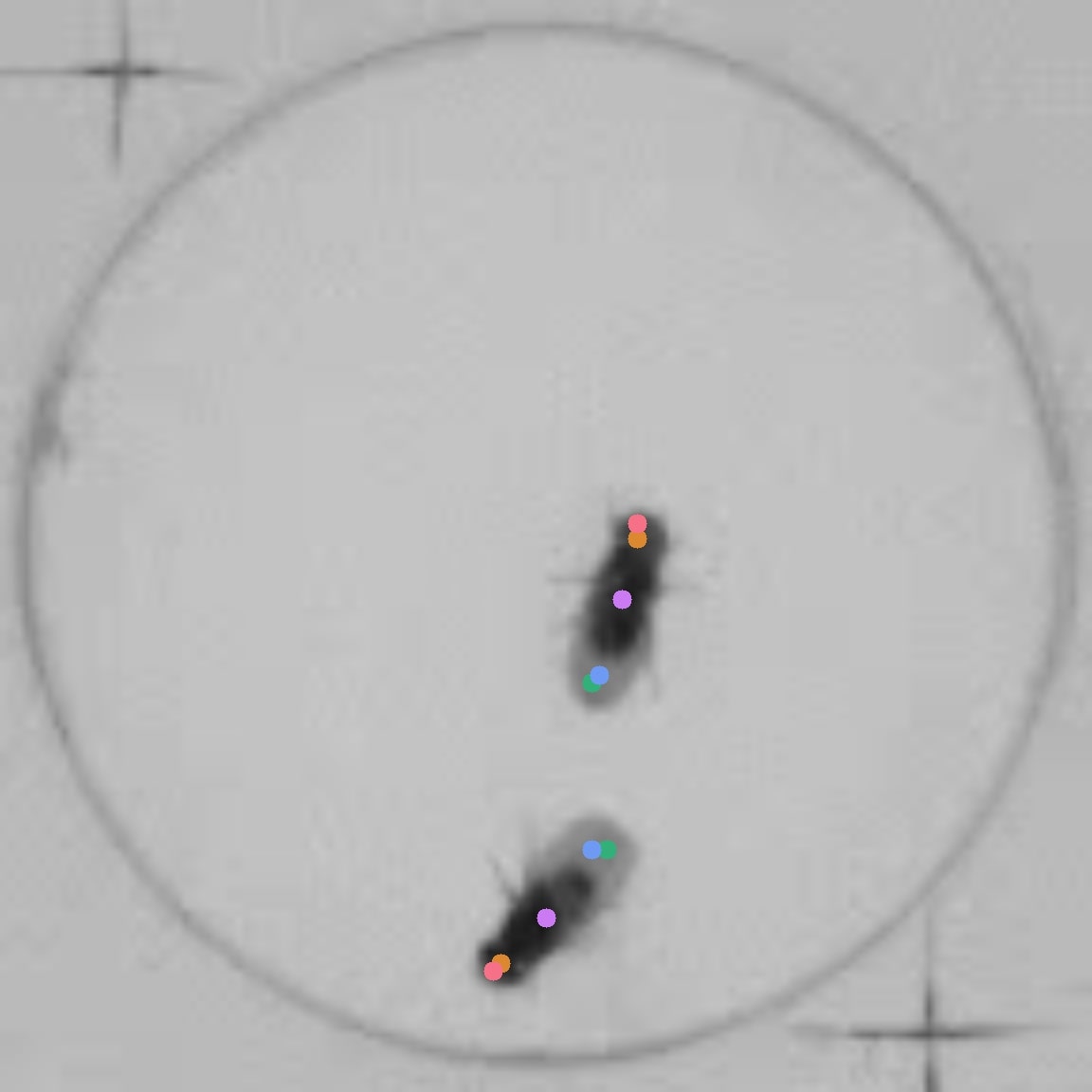}\hspace{\fighspacer} & \includegraphics[width=\figsize\textwidth]{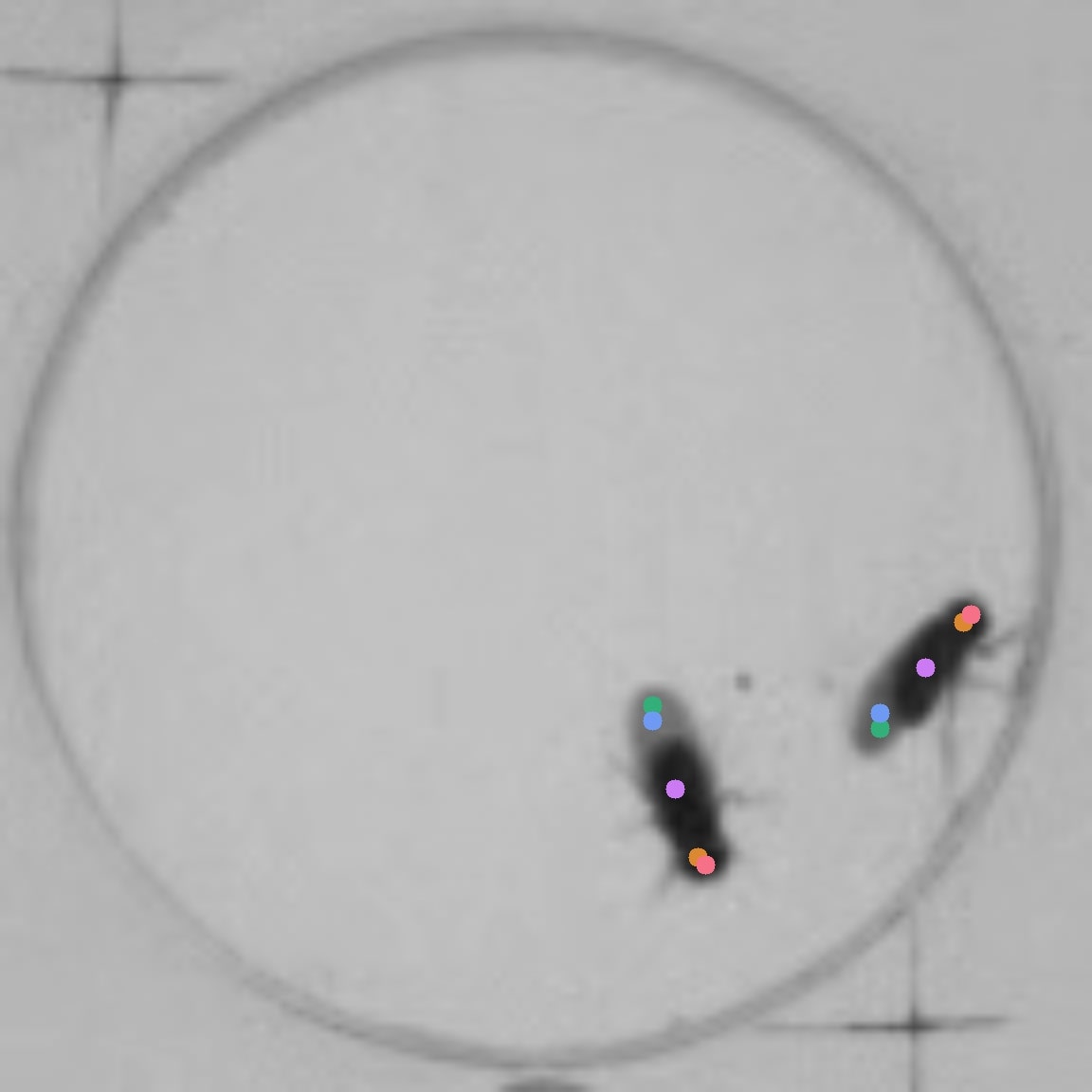}\hspace{\fighspacer} & \includegraphics[width=\figsize\textwidth]{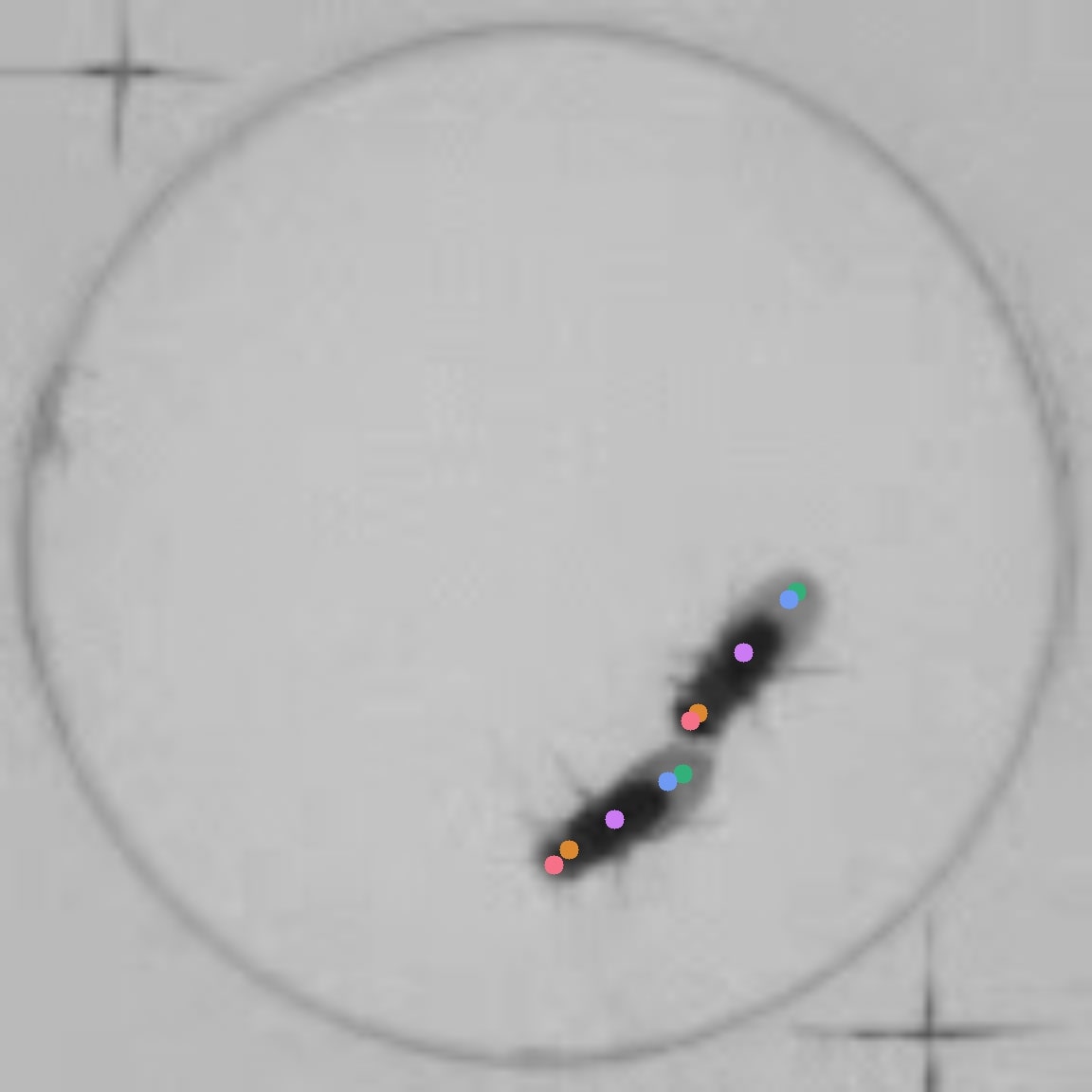}\hspace{\fighspacer} \\
\end{tabular}
\caption{\textbf{Qualitative Results on Fly-vs-Fly}. We observe that 3 keypoints are discovered on the body of the fly, with 2 on the wings (one for each wing).}
\label{fig:fly_qual}
\end{figure*}

\def\figsize{0.15}
\def\fighspace{-2mm}
\def\fighspacer{-2mm}
\begin{figure*}[h]
\centering
\begin{tabular}{cccccc}
\centering
\hspace{\fighspace}\includegraphics[width=\figsize\textwidth]{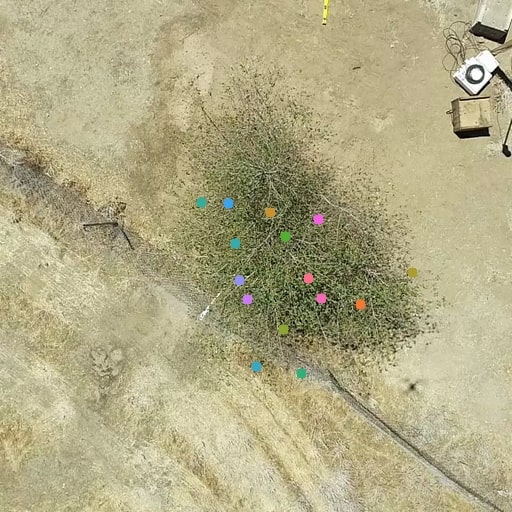}\hspace{\fighspace} & \includegraphics[width=\figsize\textwidth]{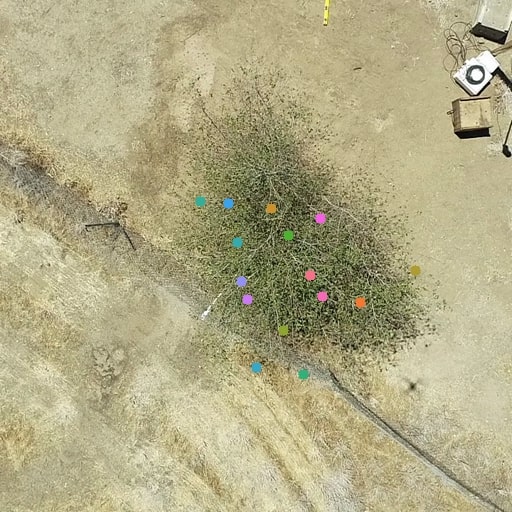}\hspace{\fighspacer} & \includegraphics[width=\figsize\textwidth]{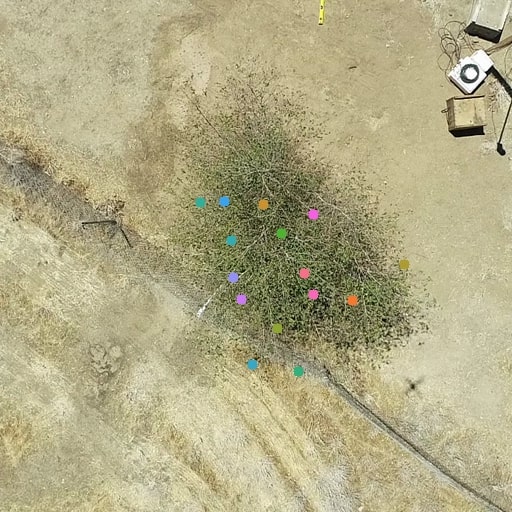}\hspace{\fighspace} & 
\includegraphics[width=\figsize\textwidth]{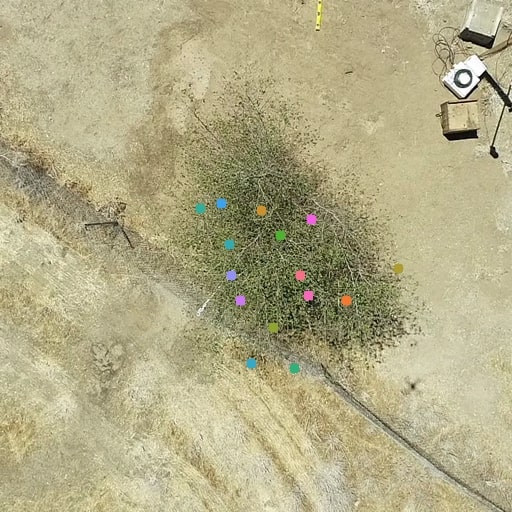}\hspace{\fighspace} & \includegraphics[width=\figsize\textwidth]{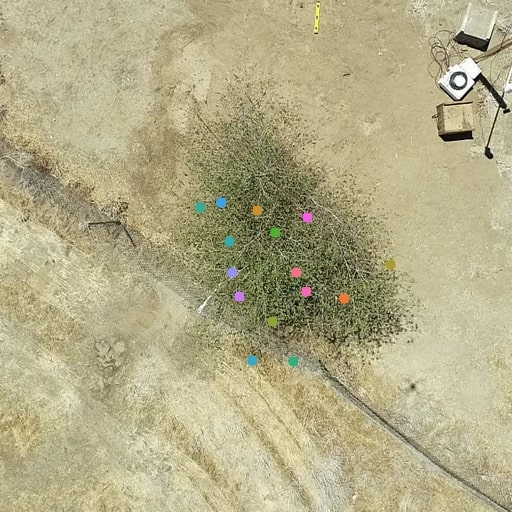}\hspace{\fighspacer} & \includegraphics[width=\figsize\textwidth]{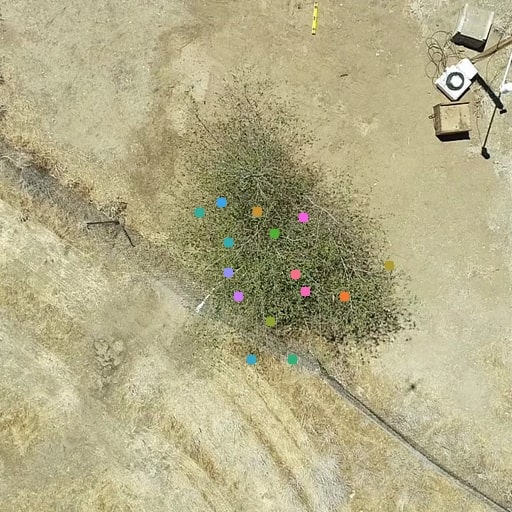}\hspace{\fighspacer}  \\
\hspace{\fighspace}\includegraphics[width=\figsize\textwidth]{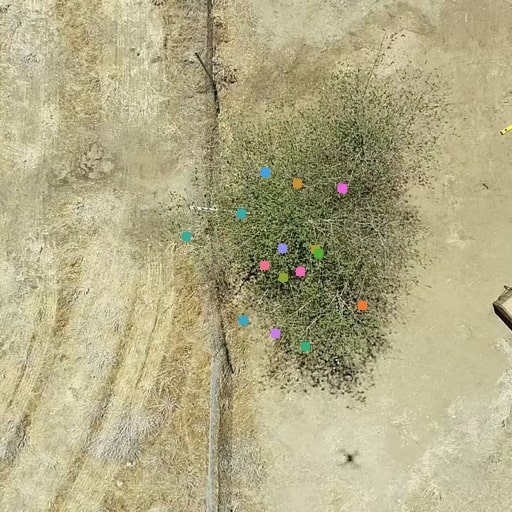}\hspace{\fighspace} & \includegraphics[width=\figsize\textwidth]{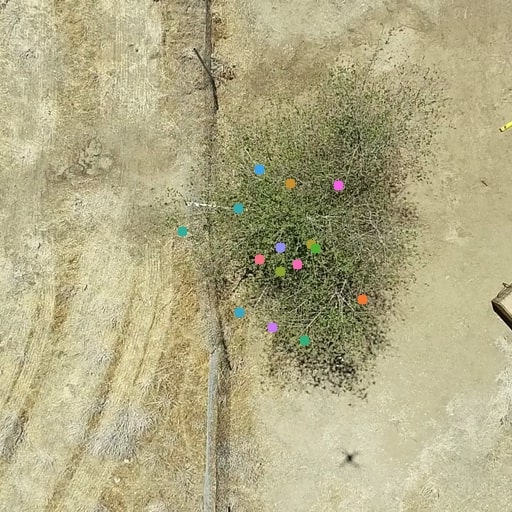}\hspace{\fighspacer} & \includegraphics[width=\figsize\textwidth]{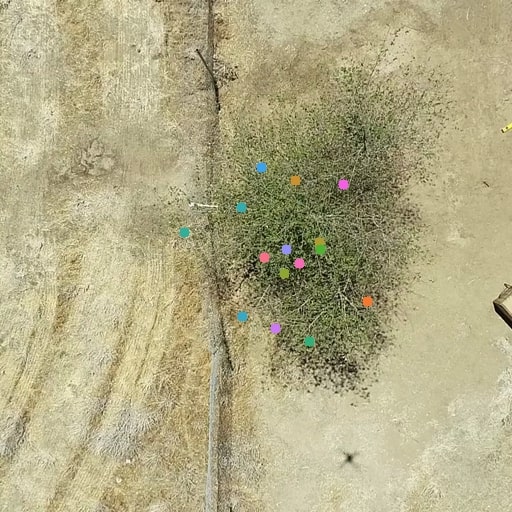}\hspace{\fighspace} & \includegraphics[width=\figsize\textwidth]{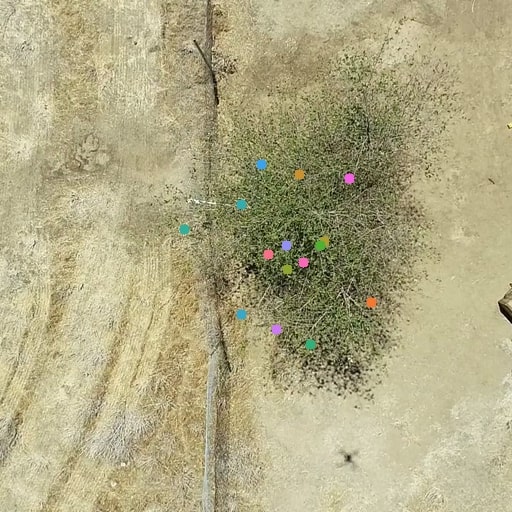}\hspace{\fighspacer} & \includegraphics[width=\figsize\textwidth]{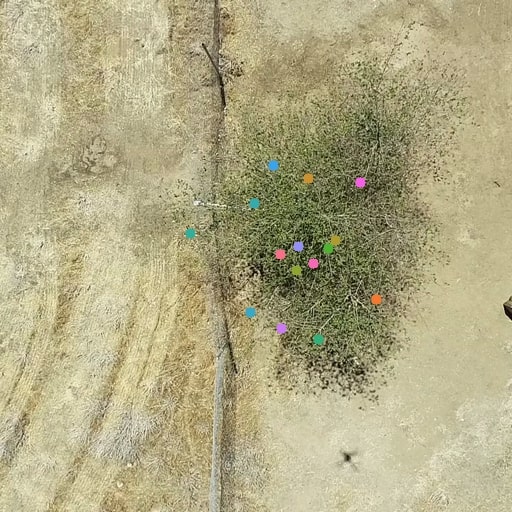}\hspace{\fighspacer} & \includegraphics[width=\figsize\textwidth]{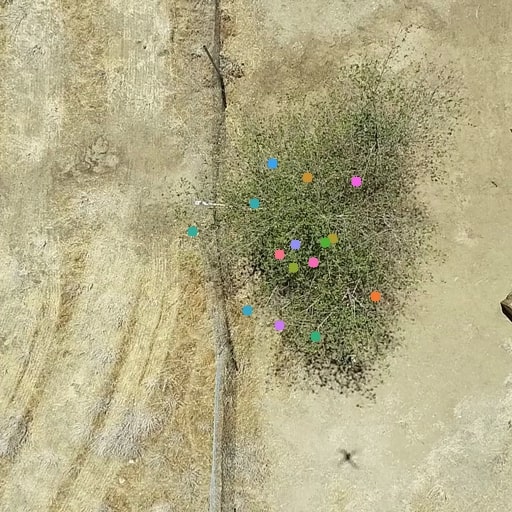}\hspace{\fighspacer} \\
\end{tabular}
\caption{\textbf{Qualitative Results on Vegetations}. Each row shows different frames with discovered keypoints from a single video. Our model can discover and track consistent keypoints within the same video.}\vspace{-0.2cm}
\label{fig:tree_qual}
\end{figure*}

\begin{figure*}[h]
\centering
\begin{tabular}{cccccc}
\centering
\hspace{\fighspace}\includegraphics[width=\figsize\textwidth]{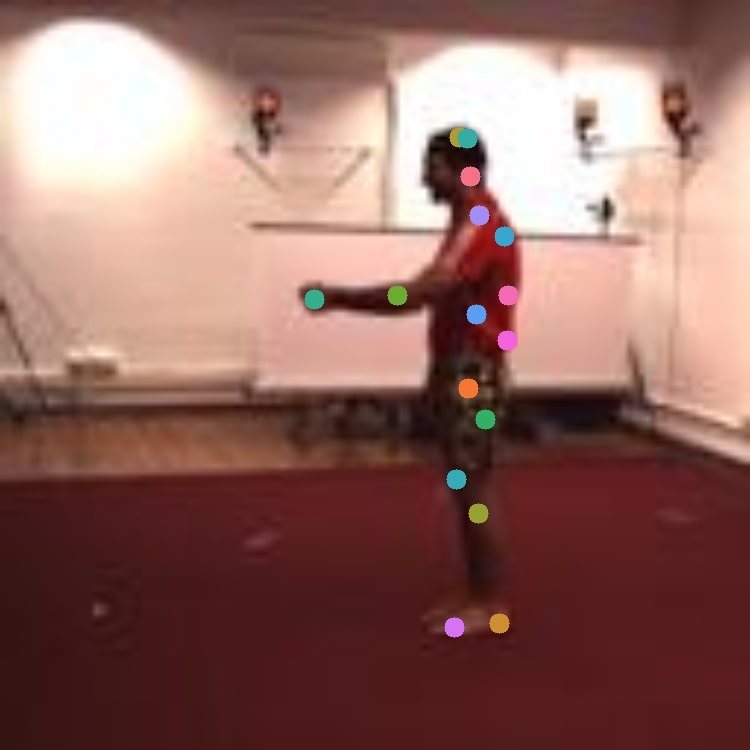}\hspace{\fighspace} & \includegraphics[width=\figsize\textwidth]{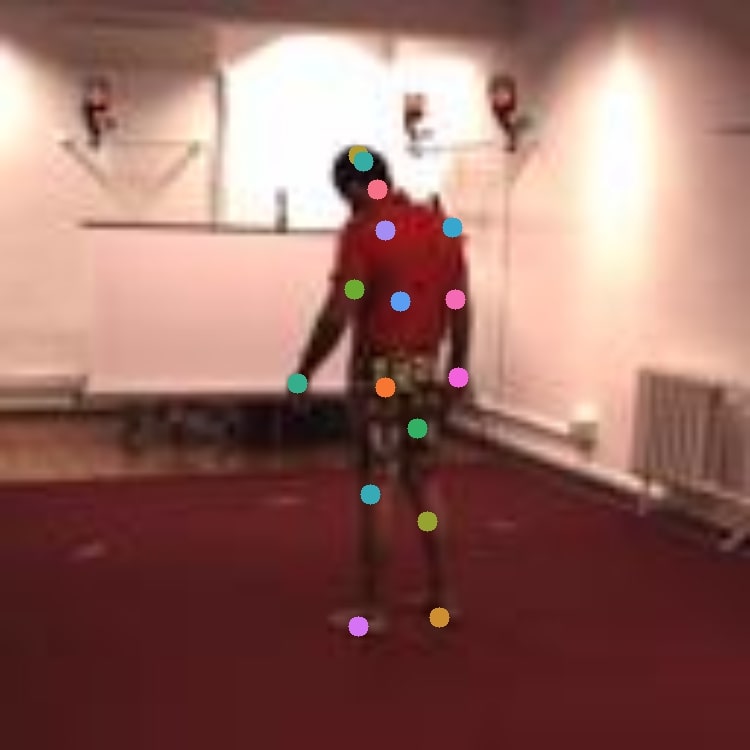}\hspace{\fighspacer} & \includegraphics[width=\figsize\textwidth]{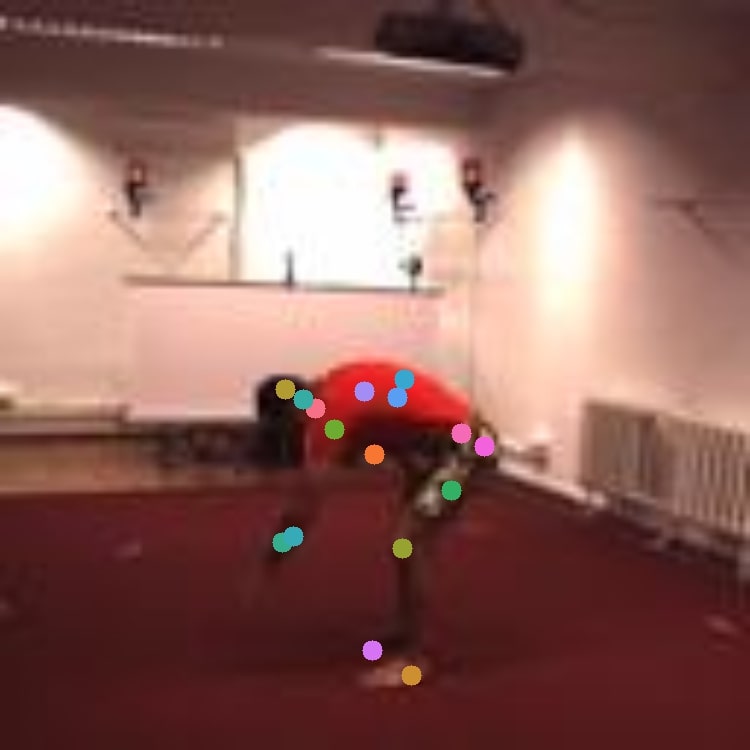}\hspace{\fighspace} & 
\includegraphics[width=\figsize\textwidth]{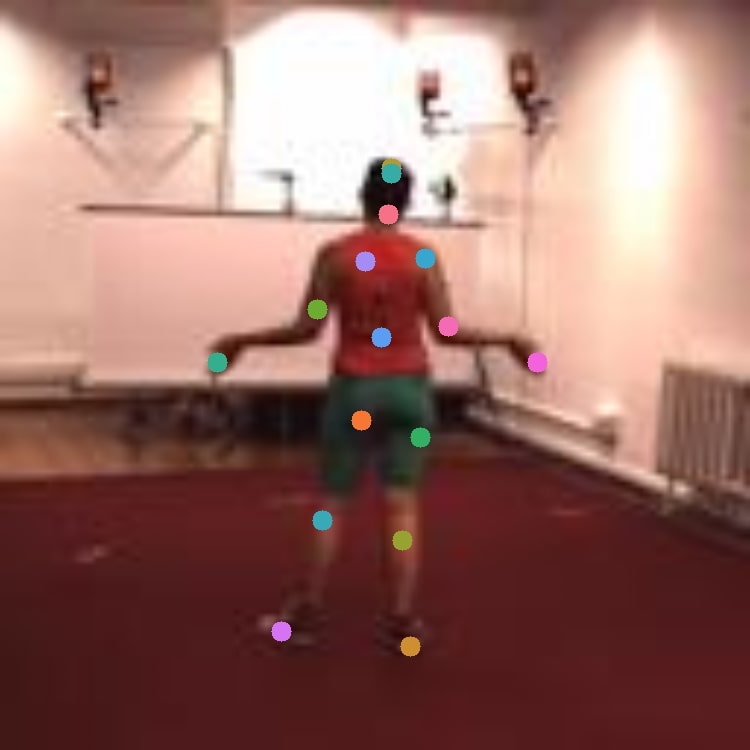}\hspace{\fighspace} & \includegraphics[width=\figsize\textwidth]{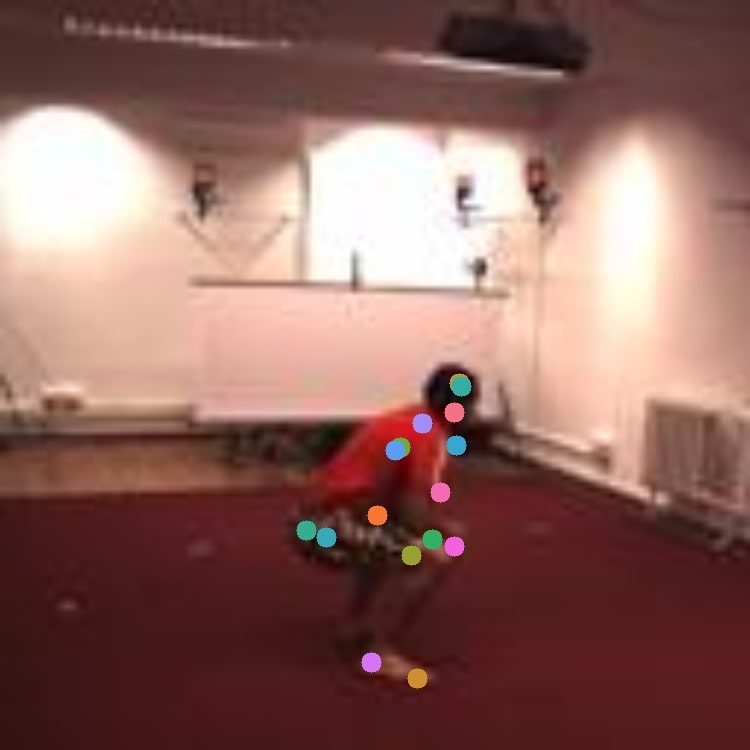}\hspace{\fighspacer} & \includegraphics[width=\figsize\textwidth]{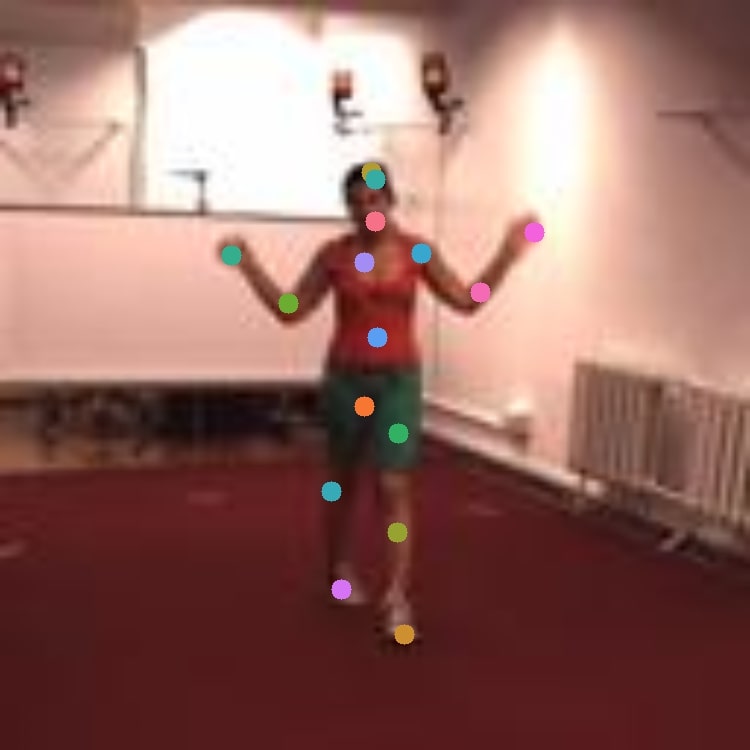}\hspace{\fighspacer}  \\
\hspace{\fighspace}\includegraphics[width=\figsize\textwidth]{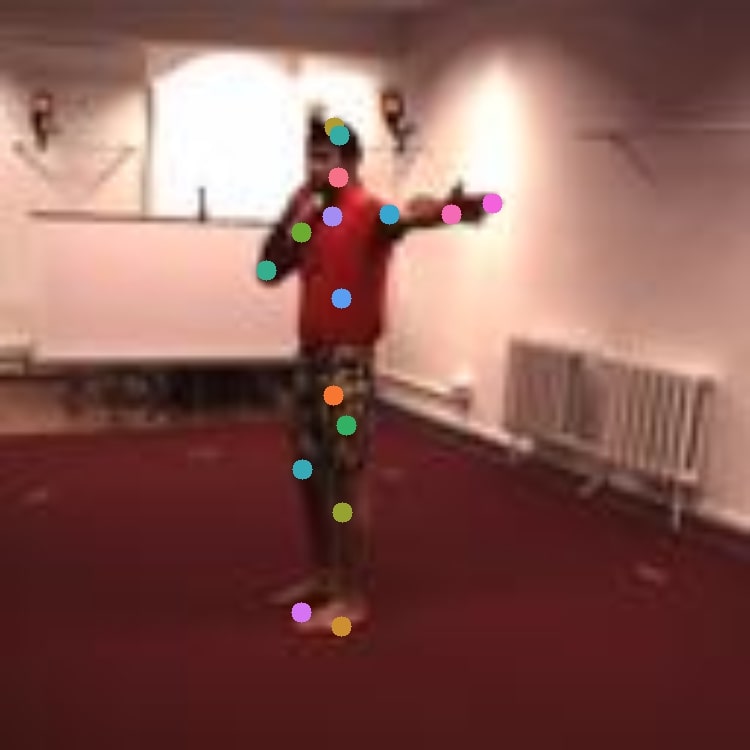}\hspace{\fighspace} & \includegraphics[width=\figsize\textwidth]{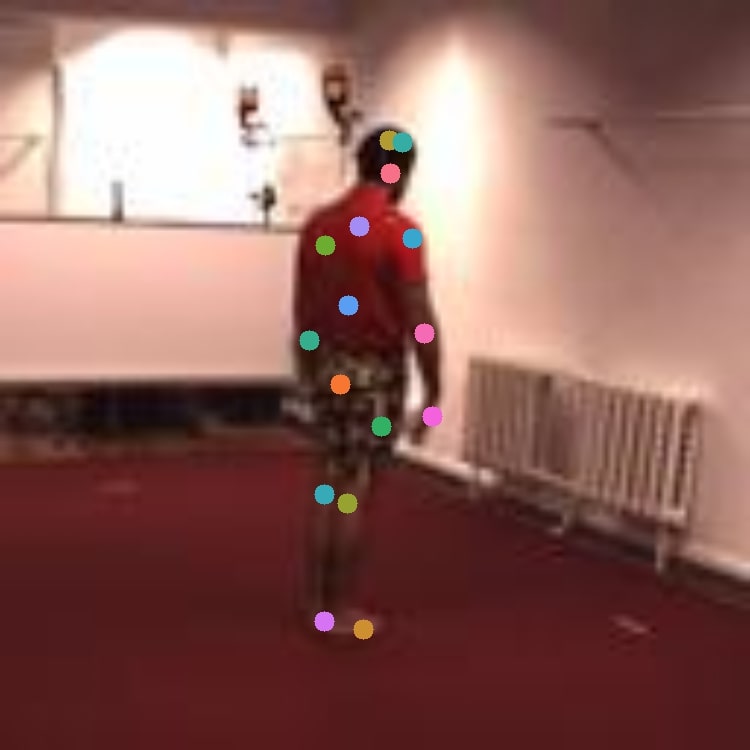}\hspace{\fighspacer} & \includegraphics[width=\figsize\textwidth]{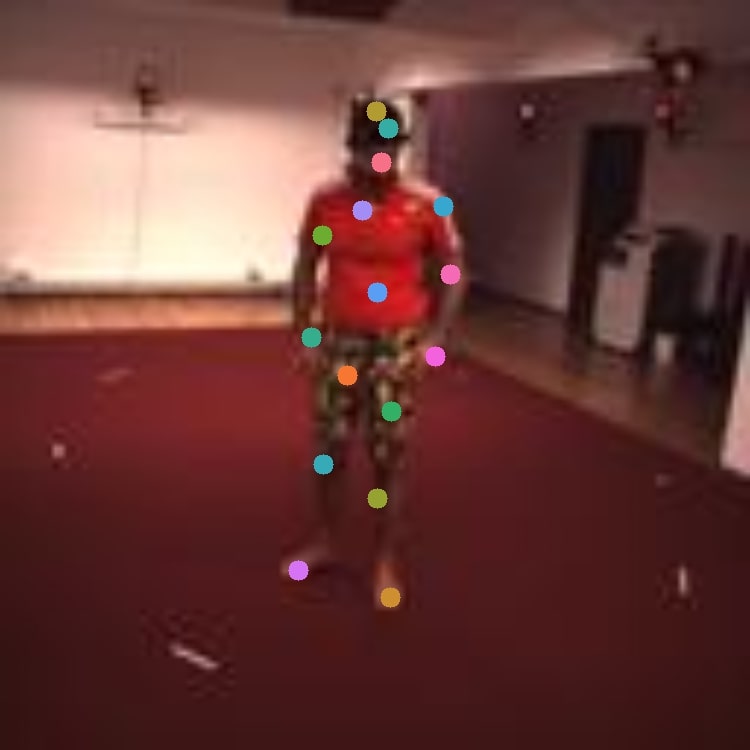}\hspace{\fighspace} & \includegraphics[width=\figsize\textwidth]{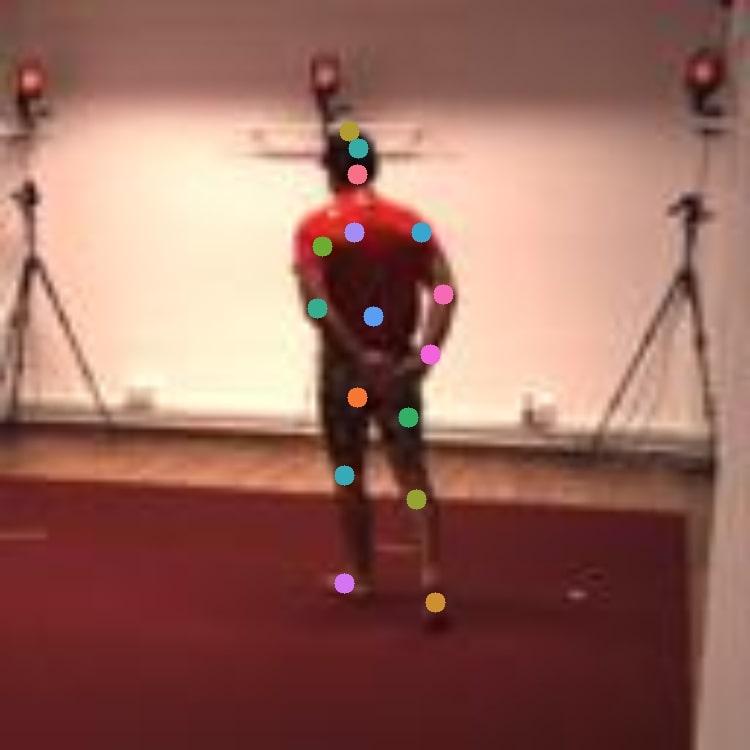}\hspace{\fighspacer} & \includegraphics[width=\figsize\textwidth]{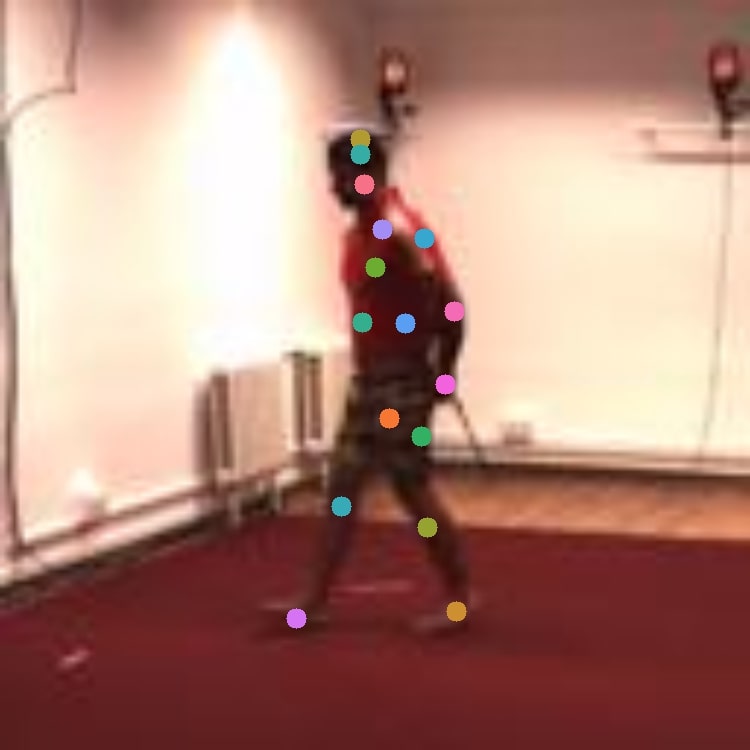}\hspace{\fighspacer} & \includegraphics[width=\figsize\textwidth]{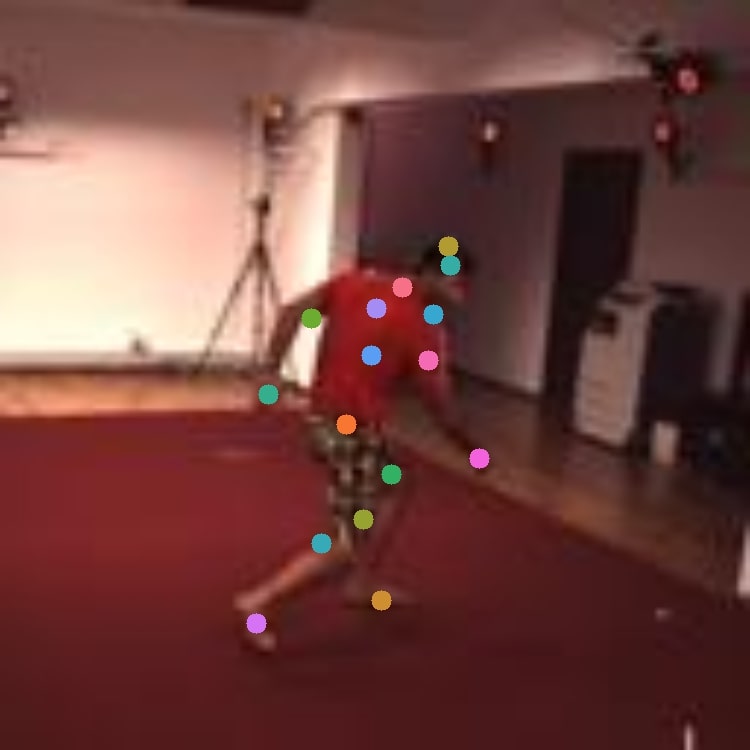}\hspace{\fighspacer} \\
\end{tabular}
\caption{\textbf{Qualitative Results on Simplified Human 3.6M}. We observe that keypoints are generally discovered on visible joints and end points of humans, such as head, elbows, hands, upper legs, knees and feet. We note that there is left/right swapping of body parts, since when the human is facing forwards or backwards, keypoints are generally on the same side.}
\label{fig:human_qual}
\end{figure*}

\begin{figure*}[h]
\centering
\begin{tabular}{cccccc}
\centering
\hspace{\fighspace}\includegraphics[width=\figsize\textwidth,height=\figsize\textwidth]{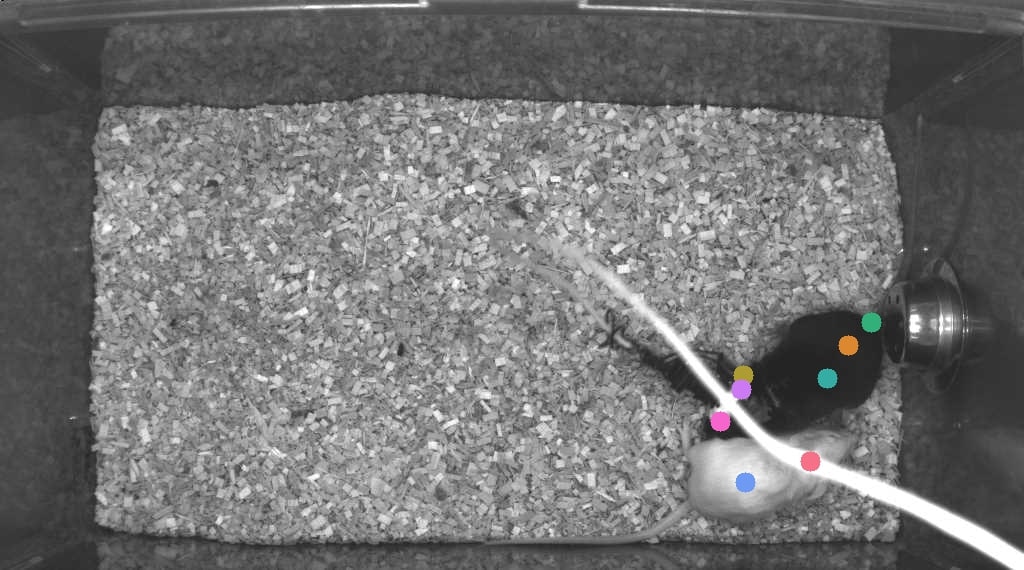}\hspace{\fighspace} & \includegraphics[width=\figsize\textwidth,height=\figsize\textwidth]{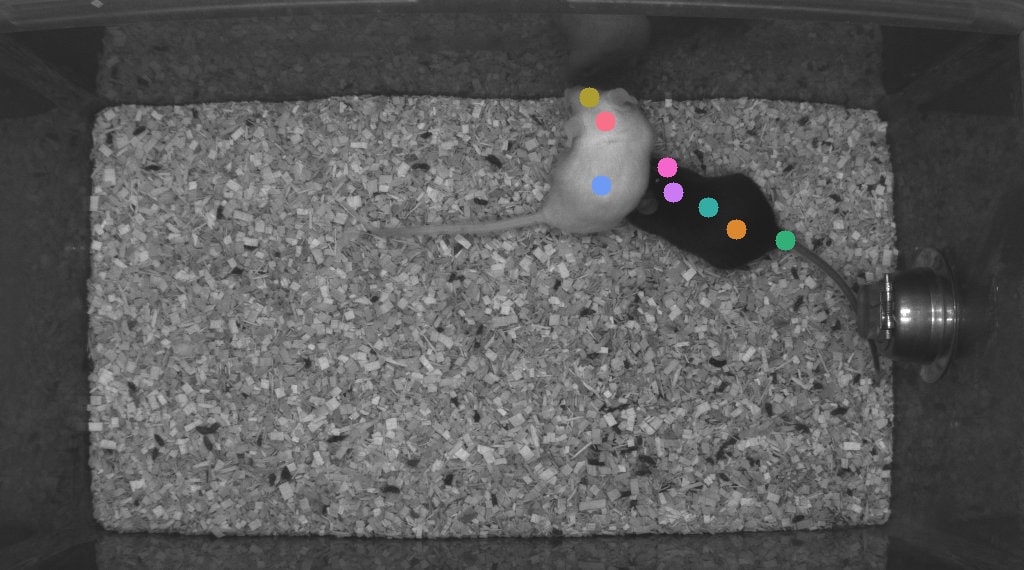}\hspace{\fighspacer} & \includegraphics[width=\figsize\textwidth]{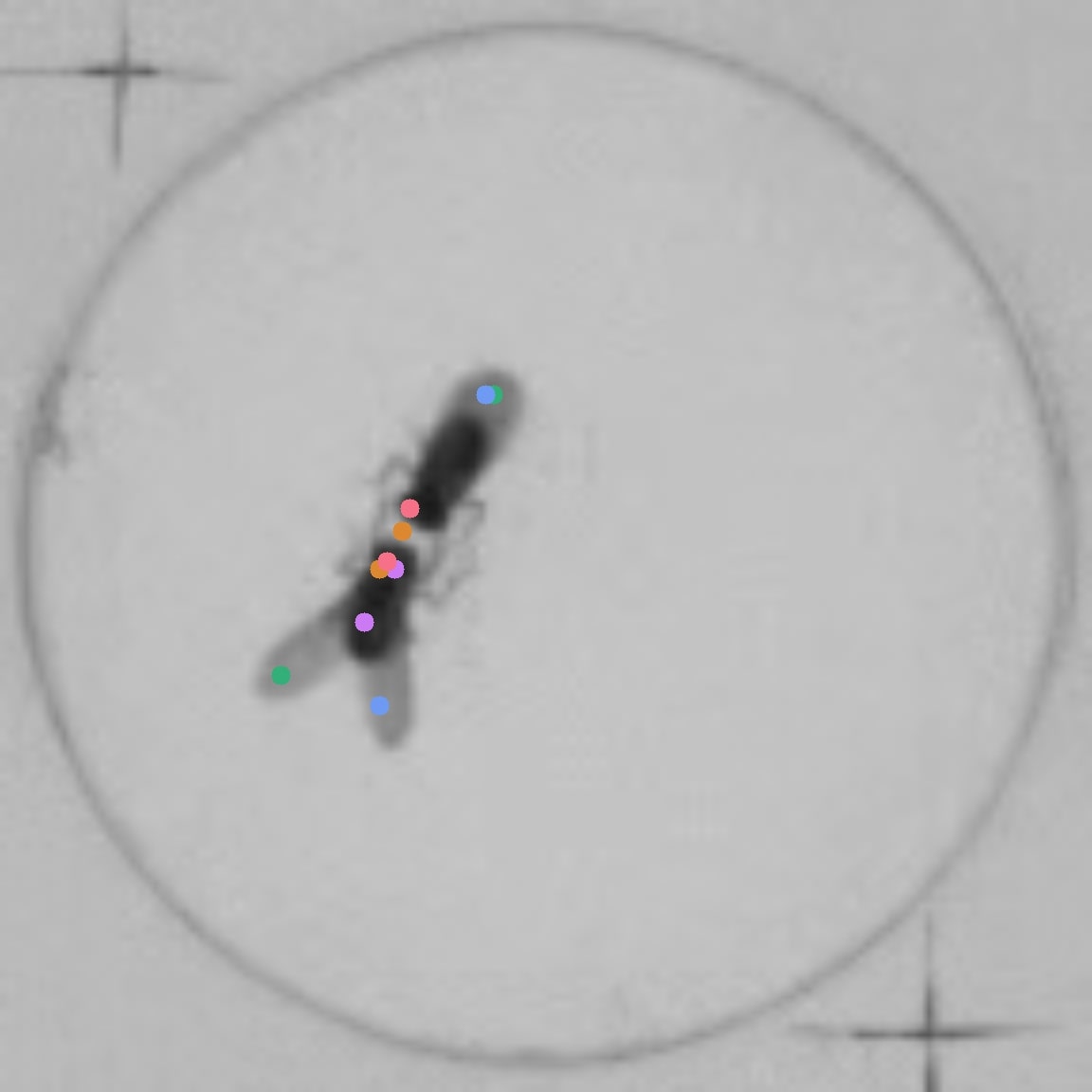}\hspace{\fighspace} & 
\includegraphics[width=\figsize\textwidth]{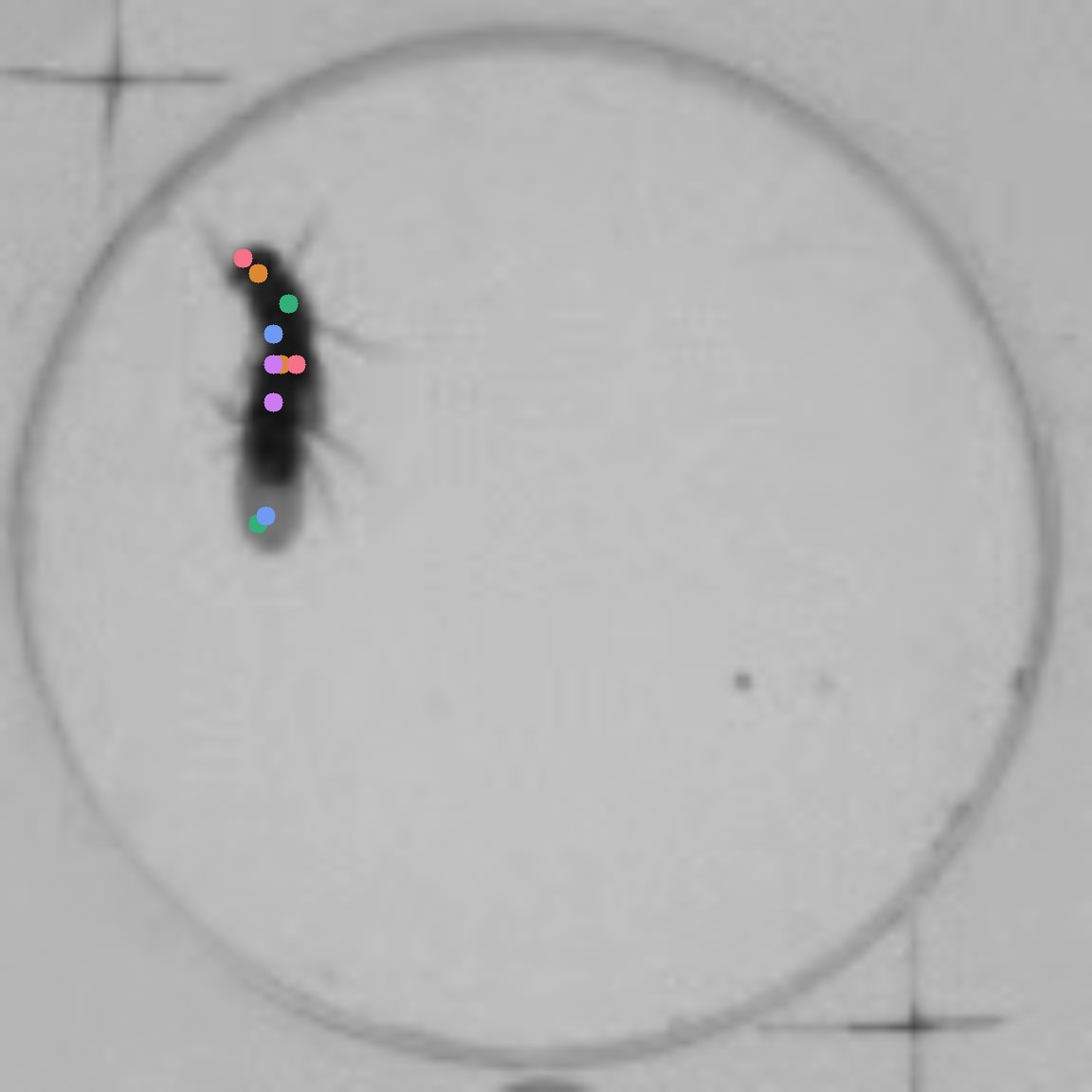}\hspace{\fighspace} & \includegraphics[width=\figsize\textwidth]{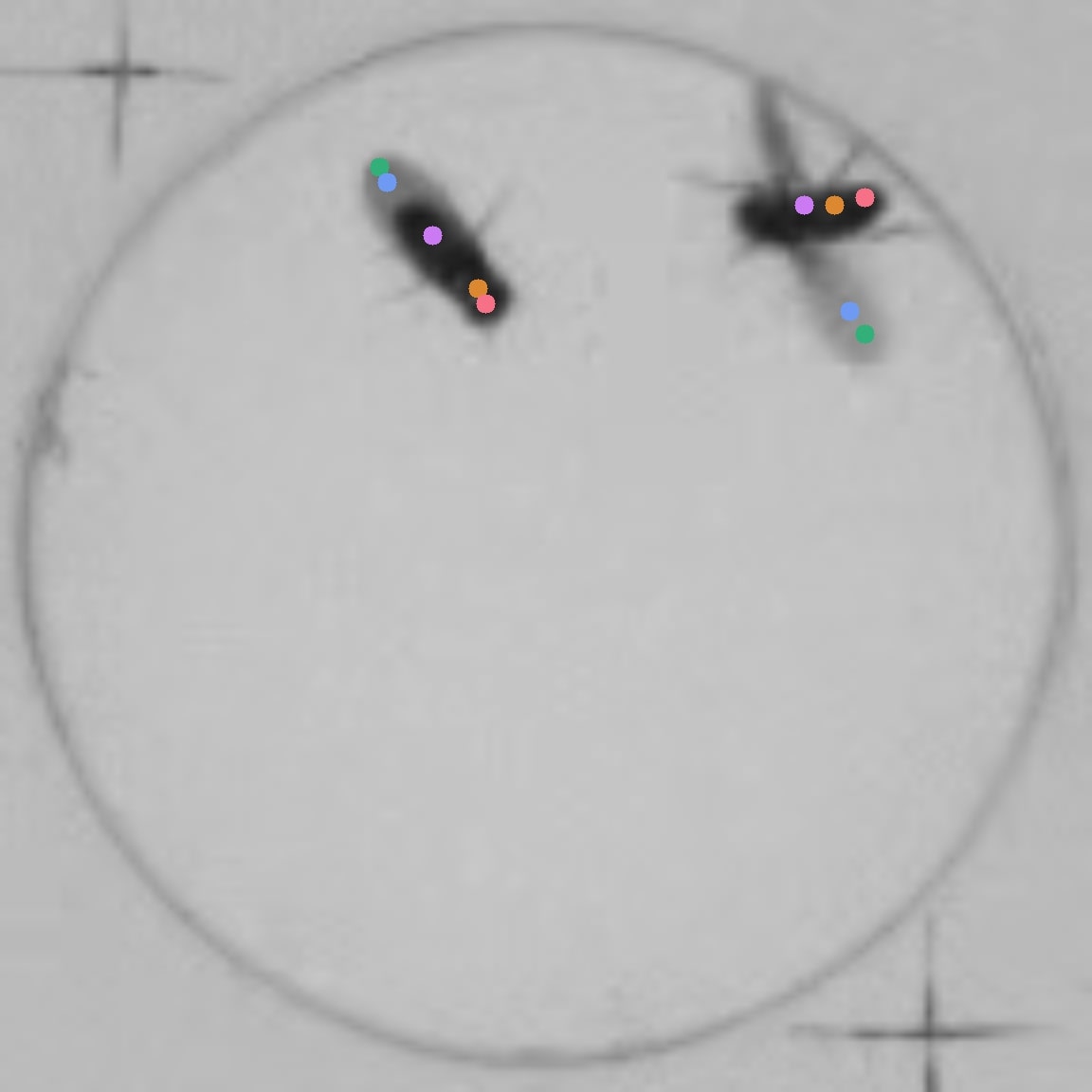}\hspace{\fighspacer} & \includegraphics[width=\figsize\textwidth]{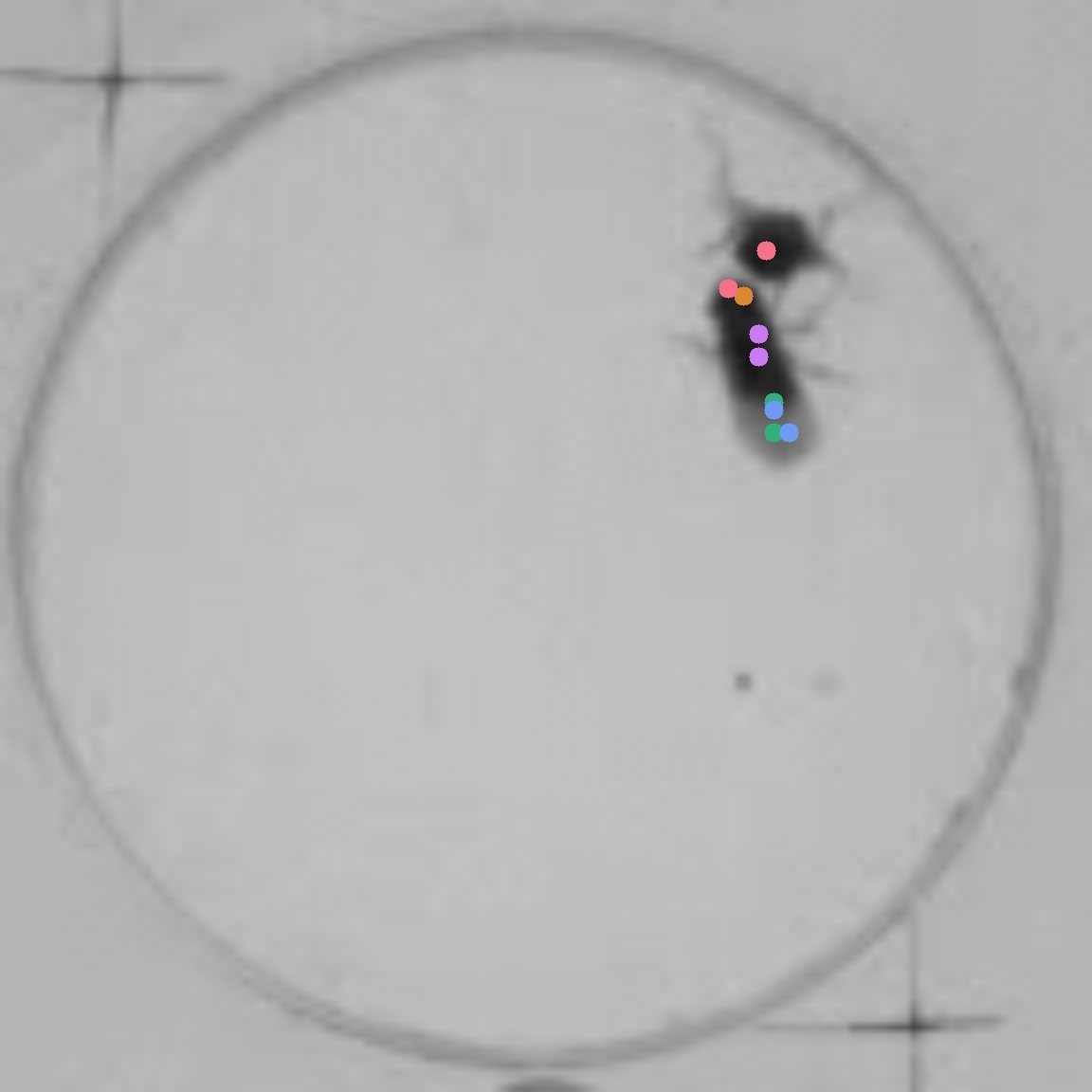}\hspace{\fighspacer}  \\
\hspace{\fighspace}\includegraphics[width=\figsize\textwidth,height=\figsize\textwidth]{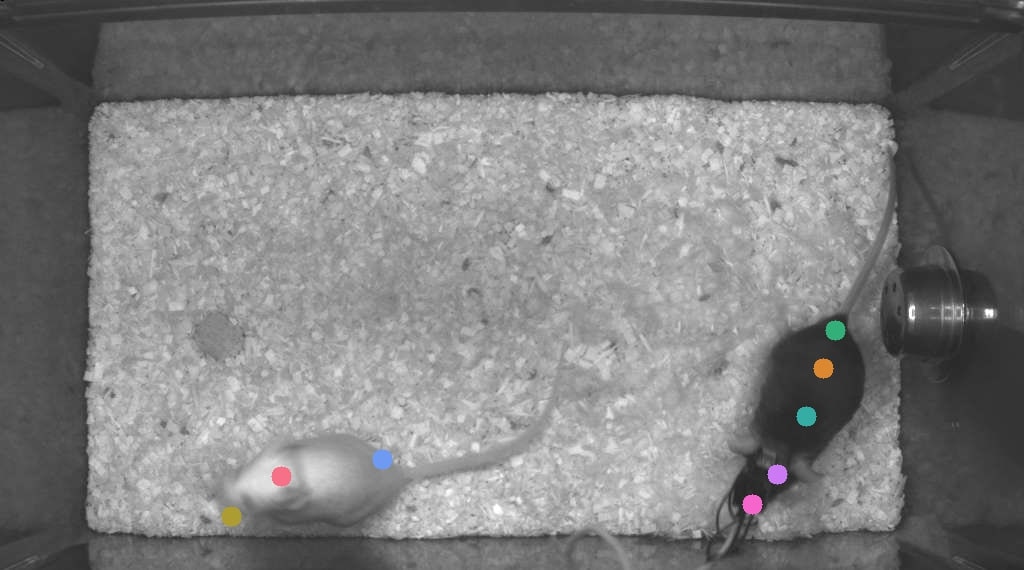}\hspace{\fighspace} & \includegraphics[width=\figsize\textwidth,height=\figsize\textwidth]{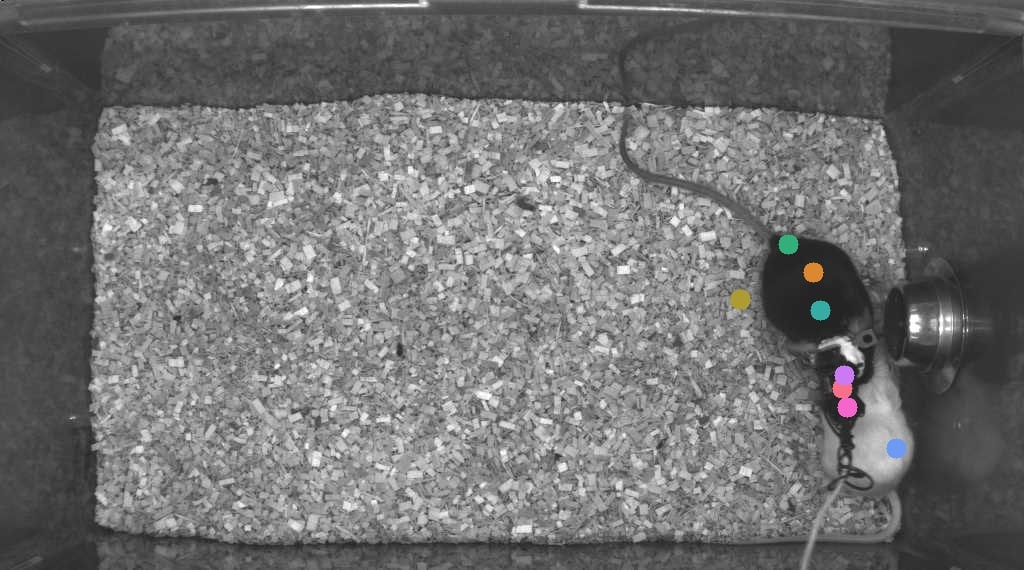}\hspace{\fighspacer} & \includegraphics[width=\figsize\textwidth]{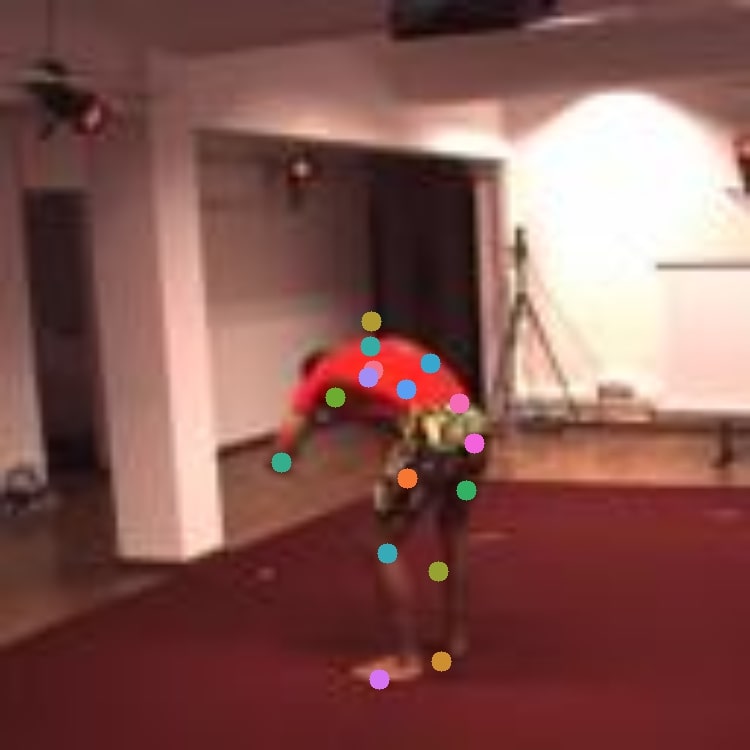}\hspace{\fighspace} & \includegraphics[width=\figsize\textwidth]{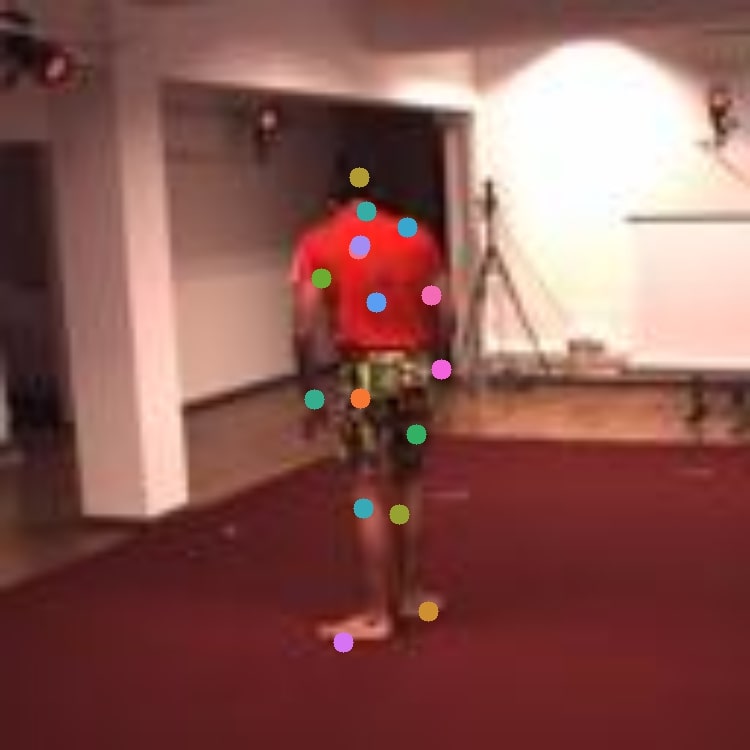}\hspace{\fighspacer} & \includegraphics[width=\figsize\textwidth]{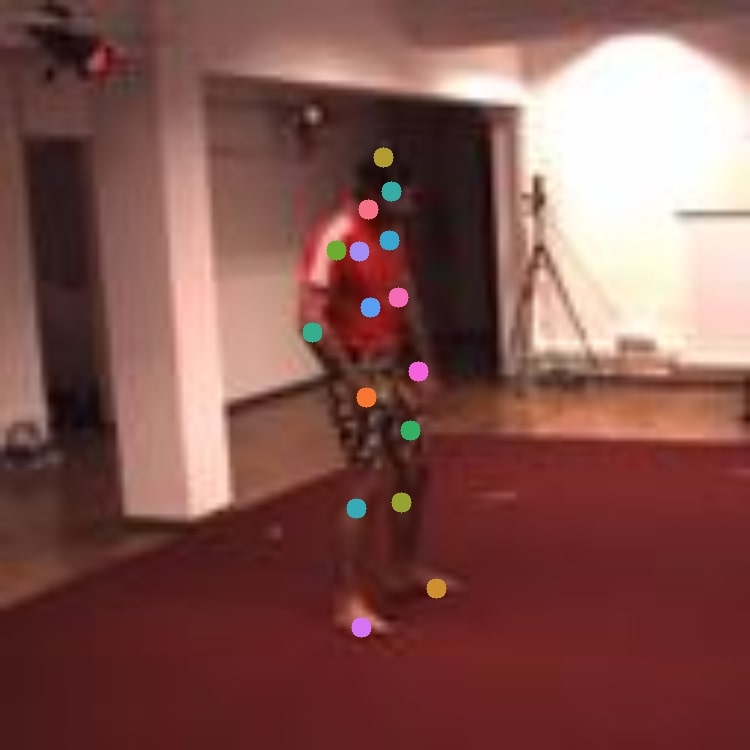}\hspace{\fighspacer} & \includegraphics[width=\figsize\textwidth]{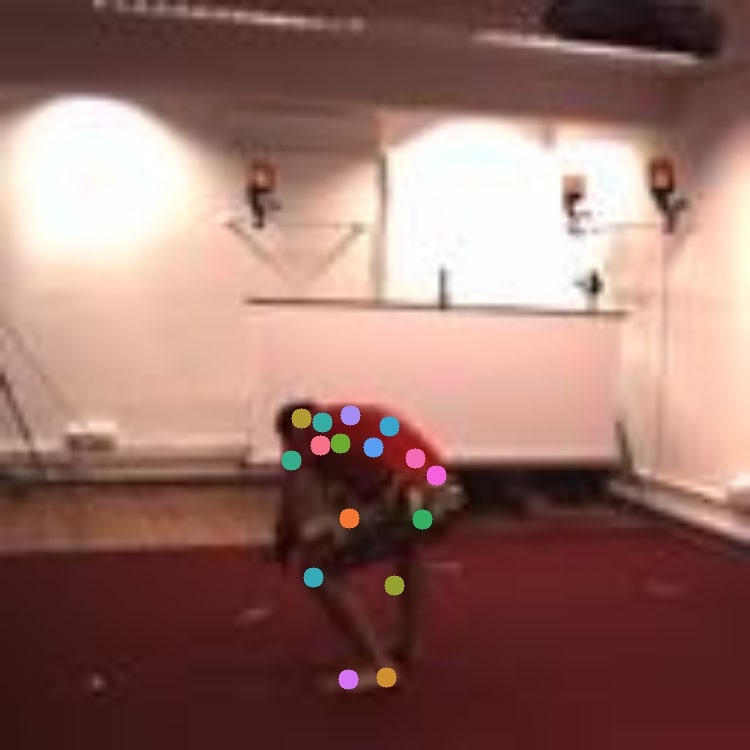}\hspace{\fighspacer} \\
\end{tabular}
\caption{\textbf{Limitations}. We visualize examples that are difficult for our model, for example from occlusion/agents being in close proximity (mouse, fly), self-occlusion (human), unusual poses (human, fly), and left-right swapping (human).}
\label{fig:failure_qual}
\end{figure*}

\end{document}